\documentclass[runningheads]{llncs}
\usepackage{graphicx}
\usepackage{comment}
\usepackage{amsmath,amssymb} % define this before the line numbering.
\usepackage{color}

\usepackage[accsupp]{axessibility}  % Improves PDF readability for those with disabilities.

\usepackage{graphicx}
\usepackage{booktabs}

\usepackage{epsfig}
\usepackage{tabularx}
\usepackage[table]{xcolor}       % for colored stuff
\usepackage{booktabs}
\usepackage{relsize}
\usepackage{multirow}
\usepackage{scrextend}           % Put footmark in the bottom of the page
\usepackage{array}               % Thicker vertical rules in tables, lines in only certain columns
\usepackage{makecell}            % For thicker horizontal rules in tables
\usepackage{subcaption}          % For subfigures
\usepackage{wasysym}             % For right filled arrow
\usepackage{lipsum}              % For random filler text
\usepackage{esvect}              % For vv command arrow above symbols
\usepackage{soul}                % For striked text command \st and hightlight
\usepackage{float}
\usepackage{bm}
\usepackage{cite}                % orders the citations in increasing order
\usepackage{paralist}            % for compactitem
\usepackage{enumitem}            % for noindent itemize
\usepackage{empheq}
\usepackage{gensymb}             % for degree symbol
\usepackage{wrapfig}             % for wrapped figure, wraptable

\definecolor{lightgreen}{HTML}{90EE90}

\allowdisplaybreaks              % Break aligned equations

\graphicspath{{images/}}

\def\cccolorbox#1#2{\ifx#2\relax\let\next\allowbreak\else
       \def\next{\colorbox{#1}{\strut #2}\allowbreak\cccolorbox{#1}}\fi\next}
\def\ccolorbox#1#2{\fboxsep0pt\cccolorbox{#1}#2\relax}

\usepackage{tikz} % for some figures
\usetikzlibrary{calc}  % for centerarc
\usetikzlibrary{arrows}
\usetikzlibrary{arrows.meta}
\usetikzlibrary{shapes.arrows}
\usetikzlibrary{fadings, shadows}
\usetikzlibrary{decorations.text}
\def\centerarc[#1](#2)(#3:#4:#5)% Syntax: [draw options] (center) (initial angle:final angle:radius)
    { \draw[#1] ($(#2)+({#5*cos(#3)},{#5*sin(#3)})$) arc (#3:#4:#5); }

\setlist[itemize]{noitemsep,topsep=0pt,leftmargin=*,label={\large\textbullet}}
\definecolor{my_blue}{rgb}{0.2, 0.6, 1}  % dodgerblue
\definecolor{my_magenta}{rgb}{1.0, 0.2, 0.6} % triadic to dodgerblue
\definecolor{my_yellow}{rgb}{1.0, 0.8, 0.2} % triadic to magenta
\definecolor{my_green}{rgb}{0.0, 0.9, 0.24}
\definecolor{my_green_2}{rgb}{0.0, 0.4, 0.0}

\definecolor{set1_cyan}{rgb}{0.23, 0.87, 1.0}

\definecolor{backward_color}{rgb}{1.0, 0.6, 0.2}
\definecolor{forward_color}{rgb}{0.2, 1.0, 0.6}

\definecolor{my_violet}{rgb}{0.79, 0.40, 1} %{0.73,0.62,0.91}  %
\definecolor{my_red}{rgb}{1,0,0}
\definecolor{my_purple}{rgb}{0.27,0.8, 0.8}
\definecolor{new_green}{rgb}{0.75,0.97,0.44}
\definecolor{my_orange}{rgb}{1.0,0.6,0.35}

\definecolor{lightgray}{rgb}{0.96, 0.96, 0.96}
\definecolor{white}{rgb}{1.0, 1.0, 1.0}

\definecolor{projectionColor}{rgb}{0.2, 0.6, 1}
\definecolor{rayColor}{rgb}{0.0,0.0,0.0}
\definecolor{axisColor}{rgb}{0.0, 0.0, 0.0}
\definecolor{rayColor}{rgb}{0.0, 0.0, 0.0}

\colorlet{projectionBorderShade}{rayColor!100}
\colorlet{projectionFillShade}{projectionColor!20}
\colorlet{rayShade}{my_yellow!100}
\colorlet{axisShade}{axisColor!20}
\colorlet{axisShadeDark}{axisColor!100}

\definecolor{LightCyan}{rgb}{0.88,1,1}

\definecolor{darkGreen}{rgb}{0.01, 0.8, 0.24}%{0.31, 0.94, 0.3}%{0.09,0.88,0.00}{0.29, 0.83, 0.38}

\colorlet{proposedShade}{blue}
\colorlet{vanillaShade}{red!90}

\newcommand{\scaleFraction}{0.72}
\newcommand{\figureScaleFraction}{1.0}
\newcommand{\waymoFigureScaleFraction}{0.7}

\newcommand{\identity}{{\bf{I}}}

\newcommand{\oneD}{$1$D}
\newcommand{\twoD}{$2$D}
\newcommand{\threeD}{$3$D}
\newcommand{\fourD}{$4$D}
\newcommand{\fiveD}{$5$D}
\newcommand{\twoDMath}{2\text{D}}
\newcommand{\threeDMath}{3\text{D}}
\newcommand{\iou}{IoU}
\newcommand{\iouTwoD}{\iou$_{2\text{D}}$}
\newcommand{\iouThreeD}{\iou$_{3\text{D}}$}
\newcommand{\lidar}{LiDAR}

\newcommand{\equivariant} {equivariant}
\newcommand{\Equivariant} {Equivariant}
\newcommand{\equivariance}{equivariance}
\newcommand{\Equivariance}{Equivariance}
\newcommand{\scaleEquivariant} {scale \equivariant}
\newcommand{\ScaleEquivariant} {Scale \Equivariant}
\newcommand{\scaleEquivariance}{scale \equivariance}
\newcommand{\ScaleEquivariance}{Scale \Equivariance}
\newcommand{\depthEquivariant} {depth \equivariant}
\newcommand{\DepthEquivariant} {Depth \Equivariant}
\newcommand{\depthEquivariance} {depth \equivariance}

\newcommand{\MaxScale}{Scale-Projection}
\newcommand{\ses}{SES}
\newcommand{\se}{SE}
\newcommand{\weight}{\mathbf{w}}
\newcommand{\plane}{patch plane}
\newcommand{\transformationGroup}{G}
\newcommand{\transformationGroupMember}{g}
\newcommand{\inputData}{h}%\mathcal{H}}
\newcommand{\outputData}{y}%\mathcal{Y}}
\newcommand{\mapping}{\Phi}
\newcommand{\transformationMath}{\mathcal{T}}
\newcommand{\transformationInput}{\transformationMath_\transformationGroupMember^\inputData}
\newcommand{\transformationOutput}{\transformationMath_\transformationGroupMember^\outputData}
\newcommand{\manifold}{manifold}

\newcommand{\transformation}{transformation}

\newcommand{\mySign}{\!+\!}
\newcommand{\mathDash}{$-$}

\newcommand{\varX}{x}
\newcommand{\varY}{y}
\newcommand{\varZ}{z}
\newcommand{\posX}{X}
\newcommand{\posY}{Y}
\newcommand{\posZ}{Z}

\newcommand{\norm}[1]{\left|#1\right|}

\newcommand{\pointCloudOne}{H}

\newcommand{\projectionOperator}{\mathbf{K}}%\mathrm{\pi}}

\newcommand{\rotation}{\mathbf{R}}
\newcommand{\translation}{\mathbf{t}}
\newcommand{\transX}{t_\varX}
\newcommand{\transY}{t_\varY}
\newcommand{\transZ}{t_\varZ}
\newcommand{\transXTwo}{\overline{t}_\varX}
\newcommand{\transYTwo}{\overline{t}_\varY}
\newcommand{\transZTwo}{\overline{t}_\varZ}

\newcommand{\predictionX}{\Hat{x}}
\newcommand{\predictionY}{\Hat{y}}
\newcommand{\predictionZ}{\Hat{z}}

\newcommand{\pixU}{u}
\newcommand{\pixV}{v}
\newcommand{\pixUTwo}{u'}
\newcommand{\pixVTwo}{v'}

\newcommand{\ppointU}{u_0} % principal point one
\newcommand{\ppointV}{v_0} % principal point two
\newcommand{\focal}{f}%{\mathfrak{f}}
\newcommand{\focalNormalized}{\bar{\focal}}
\newcommand{\projectionOne}{h} % We do not use g because group member is denoted by g.
\newcommand{\projectionTwo}{h'}
\newcommand{\projectionOneIndexed}{h_i}
\newcommand{\pixUMinusPointU}{(\!\pixU\!-\!\ppointU\!)}

\newcommand{\pixVMinusPointV}{(\!\pixV\!-\!\ppointV\!)}

\newcommand{\transU}{t_u}
\newcommand{\transV}{t_v}
\newcommand{\predictionU}{\Hat{\pixU}}
\newcommand{\predictionV}{\Hat{\pixV}}
\newcommand{\scaleNotation}{s}

\newcommand{\conv}{*}
\newcommand{\filter}{\Psi}

\newcommand{\vanillaBlocks}{vanilla blocks}
\newcommand{\vanillaConv}{vanilla convolution}
\newcommand{\logPolar}{log-polar}
\newcommand{\LogPolar}{Log-polar}

\newcommand{\dla}{DLA-34}
\newcommand{\dlaThirtyFour}{DLA-34}
\newcommand{\dlaOneZeroTwo}{DLA-102}
\newcommand{\dlaOneSixNine}{DLA-169}

\newcommand{\resNetEighteen}{ResNet-18}

\newcommand{\pretrain}{pretrain}

\newcommand{\kitti}{KITTI}
\newcommand{\nuscenes}{nuScenes}
\newcommand{\waymo}{Waymo}

\newcommand{\imageNet}{ImageNet}

\newcommand{\valOne}{Val}

\newcommand{\val}{Val}
\newcommand{\frontal}{frontal}

\newcommand{\loss}{\mathcal{L}}

\newcommand{\lOne}{L_1}

\newcommand{\size}{\text{size}}
\newcommand{\offset}{\text{offset}}
\newcommand{\heatmap}{\text{heatmap}}
\newcommand{\class}{{class}}

\newcommand{\ap}{AP}

\newcommand{\apThreeD}{\ap$_{3\text{D}}$}
\newcommand{\aphThreeD}{APH$_{3\text{D}}$}
\newcommand{\apThreeDForty}{\ap$_{3\text{D}|R_{40}}$}
\newcommand{\apBevForty}{\ap$_{\text{BEV}|R_{40}}$}

\newcommand{\apTwoDForty}{\ap$_{2\text{D}|R_{40}}$}
\newcommand{\levelOne}{Level\_1}
\newcommand{\levelTwo}{Level\_2}
\newcommand{\bracketPercentage}{[\%]}
\newcommand{\imageOnly}{image-only}

\newcommand{\first}[1]{$\textcolor{red}{\mathbf{#1}}$}
\newcommand{\second}[1]{$\textcolor{blue}{\mathbf{#1}}$}
\newcommand{\firstkey}[1]{\textcolor{red}{\textbf{#1}}}
\newcommand{\secondkey}[1]{\textcolor{blue}{\textbf{#1}}}
\newcommand{\sota}{SOTA}
\newcommand{\best}[1]{$\mathbf{#1}$}
\newcommand{\bestKey}[1]{\textbf{#1}}

\newcommand{\mthreeDRPN}{M3D-RPN}

\newcommand{\groomedNMS}{GrooMeD-NMS}
\newcommand{\monoDISMulti}{MonoDIS-M}
\newcommand{\gupNet}{GUP Net}
\newcommand{\pseudoLidar}{Pseudo-{\lidar}}
\newcommand{\ddThreeD}{DD3D}
\newcommand{\ddmp}{DDMP-3D}
\newcommand{\caddn}{CaDDN}
\newcommand{\patchNet}{PatchNet}
\newcommand{\detrThreeD}{DETR3D}

\newcommand{\forExample}{\textit{e.g.}}
\newcommand{\thatIs}{\textit{i.e.}}

\newcommand{\myTopRule}{\Xhline{2\arrayrulewidth}}

\newcolumntype{t}{!{\vrule width 1.5\arrayrulewidth}}
\newcolumntype{m}{!{\vrule width 2.5\arrayrulewidth}}

\colorlet{cyan_highlight}{my_blue!85}

\colorlet{darkGreen_highlight}{darkGreen!75}

\colorlet{my_magenta_highlight}{my_magenta!50}
\newcommand{\COFix}{\cellcolor{my_magenta_highlight}}
\colorlet{my_yellow_highlight}{my_yellow!55}

\providecommand\rightarrowRHD{\relbar\joinrel\mathrel\RHD}

\providecommand\longrightarrowRHD{\relbar\joinrel\relbar\joinrel\mathrel\RHD}

\newcommand{\uparrowRHD}  {\rotatebox[origin=c]{90}{$\rightarrowRHD$}}
\newcommand{\downarrowRHD}{\rotatebox[origin=c]{270}{$\rightarrowRHD$}}
\newcommand{\uparrowRHDSmall}  {\raisebox{0.05\normalbaselineskip}{\scalebox{0.7}{\uparrowRHD}}}   %$\uparrow$
\newcommand{\downarrowRHDSmall}{\raisebox{0.07\normalbaselineskip}{\scalebox{0.7}{\downarrowRHD}}} %$\downarrow$

\usepackage{pifont}% http://ctan.org/pkg/pifont
\newcommand{\cmark}{\checkmark}%\ding{51}}%
\newcommand{\xmark}{\ding{53}}%

\newcommand{\noIndentHeading}[1]{\noindent\textbf{#1}}

\newcommand{\refSupTable}[1]{Tab. \textcolor{red}{#1}}
\newcommand{\refSupFigure}[1]{Fig. \textcolor{red}{#1}}
\newcommand{\refSupApp}[1]{Appendix \textcolor{red}{A#1}}

\definecolor{XLcolor}{rgb}{0.858, 0.188, 0.478}
\usepackage[pagebackref,breaklinks,colorlinks,citecolor={my_blue}]{hyperref}
\usepackage[capitalize]{cleveref}
\crefname{section}{Sec.}{Secs.}
\Crefname{section}{Section}{Sections}
\crefname{table}{Tab.}{Tabs.}
\Crefname{table}{Table}{Tables}

\newcommand{\methodName}{DEVIANT}
\newcommand{\methodNameFull}{Depth EquiVarIAnt NeTwork}
\newcommand{\methodNameShort}{DEVIANT}

\newcommand{\paperTitle}{\methodName: \methodNameFull{} for Monocular 3D Object Detection}

\begin{document}
% \renewcommand\thelinenumber{\color[rgb]{0.2,0.5,0.8}\normalfont\sffamily\scriptsize\arabic{linenumber}\color[rgb]{0,0,0}}
% \renewcommand\makeLineNumber {\hss\thelinenumber\ \hspace{6mm} \rlap{\hskip\textwidth\ \hspace{6.5mm}\thelinenumber}}
% \linenumbers

%============================================================================
% PAPER ID  - PLEASE UPDATE
%============================================================================
\pagestyle{headings}
\mainmatter
\def\ECCVSubNumber{3798}  % Insert your submission number here

%============================================================================
% Title
%============================================================================
\title{\paperTitle} % Replace with your title

\begin{comment}
\titlerunning{ECCV-22 submission ID \ECCVSubNumber} 
\authorrunning{ECCV-22 submission ID \ECCVSubNumber} 
\author{Anonymous ECCV submission}
\institute{Paper ID \ECCVSubNumber}
\end{comment}

\titlerunning{\methodName: \methodNameFull{} for Monocular 3D Detection} 
\authorrunning{A. Kumar et al.} 
\author{
Abhinav Kumar\inst{1} \and
Garrick Brazil\inst{2} \and
Enrique Corona\inst{3} \and
Armin Parchami\inst{3} \and 
Xiaoming Liu\inst{1}
}

\institute{
Michigan State University \and
Meta AI \and
Ford Motor Company\\
\inst{1}\email{\{kumarab6, liuxm\}@msu.edu},~\inst{2}\email{brazilga@fb.com},~\inst{3}\email{\{ecoron18,~mparcham\}@ford.com}\\
\url{https://github.com/abhi1kumar/DEVIANT}
}

% \author{First E. van Author\inst{1}\orcidlink{0000-1111-2222-3333}\index{van Author, First E.} \and
% Second Author\inst{2,3}\orcidlink{1111-2222-3333-4444} \and
% Third Author\inst{3}\orcidlink{2222--3333-4444-5555}}
% %
% \institute{Princeton University, Princeton NJ 08544, USA \and
% Springer Heidelberg, Tiergartenstr. 17, 69121 Heidelberg, Germany
% \email{lncs@springer.com}\\
% \url{http://www.springer.com/gp/computer-science/lncs} \and
% ABC Institute, Rupert-Karls-University Heidelberg, Heidelberg, Germany\\
% \email{\{abc,lncs\}@uni-heidelberg.de}}

\maketitle

%============================================================================
%============================================================================
%============================================================================
\begin{abstract}
% The abstract should summarize the contents of the paper. LNCS guidelines
% indicate it should be at least 70 and at most 150 words. It should be set in 9-point
% font size and should be inset 1.0~cm from the right and left margins.
    Modern neural networks use building blocks such as convolutions that are \equivariant{} to arbitrary \twoD{} translations. 
    However, these \vanillaBlocks{} are not \equivariant{} to arbitrary \threeD{} translations in the projective \manifold. 
    Even then, all monocular \threeD{} detectors use \vanillaBlocks{} to obtain the \threeD{} coordinates, a task for which the \vanillaBlocks{} are not designed for. 
    This paper takes the first step towards convolutions equivariant to arbitrary \threeD{} translations in the projective \manifold.
    Since the depth is the hardest to estimate for monocular detection, this
    paper proposes \methodNameFull{} (\methodName) built with existing \scaleEquivariant{} steerable blocks. 
    As a result, \methodName{} is \equivariant{} to the depth translations in the projective \manifold{} whereas vanilla networks are not.
    The additional \depthEquivariance{} forces the \methodName{} to learn consistent depth estimates, and therefore,
    \methodName{} achieves state-of-the-art monocular \threeD{} detection 
    results on \kitti{} and \waymo{} datasets in the \imageOnly{} category and performs competitively to methods using extra information.
    Moreover, \methodName{} works better than vanilla networks in cross-dataset evaluation. 
    %, \kitti, \waymo}
\end{abstract}

%============================================================================
%============================================================================
%============================================================================
\section{Introduction}\label{sec:intro}

    Monocular \threeD{} object detection is a fundamental task in computer vision, where the task is to infer \threeD{} information including depth from a single monocular image. 
    It has applications in augmented reality \cite{alhaija2018augmented},  gaming \cite{rematas2018soccer}, robotics \cite{saxena2008robotic}, and more recently in autonomous driving \cite{brazil2019m3d, simonelli2020disentangling} as a fallback solution for \lidar{}. 

    Most of the monocular \threeD{} methods attach extra heads to the \twoD{} Faster-RCNN \cite{ren2015faster} or CenterNet \cite{zhou2019objects} for \threeD{} detections.
    Some change architectures \cite{liu2019deep,li2020rtm3d,tang2020center3d} or losses \cite{brazil2019m3d,chen2020monopair}. 
    Others incorporate augmentation \cite{simonelli2020towards}, or confidence \cite{ liu2019deep,brazil2020kinematic}. 
    Recent ones use in-network ensembles \cite{zhang2021objects, lu2021geometry} for better depth estimation.
    
    Most of these methods use \vanillaBlocks{} such as convolutions that are \textit{equivariant} to arbitrary \twoD{} translations \cite{rath2020boosting, bronstein2021convolution}. 
    In other words, whenever we shift the ego camera in \twoD{} (See {\color{vanillaShade}{$\transU$}} of \cref{fig:teaser}), the new image (projection) is a translation of 
    \begin{wraptable}{r}{5.5cm}
        \caption{\textbf{\Equivariance{} comparisons}. [Key: Proj.= Projected, ax= axis]}
        \label{tab:equivariance_comparison}
        \centering
        \scalebox{\scaleFraction}{
            \footnotesize
            \setlength\tabcolsep{0.1cm}
            \begin{tabular}{m l t c c c t c cm}
                \myTopRule
                 & \multicolumn{3}{ct}{\threeD{}} & \multicolumn{2}{cm}{Proj. \twoD{}}\\
                \cline{2-6}
                \textbf{Translation} \scalebox{0.85}{$\rightarrowRHD$}& $x-$ax & $y-$ax & $z-$ax & $u$-ax & $v$-ax\\
                & $(\transX)$ & $(\transY)$ & $(\transZ)$ & $(\transU)$ & $(\transV)$\\
                \myTopRule
                Vanilla CNN & \mathDash{}& \mathDash{}& \mathDash{}& \checkmark & \checkmark \\
                \LogPolar{} \cite{zwicke1983new} & \mathDash{}& \mathDash{}& \checkmark & \mathDash{}& \mathDash{}\\
                \textbf{\methodName} & \mathDash{}& \mathDash{}& \checkmark & \checkmark & \checkmark \\
                Ideal & \checkmark & \checkmark & \checkmark & \mathDash{}& \mathDash{}\\ 
                \myTopRule
            \end{tabular}
        }
    \end{wraptable}
    the original image, and therefore, these methods output a translated feature map. 
    However, in general, the camera moves in depth in driving scenes instead of \twoD{} (See {\color{proposedShade}{$\transZ$}} of \cref{fig:teaser}). 
    So, the new image is not a translation of the original input image due to the projective transform.
    Thus, using \vanillaBlocks{} in monocular methods is a mismatch between the assumptions and the regime where these blocks operate.
    Additionally, there is a huge generalization gap between training and validation for monocular \threeD{} detection (See %\cref{tab:kitti_compare_generalization_big}
    \refSupTable{14} in the supplementary).
    Modeling translation \equivariance{} in the correct \manifold{} improves generalization for tasks in spherical \cite{cohen2018spherical} and hyperbolic \cite{ganea2018hyperbolic} manifolds. 
    Monocular detection involves processing pixels (\threeD{} point projections) to obtain the \threeD{} information, and is thus a task in the projective \manifold.
    Moreover, the depth in monocular detection is ill-defined \cite{tang2020center3d}, and thus, the hardest to estimate \cite{ma2021delving}.
    Hence, using building blocks \textit{\equivariant{} to depth translations in the projective \manifold} is a natural choice for improving generalization and is also at the core of this work (See \refSupApp{1.8}).%cref{sec:why_better_generalize}).
    
    \begin{figure}[!tb]
        \centering
        \begin{subfigure}[align=bottom]{.35\linewidth}
            \centering
            \input{images/teaser.tex}
            \caption{Idea.}
        \end{subfigure}
        \hfill
        \begin{subfigure}[align=bottom]{.63\linewidth}
              \centering
              \input{images/depth_equivariance}
              \caption{Depth \Equivariance{}.}
        \end{subfigure}
        \caption{
        \textbf{(a) Idea.} Vanilla CNN is \equivariant{} to {\color{vanillaShade}{projected \twoD{} translations $\transU,\transV$}} of the ego camera. 
        The ego camera moves in \threeD{} in driving scenes which breaks this assumption. 
        We propose \methodName{} which is additionally \equivariant{} to {\color{proposedShade}{depth translations $\transZ$}} in the projective \manifold.
        \textbf{(b) Depth \Equivariance{}}. 
        \methodName{} enforces additional consistency among the feature maps of an image and its transformation caused by the ego depth translation.
        $\transformationMath_\scaleNotation\!=\!$ scale transformation, $*\!=\!$ vanilla convolution.
        }
        \label{fig:teaser}
    \end{figure}

    Recent monocular methods use flips \cite{brazil2019m3d}, scale \cite{simonelli2020towards, lu2021geometry}, mosaic \cite{bochkovskiy2020yolov4, sugirtha2021exploring} or copy-paste \cite{lian2021geometry} augmentation, depth-aware convolution \cite{brazil2019m3d}, or geometry \cite{liu2021ground, lu2021geometry, shi2021geometry, zhang2021learning} to improve generalization. 
    Although all these methods improve performance, a major issue is that their backbones are not designed for the projective world. 
    This results in the depth estimation going haywire with a slight ego movement \cite{zhou2021monoef}. 
    Moreover, data augmentation, \forExample, flips, scales, mosaic, copy-paste, is not only limited for the projective tasks, but also does not guarantee desired behavior \cite{gandikota2021training}.
        
    To address the mismatch between assumptions and the operating regime of the vanilla blocks and improve generalization, we take the first step towards convolutions \equivariant{} to arbitrary \threeD{} translations in the projective manifold. 
    We propose \methodNameFull{} (\methodName)
    which is additionally equivariant to depth translations in the projective \manifold{} as shown in \cref{tab:equivariance_comparison}.
    Building upon the classic result from \cite{hartley2003multiple}, we simplify it under reasonable assumptions about the camera movement in autonomous driving to get scale transformations.
    The \scaleEquivariant{} blocks are well-known in the literature \cite{ghosh2019scale, zhu2019scale, sosnovik2020sesn, jansson2021scale}, and consequently, we replace the vanilla blocks in the backbone with their \scaleEquivariant{} steerable counterparts \cite{sosnovik2020sesn} to additionally embed \equivariance{} to depth translations in the projective \manifold. 
    %Thus, \methodNameShort{} backbone is explicitly ``aware'' of the depth translations in the projective world.
    Hence, \methodName{} learns consistent depth estimates and improves monocular detection.

    In summary, the main contributions of this work include:
    \begin{itemize}
        \item We study the modeling error in monocular \threeD{} detection  and propose \depthEquivariant{} networks built with \scaleEquivariant{} steerable blocks as a solution.
        \item We achieve state-of-the-art (\sota) monocular \threeD{} object detection results on the \kitti{} and \waymo{} datasets in the \imageOnly{} category and perform competitively to methods which use extra information.
        \item  We experimentally show that \methodName{} works better in cross-dataset evaluation suggesting better generalization than vanilla CNN backbones.
    \end{itemize}

%============================================================================
%============================================================================
%============================================================================
\section{Literature Review}

    %============================================================================
    \noIndentHeading{\Equivariant{} Neural Networks.}
        The success of convolutions in CNN has led people to look for their generalizations \cite{cohen2016group,weiler2021coordinate}.
        %(Circular) 
        Convolution is the unique solution to \twoD{} translation \equivariance{} in the Euclidean \manifold{} \cite{bronstein2021convolution, bronstein2021geometric, rath2020boosting}.
        Thus, convolution in CNN is a prior in the Euclidean \manifold.
        Several works explore other group actions in the Euclidean \manifold{} such as \twoD{} rotations \cite{cohen2014learning, dieleman2016exploiting, marcos2017rotation, weiler2018learning}, scale \cite{kanazawa2014locally, marcos2018scale}, flips \cite{yeh2019chirality}, or their combinations \cite{worrall2017harmonic, wang2021incorporating}. 
        Some consider \threeD{} translations \cite{worrall2018cubenet} and rotations \cite{thomas2018tensor}. % on the \threeD{} manifold.
        Few \cite{dosovitskiy2021image, wilk2018learning, zhou2020meta} attempt learning the equivariance from the data, but such methods have significantly higher data requirements \cite{worrall2018cubenet}.
        Others change the \manifold{} to spherical \cite{cohen2018spherical}, hyperbolic \cite{ganea2018hyperbolic}, graphs \cite{micheli2009neural}, or arbitrary manifolds \cite{jing2020physical}. 
        Monocular \threeD{} detection involves operations on pixels which are projections of \threeD{} point and thus, works in a different \manifold{} namely projective \manifold.
        \cref{tab:equivariance} summarizes all these \equivariance{}s known thus far.
        \begin{table}[!tb]
            \caption{\Equivariance s known in the literature.}
            \label{tab:equivariance}
            \centering
            \scalebox{\scaleFraction}{
            \footnotesize
            \setlength\tabcolsep{0.1cm}
            \begin{tabular}{m l t c|c|c|c|cm}
                \myTopRule
                \textbf{Transformation} \scalebox{0.85}{$\rightarrowRHD$} & \multirow{2}{*}{Translation} & \multirow{2}{*}{Rotation} & \multirow{2}{*}{Scale} & 
                \multirow{2}{*}{Flips} & \multirow{2}{*}{Learned}\\
                \cline{1-1}
                \textbf{Manifold} {\raisebox{0.1\normalbaselineskip}{\scalebox{0.85}{\downarrowRHD}}}
                &  &  & & &\\
                \myTopRule
                \multirow{2}{*}{Euclidean} & \multirow{2}{*}{Vanilla CNN\cite{lecun1998gradient}} & Polar, & \LogPolar\cite{henriques2017warped}, & \multirow{2}{*}{ChiralNets\cite{yeh2019chirality}} & \multirow{2}{*}{Transformers\cite{dosovitskiy2021image}}\\
                & & Steerable\cite{worrall2017harmonic} &  Steerable\cite{ghosh2019scale} & & \\
                \hline
                Spherical & Spherical CNN\cite{cohen2018spherical} & \mathDash{}& \mathDash{}& \mathDash{}& \mathDash\\
                \hline
                Hyperbolic & Hyperbolic CNN\cite{ganea2018hyperbolic} & \mathDash{}& \mathDash{}& \mathDash{}& \mathDash\\
                \hline
                Projective & Monocular Detector & \mathDash{}& \mathDash{}& \mathDash{}& \mathDash\\
                \myTopRule
            \end{tabular}
            }
        \end{table}

    %============================================================================
    \noIndentHeading{\ScaleEquivariant{} Networks.}
        %\ccolorbox{my_yellow_highlight}{
        Scale \equivariance{} in the Euclidean \manifold{} is more challenging than the rotations because of its acyclic and unbounded nature \cite{rath2020boosting}.
        There are two major lines of work for \scaleEquivariant{} networks. 
        The first \cite{henriques2017warped, esteves2018polar} infers the global scale using \logPolar{} transform \cite{zwicke1983new},
        while the other infers the scale locally by convolving with multiple scales of images \cite{kanazawa2014locally} or filters \cite{xu2014scale}. 
        Several works \cite{ghosh2019scale, zhu2019scale, sosnovik2020sesn, jansson2021scale} extend the local idea, using steerable filters \cite{freeman1991design}.     
        Another work \cite{worrall2019deep} constructs filters for integer scaling.
        We compare the two kinds of \scaleEquivariant{} convolutions on the monocular \threeD{} detection task
        and show that steerable convolutions are better suited to embed depth (scale) \equivariance.
        Scale equivariant networks have been used for classification \cite{esteves2018polar, ghosh2019scale, sosnovik2020sesn}, \twoD{} tracking \cite{sosnovik2021siamese} and \threeD{} object classification \cite{esteves2018polar}.
        We are the first to use \scaleEquivariant{} networks for monocular \threeD{} detection.

    %============================================================================
    \noIndentHeading{3D Object Detection.}
        Accurate \threeD{} object detection uses sparse data from \lidar{}s \cite{shi2019pointrcnn}, which are expensive and do not work well in severe weather \cite{tang2020center3d} and glassy environments. 
        Hence, several works have been on monocular camera-based \threeD{} object detection, which is simplistic  but has scale/depth ambiguity \cite{tang2020center3d}.
        Earlier approaches \cite{payet2011contours, fidler20123d, pepik2015multi, chen2016monocular} use hand-crafted features, while the recent ones use deep learning.
        Some change architectures \cite{liu2019deep,li2020rtm3d,tang2020center3d,liu2022learning} or losses \cite{brazil2019m3d,chen2020monopair}. 
        Some use scale \cite{simonelli2020towards, lu2021geometry}, mosaic \cite{sugirtha2021exploring} or copy-paste \cite{lian2021geometry} augmentation.
        Others incorporate depth in convolution \cite{brazil2019m3d, ding2020learning}, or confidence \cite{kumar2020luvli, liu2019deep,brazil2020kinematic}. 
        More recent ones use in-network ensembles to predict the depth deterministically \cite{zhang2021objects} or probabilistically \cite{lu2021geometry}.
        A few use temporal cues \cite{brazil2020kinematic}, NMS \cite{kumar2021groomed}, or corrected camera extrinsics \cite{zhou2021monoef} in the training pipeline.
        Some also use CAD models \cite{chabot2017deep, liu2021autoshape} or \lidar{} \cite{reading2021categorical} in training.
        Another line of work called \pseudoLidar{} \cite{wang2019pseudo, ma2019accurate,ma2020rethinking,simonelli2021we,park2021pseudo} estimates the depth first, and then uses a point cloud-based \threeD{} object detector.
        We refer to \cite{ma20223d} for a detailed survey.
        Our work is the first to use \scaleEquivariant{} 
        blocks in the backbone for monocular \threeD{} detection.

%============================================================================
%============================================================================
%============================================================================
\section{Background}
    We first provide the necessary definitions which are used throughout this paper. These are not our contributions and can be found in the literature \cite{worrall2018cubenet, burns1992non, hartley2003multiple}.
    
    %============================================================================
    \noIndentHeading{\Equivariance.} 
        Consider a group of transformations $\transformationGroup$, whose individual members are $\transformationGroupMember$. 
        Assume $\mapping$ denote the mapping of the inputs $\inputData$ to the outputs $\outputData$. 
        Let the inputs and outputs undergo the transformation $\transformationInput$ and $\transformationOutput$ respectively. 
        Then, the mapping $\mapping$ is equivariant to the group $\transformationGroup$\cite{worrall2018cubenet} if
            $\mapping (\transformationInput \inputData) = \transformationOutput (\mapping \inputData), \forall~\transformationGroupMember\in\transformationGroup.$
        Thus, \equivariance{} provides an explicit relationship~between input transformations and feature-space transformations at each layer of the neural network \cite{worrall2018cubenet}, and intuitively makes the learning easier. 
        The mapping $\mapping$ is the \vanillaConv~when the $\transformationInput= \transformationOutput= \transformationMath_\translation$ where $\transformationMath_\translation$ denotes the translation $\translation$ on the discrete grid \cite{bronstein2021convolution, bronstein2021geometric, rath2020boosting}.
        These \vanillaConv~introduce weight-tying \cite{lecun1998gradient} in fully connected neural networks resulting in a greater generalization.
        A special case of \equivariance{} is the invariance \cite{worrall2018cubenet} which is given by $\mapping (\transformationInput \inputData) = \mapping \inputData, \forall~\transformationGroupMember\in\transformationGroup.$

    %============================================================================
    \noIndentHeading{Projective Transformations.} 
        Our idea is to use \equivariance{} to depth translations in the projective \manifold{} since the monocular detection task belongs to this manifold.
        A natural question to ask is whether such \equivariant{}s exist in the projective manifold.
        \cite{burns1992non} answers this question in negative, and says that such \equivariant{}s do not exist in general. 
        However, such equivariants exist for special classes, such as planes.
        An intuitive way to understand this is  to infer the rotations and translations by looking at the two projections (images). 
        For example, the result of \cite{burns1992non} makes sense if we consider a car with very different front and back sides as in
        \refSupFigure{6}.
        % \cref{fig:non_existence}. 
        A $180\degree$ ego rotation around the car means the projections (images) are its front and the back sides, which are different. 
        Thus, we can not infer the translations and rotations 
        from these two projections.
        Based on this result, we stick with \textbf{locally} planar objects \thatIs{} we assume that a \threeD{} object is made of several \textit{patch planes}. 
        (See last row of \cref{fig:receptive} as an example). 
        It is important to stress that we do \textbf{NOT} assume that the \threeD{} object such as car is planar. 
        The local planarity also agrees with the property of manifolds that manifolds locally resemble $n$-dimensional Euclidean space and because the projective transform maps planes to planes, the patch planes in \threeD{} are also locally planar. 
        We show a sample planar patch and the \threeD{} object in \refSupFigure{5} in the appendix.%\cref{fig:eqv_exists}.

    %============================================================================
    \noIndentHeading{Planarity and Projective Transformation.}                              
        Example 13.2 from \cite{hartley2003multiple} links the planarity and projective transformations.
        Although their result is for stereo with two different cameras $(\projectionOperator, \projectionOperator')$, we substitute $\projectionOperator\!=\!\projectionOperator'$ to get \cref{th:projective_bigboss}. 
        \begin{theorem}\label{th:projective_bigboss}\cite{hartley2003multiple}
            Consider a \threeD{} point lying on a \plane~$mx\!+\!ny\!+\!oz\!+\!p\!=\!0$, and observed by an ego camera in a pinhole setup to give an image $\projectionOne$. 
            Let $\translation\!=\!(\transX,\transY,\transZ)$ and $\rotation\!=\![r_{ij}]_{3\times3}$ denote a translation and rotation of the ego camera respectively. 
            Observing the same \threeD{} point from a new camera position leads to an image $\projectionTwo$. 
            Then, the image $\projectionOne$ is related to the image $\projectionTwo$ by the projective \transformation{} 
            \begin{align}
                &\transformationMath: \projectionOne(\pixU-\ppointU, \pixV-\ppointV) =  
                \label{eq:bigboss}\\
                &\!\projectionTwo\!\left(\!\focal\!\dfrac
                {\left(\!r_{11}\mySign\transXTwo\!\frac{m}{p}\right)\!\pixUMinusPointU
                \!+\!\left(\!r_{21}\mySign\transXTwo\!\frac{n}{p}\right)\!\pixVMinusPointV
                \!+\!\left(\!r_{31}\mySign\transXTwo\!\frac{o}{p}\right)\!\focal
                }
                {\left(\!r_{13}\mySign\transZTwo\frac{m}{p}\right)\!\pixUMinusPointU
                \!+\!\left(\!r_{23}\mySign\transZTwo\!\frac{n}{p}\right)\!\pixVMinusPointV
                \!+\!\left(\!r_{33}\mySign\transZTwo\!\frac{o}{p}\right)\!\focal
                },
                \right.\nonumber \\
                &\!\left.\!\focal\!\dfrac
                {\left(\!r_{12}\mySign\transYTwo\!\frac{m}{p}\right)\!\pixUMinusPointU 
                \!+\!\left(\!r_{22}\mySign\transYTwo\!\frac{n}{p}\right)\!\pixVMinusPointV  
                \!+\!\left(\!r_{32}\mySign\transYTwo\!\frac{o}{p}\right)\!\focal
                }
                {\left(\!r_{13}\mySign\transZTwo\!\frac{m}{p}\right)\!\pixUMinusPointU
                \!+\!\left(\!r_{23}\mySign\transZTwo\!\frac{n}{p}\right)\!\pixVMinusPointV
                \!+\!\left(\!r_{33}\mySign\transZTwo\!\frac{o}{p}\right)\!\focal
                } 
                \!\right), \nonumber
            \end{align}
             where $\focal$ and $(\ppointU, \ppointV)$ denote the 
            focal length and principal point of the ego camera, and  $(\transXTwo, \transYTwo, \transZTwo) = \rotation^T\translation$. 
        \end{theorem}

%============================================================================
%============================================================================
%============================================================================
\section{\DepthEquivariant{} Backbone}\label{sec:proposed}
    The projective transformation in \cref{eq:bigboss} from \cite{hartley2003multiple} is complicated and also involves rotations, and we do not know which convolution obeys this projective \transformation. 
    Hence, we simplify \cref{eq:bigboss} under reasonable assumptions to obtain a familiar \transformation{} for which the \textit{convolution} is known.
        \begin{corollary}\label{th:projective_scaled}
            % Consider a \threeD{} point lying on a \plane~$mx\!+\!ny\!+\!oz\!+\!p\!=\!0$, and observed by an ego camera in a pinhole setup to give an image $\projectionOne$. 
            % Observing the same \threeD{} point from the rotation and translation of the ego camera leads to an image $\projectionTwo$. 
            When the ego camera translates in depth without rotations 
            $(\rotation=\identity)$,
            and the \plane~is ``approximately'' parallel to the image plane, the image $\projectionOne$ locally is a scaled version of the second image $\projectionTwo$ independent of focal length, \thatIs
            \begin{align}
                &\transformationMath_\scaleNotation: \projectionOne(\pixU-\ppointU, v-\ppointV)
                \approx \projectionTwo\left(\dfrac{\pixU-\ppointU}{1\mySign\transZ\frac{o}{p}}, \dfrac{v-\ppointV}{1\mySign\transZ\frac{o}{p}}\right).
                \label{eq:projective_scaled}
            \end{align}
            where $\focal$ and $(\ppointU, \ppointV)$ denote the 
            focal length and principal point of the ego camera, and $\transZ$ denotes the ego translation.
        \end{corollary}
        % We refer to \cref{sec:approximation_proof} in the supplementary material for the detailed explanation of \cref{th:projective_scaled}.
        See \refSupApp{1.6} for the detailed explanation of \cref{th:projective_scaled}.
        \cref{th:projective_scaled} says
        \begin{align}
            &\transformationMath_\scaleNotation : \projectionOne(\pixU-\ppointU, \pixV-\ppointV)
            \approx \projectionTwo\left(\frac{\pixU-\ppointU}{s}, \frac{\pixV-\ppointV}{s}\right),
        \end{align}
        where, $s\!=\!1+\transZ\frac{o}{p}$ denotes the scale and $\transformationMath_\scaleNotation$ denotes the scale \transformation. 
        The scale $s\! <\! 1$ suggests downscaling, while $s \!>\! 1$ suggests upscaling.
        \cref{th:projective_scaled} shows that the \transformation{} $\transformationMath_\scaleNotation$  is independent of the focal length and that scale is a linear function of the depth translation.
        Hence, the depth translation in the projective \manifold{} induces scale \transformation{} and thus, the depth \equivariance{} in the projective manifold is the \scaleEquivariance{} in the Euclidean manifold. 
        Mathematically, the desired \equivariance{} is $\left[\transformationMath_{\scaleNotation}(\projectionOne)\conv\filter\right] = \transformationMath_{\scaleNotation}\left[\projectionOne \conv \filter_{\scaleNotation^{-1}}\right]$, where $\filter$ denotes the filter (See \refSupApp{1.7}). %\cref{sec:scale_eqv_proof}).
        As CNN is not a \scaleEquivariant{} (\se) architecture \cite{sosnovik2020sesn}, 
        we aim to get \se{} backbone which makes the architecture \equivariant{} to depth translations in the projective manifold.
        The scale transformation is a familiar \transformation{} and \se{} convolutions are well known \cite{ghosh2019scale, zhu2019scale, sosnovik2020sesn, jansson2021scale}.

        \begin{figure}[!tb]
            \centering
            \begin{subfigure}[align=bottom]{.31\linewidth}
                \centering
                \includegraphics[width=\linewidth]{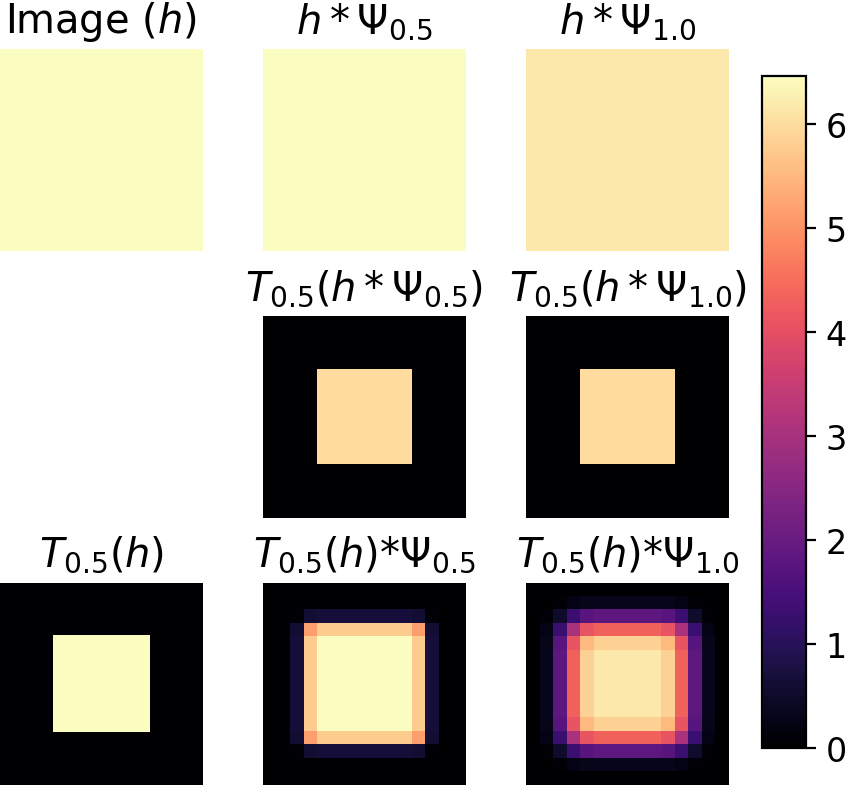}
                \caption{\ses{} Convolution Output.}
                \label{fig:scale_eq_toy}
            \end{subfigure}\hfill%
            \begin{subfigure}[align=bottom]{.27\linewidth}
                \centering
                \includegraphics[width=\linewidth]{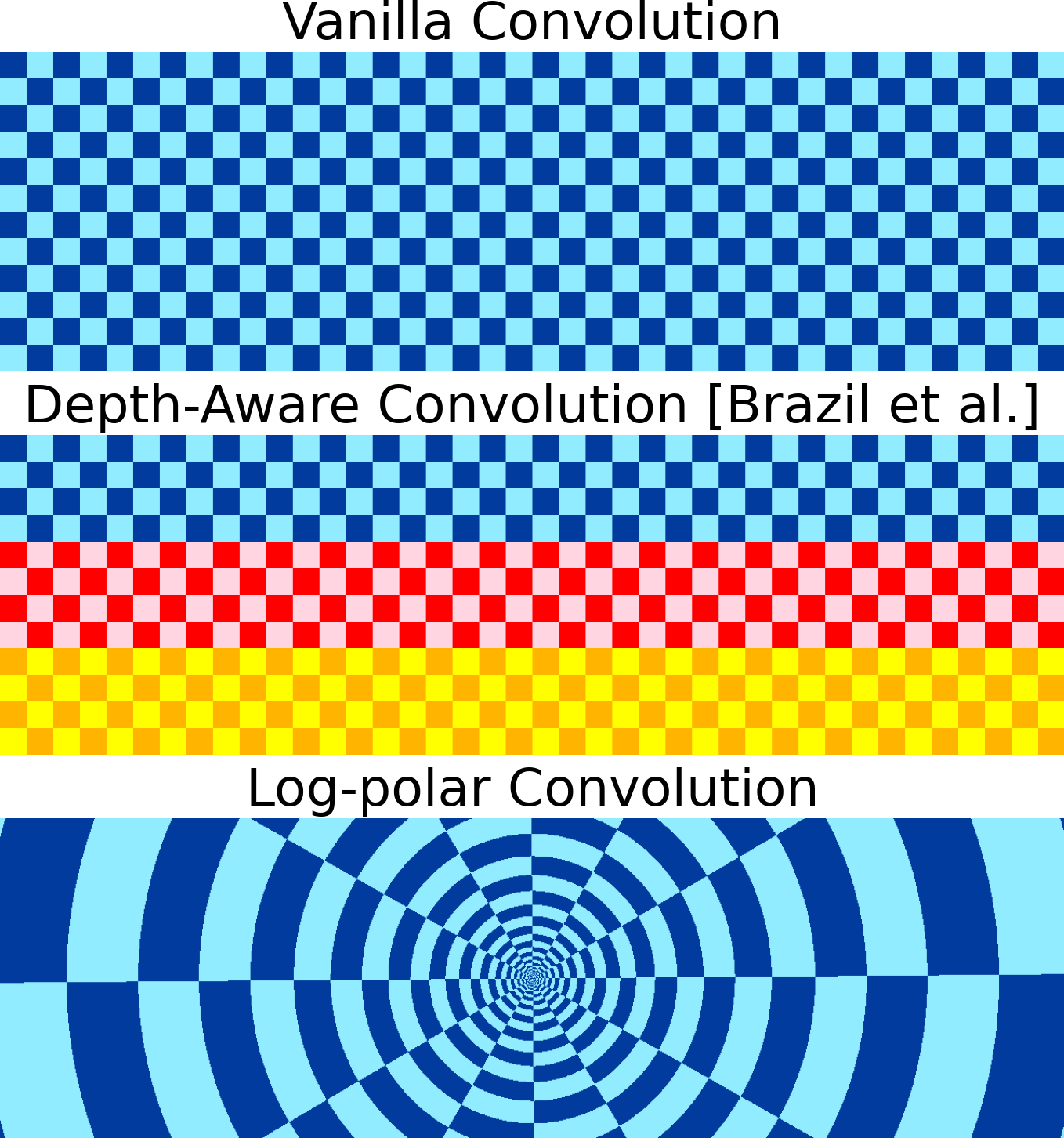}
                \caption{Receptive fields.}
                \label{fig:receptive}
            \end{subfigure}\hfill%
            \begin{subfigure}[align=bottom]{.38\linewidth}
                \centering
                \includegraphics[width=\linewidth]{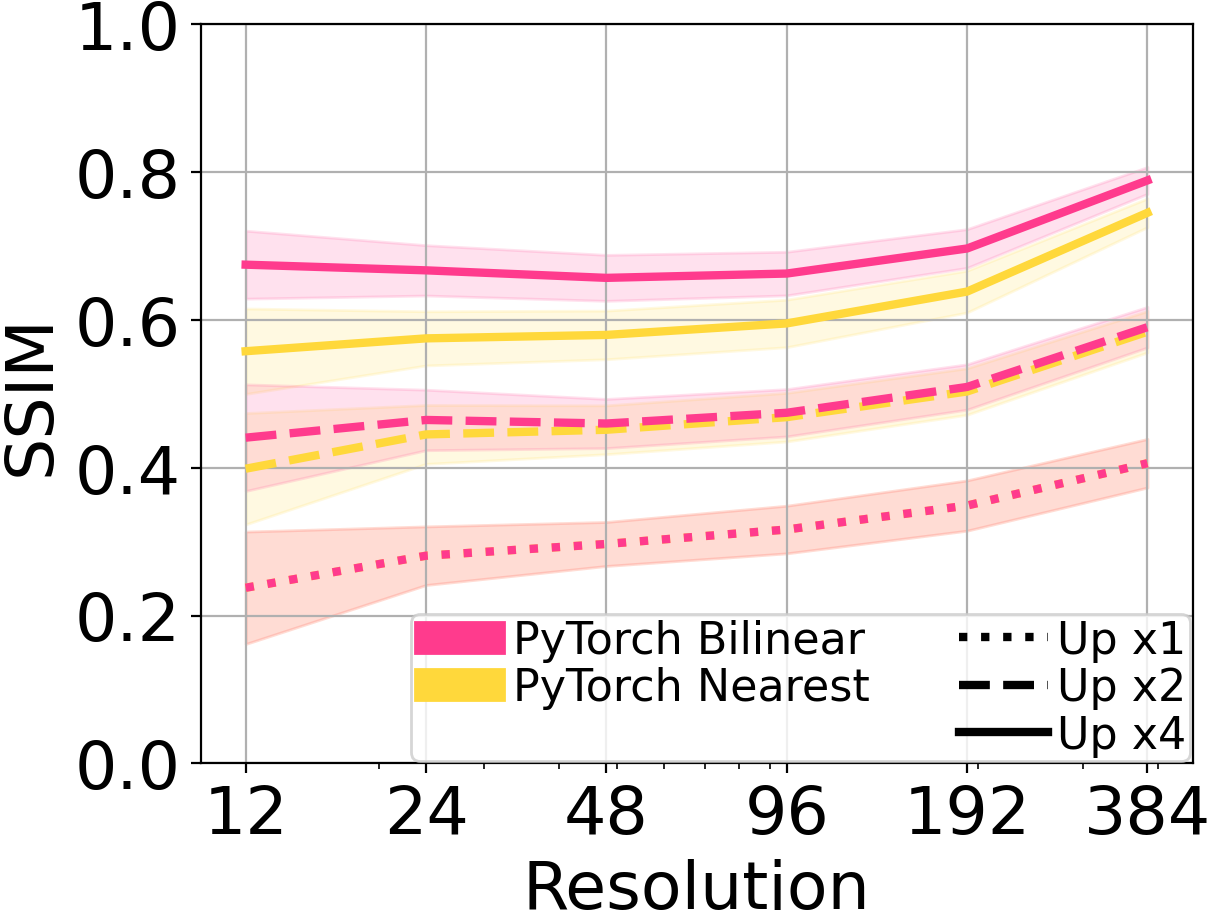}
                \caption{\LogPolar{} SSIM.}
                \label{fig:log_polar_ssim}
            \end{subfigure}
            \caption{ 
            \textbf{(a) \ScaleEquivariance}. We apply \ses{} convolution \cite{sosnovik2020sesn} with two scales on a single channel toy image $\projectionOne$. 
            \textbf{(b) Receptive fields} of convolutions in the Euclidean manifold. Colors represent different weights, while shades represent the same weight. 
            \textbf{(c) Impact of discretization on \logPolar{} convolution.} SSIM is very low at small resolutions and is not $1$ even after upscaling by $4$. 
            [Key: Up= Upscaling]
            }
        \end{figure}

        %============================================================================
        \noIndentHeading{\ScaleEquivariant{} Steerable (\ses) Blocks.}\label{sec:steerable}    
            We use the existing \ses{} blocks \cite{sosnovik2020sesn,sosnovik2021siamese} to construct our \methodNameFull{} (\methodName{}) backbone.
            As \cite{sosnovik2021siamese} does not construct \se-\dla{} backbones, we construct our \methodName{} backbone as follows.
            We replace the vanilla convolutions by the \ses{} convolutions \cite{sosnovik2021siamese} with the basis as Hermite polynomials.
            \ses{} convolutions result in multi-scale representation of an input tensor. 
            As a result, their output is five-dimensional instead of four-dimensional. 
            Thus, we replace the \twoD{} pools and batch norm (BN) by \threeD{} pools and \threeD{} BN respectively.
            The \MaxScale{} layer \cite{sosnovik2020sesn} carries a $\max$ over the extra (scale) dimension to project five-dimensional tensors to four dimensions (See \refSupFigure{9} in the supplementary). %\cref{fig:steerable_idea}).
            Ablation in \cref{sec:results_ablation} confirms that BN and Pool (BNP) should also be \se{} for the best performance.
            
            The \ses{} convolutions \cite{ghosh2019scale, zhu2019scale, sosnovik2020sesn} are based on steerable-filters \cite{freeman1991design}.
            Steerable approaches \cite{ghosh2019scale} first pre-calculate the non-trainable multi-scale basis in the Euclidean \manifold{} and then build filters by the linear combinations of the trainable weights $\weight$ (See \refSupFigure{9}).
            The number of trainable weights $\weight$ equals the number of filters at one particular scale. 
            The linear combination of multi-scale basis ensures that the filters are also multi-scale.
            Thus, \ses{} blocks bypass grid conversion and do not suffer from sampling effects.
            
            We show the convolution of toy image $\projectionOne$ with a \ses{} convolution in \cref{fig:scale_eq_toy}. Let $\filter_\scaleNotation$ denote the filter at scale $\scaleNotation$.
            The convolution between downscaled image and filter $\transformationMath_{0.5}(\projectionOne)\conv\filter_{0.5}$ matches the downscaled version of original image convolved with upscaled filter $\transformationMath_{0.5}(\projectionOne\conv\filter_{1.0})$. 
            \cref{fig:scale_eq_toy} (right column) shows that the output of a CNN  exhibits aliasing in general and is therefore, not \scaleEquivariant.

        %============================================================================
        \noIndentHeading{\LogPolar{} Convolution: Impact of Discretization.}\label{sec:log_polar}
            An alternate way to convert the depth translation $\transZ$ of \cref{eq:projective_scaled} to shift is by converting the images to \logPolar{} space \cite{zwicke1983new} around the principal point $(\ppointU,\ppointV)$, as
            \begin{align}
                \projectionOne(\ln r, \theta)  &\approx \projectionTwo
                \left(
                \ln r - \ln\left(1\mySign\transZ\frac{o}{p}\right),~~\theta \right), 
                \label{eq:log_polar}
            \end{align}
            with $r\!=\!\sqrt{(\pixU\!-\!\ppointU)^2\!+\!(\pixV\!-\ppointV)^2}$, and $\theta\!=\! \tan^{-1}\left(\frac{\pixV-\ppointV}{\pixU-\ppointU}\right)$.
            The \logPolar{} transformation converts the scale to translation, so using convolution in the \logPolar{} space is \equivariant{} to the logarithm of the depth translation $\transZ$.
            We show the receptive field of \logPolar{} convolution in \cref{fig:receptive}.
            The \logPolar{} convolution uses a smaller receptive field for objects closer to the principal point, while a larger field away from the principal point.
            We implemented \logPolar{} convolution and found that its performance (See \cref{tab:ablation}) is not acceptable, consistent with \cite{sosnovik2020sesn}.
            We attribute this behavior to the discretization of pixels and loss of \twoD{} translation \equivariance.
            \cref{eq:log_polar} is perfectly valid in the continuous world (Note the use of parentheses instead of square brackets in \cref{eq:log_polar}). 
            However, pixels reside on discrete grids, which gives rise to sampling errors \cite{kumar2013estimation}.
            We discuss the impact of discretization on \logPolar{} convolution in \cref{sec:detection_results_kitti_val1} and show it in \cref{fig:log_polar_ssim}. 
            Hence, we do not use \logPolar{} convolution for the \methodName{} backbone.

        %============================================================================
        \noIndentHeading{Comparison of \Equivariance s for Monocular 3D Detection.}
            We now compare \equivariance{}s for monocular \threeD{} detection task.
            An ideal monocular detector should be \equivariant{} to arbitrary \threeD{} translations $(\transX, \transY, \transZ)$.
            However, most monocular detectors \cite{kumar2021groomed, lu2021geometry}  estimate \twoD{} projections of \threeD{} centers and the depth, which they back-project in \threeD{} world via known camera intrinsics. 
            Thus, a good enough detector shall be \equivariant{} to \twoD{} translations $(\transU,\transV)$ for projected centers as well as \equivariant{} to depth translations $(\transZ)$. 
            
            Existing detector backbones \cite{kumar2021groomed, lu2021geometry} are only \equivariant{} to \twoD{} translations as they use vanilla convolutions that produce \fourD{} feature maps. 
            Log-polar backbones is \equivariant{} to logarithm of depth translations but not 
            to \twoD{} translations.
            \methodName{} uses \ses{} convolutions to produce \fiveD{} feature maps. 
            The extra dimension in \fiveD{} feature map~captures the changes in scale (for depth), while these feature maps individually are \equivariant{} to \twoD{} translations (for projected centers).
            Hence, \methodName{} augments the \twoD{} translation \equivariance{} $(\transU,\transV)$ of the projected  centers with the depth translation \equivariance.
            We emphasize that although \methodName{} is \textbf{not} \equivariant{} to arbitrary \threeD{} translations in the projective manifold, \methodName{} \textbf{does} provide the \equivariance{} to depth translations $(\transZ)$ and is thus a first step towards the ideal \equivariance.
            Our experiments (\cref{sec:experiments}) show that even this additional \equivariance{} benefits monocular \threeD{} detection task.
            This is expected because depth is the hardest parameter to estimate \cite{ma2021delving}.
            \cref{tab:equivariance_comparison} summarizes these \equivariance{}s.
            Moreover, \cref{tab:results_kitti_compare_2d_3d} empirically shows that \twoD{} detection does not suffer and therefore, confirms that \methodName{} indeed augments the \twoD{} \equivariance{} with the \depthEquivariance.
            An idea similar to \methodName{} is the optical expansion \cite{yang2020upgrading} which augments optical flow with the scale information and benefits depth estimation.

%============================================================================
%============================================================================
%============================================================================
\section{Experiments}\label{sec:experiments}

        Our experiments use the \kitti{} \cite{geiger2012we}, \waymo{} \cite{sun2020scalability} and \nuscenes{} datasets \cite{caesar2020nuscenes}.
        We modify the publicly-available PyTorch \cite{paszke2019pytorch} code of \gupNet{} \cite{lu2021geometry} and use the \gupNet{} model as our baseline.
        For \methodName, we keep the number of scales as three \cite{sosnovik2021siamese}. 
        \methodName{} takes $8.5$ hours for training and $0.04$s per image for inference on a single A100 GPU.
        See \refSupApp{2.2} for more details.

        %============================================================================
        \noIndentHeading{Evaluation Metrics.} 
            \kitti{} evaluates on three object categories: Easy, Moderate and Hard. 
            It assigns each object to a category based on its occlusion, truncation, and height in the image space. 
            \kitti{} uses \apThreeDForty{} percentage metric on the Moderate category to benchmark models \cite{geiger2012we} following \cite{simonelli2019disentangling, simonelli2020disentangling}.
            %The \apThreeDForty{} performance on the Moderate category compares different models in the benchmark. 
            
            \waymo{} evaluates on two object levels: \levelOne~and \levelTwo. 
            It assigns each object to a level based on the number of \lidar{} points included in its \threeD{} box. 
            %An object belongs to \levelOne~if its \threeD{} box contains more than $5$ \lidar{} points else it is assigned \levelTwo. 
            \waymo{} uses \aphThreeD~percentage metric which is the incorporation of heading information in \apThreeD{} to benchmark models.
            It also provides evaluation at three distances $[0,30)$, $[30,50)$ and $[50, \infty)$ meters.

        %============================================================================
        \noIndentHeading{Data Splits.}
            We use the following splits of the \kitti,\waymo{} and \nuscenes: %~datasets:
            \begin{itemize}
                \item \textit{\kitti{} Test (Full) split}: Official \kitti{} \threeD{} benchmark \cite{kitti2012benchmark}
            consists of $7{,}481$ training and $7{,}518$ testing images \cite{geiger2012we}.
        
                \item \textit{\kitti{} \valOne{} split}: It partitions the $7{,}481$ training images into $3{,}712$ training and $3{,}769$ validation images \cite{chen20153d}.
                
                \item \textit{\waymo{} \val{} split}: This~split \cite{reading2021categorical, wang2021progressive} contains $52{,}386$ training and $39{,}848$ validation images from the front camera. 
                We construct its training set by sampling every third frame from the training sequences as in \cite{reading2021categorical, wang2021progressive}.

                \item \textit{\nuscenes{} \val{} split:} It consists of $28{,}130$ training and $6{,}019$ validation images from the front camera \cite{caesar2020nuscenes}. We use this split for evaluation \cite{shi2021geometry}.
            \end{itemize}

        \begin{table}[!tb]
            \caption{\textbf{Results on \kitti{} Test cars} 
            at \iouThreeD{} $\geq\!0.7$. Previous results are from the leader-board or papers.
            We show $3$ methods in each Extra category and $6$ methods in the \imageOnly{} category. [Key: \firstkey{Best}, \secondkey{Second Best}]
            }
            \label{tab:detection_results_kitti_test_cars}
            \centering
            \scalebox{\scaleFraction}{
                \footnotesize
                \setlength\tabcolsep{0.1cm}
                \rowcolors{3}{lightgray}{white}
                \begin{tabular}{ml m c m ccc  m cccm}
                    \myTopRule
                    \addlinespace[0.01cm]
                    \multirow{2}{*}{Method} & \multirow{2}{*}{Extra} &\multicolumn{3}{cm}{\apThreeDForty \bracketPercentage (\uparrowRHDSmall)} & \multicolumn{3}{cm}{\apBevForty \bracketPercentage (\uparrowRHDSmall)}\\ 
                    & & Easy & Mod & Hard & Easy & Mod & Hard\\ 
                    \myTopRule
                    %FQNet \cite{liu2019deep}                  & $??$  &  $2.77$        &  $1.51$        & $1.01$        &  $5.40$        &  $3.23$        &  $2.46$       \\
                    %ROI-10D \cite{manhardt2019roi}            & $??$  &  $4.32$        &  $2.02$        & $1.46$        &  $9.78$        &  $4.91$        &  $3.74$       \\
                    %GS3D \cite{li2019gs3d}                    & $??$  &  $4.47$        &  $2.90$        & $2.47$        &  $8.41$        &  $6.08$        &  $4.94$       \\
                    AutoShape \cite{liu2021autoshape}         & CAD   &  $22.47$ & $14.17$ & $11.36$       &  $30.66$ & $20.08$    & $15.59$       \\
                    %AM3D \cite{ma2019accurate}                & Depth & $16.50$        & $10.74$        & $9.52$        & $25.03$        & $17.32$        & $14.91$       \\                
                    %DA-3Ddet \cite{ye2020monocular}           & Depth & $16.77$        & $11.50$        & $8.93$        & \mathDash{}           & \mathDash{}           & \mathDash{}           \\
                    %D4LCN \cite{ding2020learning}             & Depth & $16.65$        & $11.72$        & $9.51$        & $22.51$        & $16.02$        & $12.55$       \\
                    
                    %\ddmp \cite{wang2021depth}              & Depth & $19.71$        & $12.78$        & $9.80$        & $28.08$        & $17.89$        & $13.44$       \\
                    PCT \cite{wang2021progressive}                    & Depth & $21.00$        & $13.37$        & $11.31$       & $29.65$        & $19.03$        & $15.92$ \\
                    DFR-Net \cite{zou2021devil}               & Depth & $19.40$        & $13.63$        & $10.35$       & $28.17$        & $19.17$        & $14.84$       \\
                    MonoDistill \cite{chong2022monodistill} & Depth & $22.97$ & $16.03$ & $13.60$ & $31.87$ & $22.59$ & $19.72$ \\
                    %MonoPSR \cite{ku2019monocular}            & \lidar& $10.76$        &  $7.25$        & $5.85$        & $18.33$        & $12.58$        &  $9.91$       \\
                    %MonoRUn \cite{chen2021monorun}            & \lidar& $19.65$        & $12.30$        & $10.58$       & $27.94$        & $17.34$        & $15.24$       \\
                    PatchNet-C \cite{simonelli2021we}         & \lidar& $22.40$        & $12.53$        & $10.60$        & \mathDash{}           & \mathDash{}           & \mathDash{}           \\
                    CaDDN \cite{reading2021categorical}       & \lidar& $19.17$        & $13.41$        & $11.46$       & $27.94$        & $18.91$        & $17.19$       \\
                    DD3D \cite{park2021pseudo}               & \lidar &  $23.22$ &  $16.34$ &  $14.20$ &  $30.98$ & $22.56$ & $20.03$\\
                    MonoEF \cite{zhou2021monoef}              & Odometry   & $21.29$        & $13.87$        & $11.71$       & $29.03$        & $19.70$        & $17.26$       \\
                    Kinematic \cite{brazil2020kinematic}      & Video & $19.07$        & $12.72$        & $9.17$        & $26.69$        & $17.52$        & $13.10$       \\
                    \myTopRule
                    \groomedNMS \cite{kumar2021groomed}           & \mathDash{}  & $18.10$        & $12.32$        & $9.65$        & $26.19$        & $18.27$        & $14.05$       \\
                    MonoRCNN \cite{shi2021geometry}           & \mathDash{}  & $18.36$        & $12.65$        & $10.03$       & $25.48$        & $18.11$        & $14.10$       \\
                    \monoDISMulti \cite{simonelli2020disentangling}& \mathDash{} & $16.54$        & $12.97$        & $11.04$       & $24.45$        & $19.25$        & $16.87$       \\
                    Ground-Aware \cite{liu2021ground}               & \mathDash{}  & \second{21.65}        & $13.25$        & $9.91$        & \first{29.81}        & $17.98$        & $13.08$       \\
                    MonoFlex \cite{zhang2021objects}          & \mathDash{}  & $19.94$        & $13.89$        & \first{12.07}& $28.23$        & \second{19.75}        & \second{16.89}       \\
                    \gupNet{} \cite{lu2021geometry} & \mathDash{}  & $20.11$ & \second{14.20} & $11.77$ & \mathDash{}& \mathDash{}& \mathDash\\
                    \hline
                    \bestKey{\methodName{} (Ours)}                        & \mathDash{}  & \first{21.88} & \first{14.46} & \second{11.89} & \second{29.65} & \first{20.44} & \first{17.43}   \\
                    \myTopRule
                \end{tabular}
            }
        \end{table}          
        \begin{table}[!tb]
            \caption{\textbf{Results on \kitti{} Test cyclists and pedestrians} (Cyc/Ped) at \iouThreeD$\geq\!0.5$. Previous results are from the leader-board or papers. [Key: \firstkey{Best}, \secondkey{Second Best}]
            }
            \label{tab:detection_results_kitti_test_ped_cyclist}
            \centering
            \scalebox{\scaleFraction}{
                \footnotesize
                \rowcolors{3}{lightgray}{white}
                \setlength\tabcolsep{0.1cm}
                \begin{tabular}{ml m c m ccc  m cccm}
                    \myTopRule
                    \addlinespace[0.01cm]
                    \multirow{2}{*}{Method} & \multirow{2}{*}{Extra} &\multicolumn{3}{cm}{Cyc \apThreeDForty \bracketPercentage (\uparrowRHDSmall)} & \multicolumn{3}{cm}{Ped \apThreeDForty \bracketPercentage (\uparrowRHDSmall)}\\ 
                    & & Easy & Mod & Hard & Easy & Mod & Hard\\ 
                    \myTopRule
                    \ddmp \cite{wang2021depth}             & Depth    & $4.18$        & $2.50$        & $2.32$  & $4.93$        & $3.55$        & $3.01$     \\
                    DFR-Net \cite{zou2021devil}               & Depth    & $5.69$        & $3.58$        & $3.10$  & $6.09$        & $3.62$        & $3.39$     \\
                    MonoDistill \cite{chong2022monodistill}       & Depth  & $5.53$        & $2.81$        & $2.40$  & $12.79$        & $8.17$        & $7.45$\\            
                    CaDDN \cite{reading2021categorical}       & \lidar  & $7.00$        & $3.41$        & $3.30$  & $12.87$        & $8.14$        & $6.76$            \\
                    DD3D \cite{park2021pseudo}               & \lidar &  $2.39$ &  $1.52$ & $1.31$ &  $13.91$ & $9.30$ &  $8.05$ \\
                    MonoEF \cite{zhou2021monoef}              & Odometry   & $1.80$ & $0.92$        & $0.71$ & $4.27$        & $2.79$        & $2.21$       \\
                    \myTopRule
                    %\mthreeDRPN \cite{brazil2019m3d}              & \mathDash{}  & $0.94$        & $0.65$        & $0.47$     & $4.92$        &  $3.48$        & $2.94$           \\
                    \monoDISMulti \cite{simonelli2020disentangling}& \mathDash{}   & $1.17$        & $0.54$        & $0.48$    & $7.79$        & $5.14$        & $4.42$          \\
                    MonoFlex \cite{zhang2021objects}          & \mathDash{}  & $3.39$        & $2.10$        & $1.67$      & $11.89$        & $8.16$       & $6.81$  \\
                    \gupNet{} \cite{lu2021geometry} & \mathDash{}  & \second{4.18} & \second{2.65} & \second{2.09} & \first{14.72} & \first{9.53} & \first{7.87} \\
                    \hline	 
                    \rowcolor{white}
                    \bestKey{\methodName{} (Ours)}                        & \mathDash{}  & \first{5.05} & \first{3.13} & \first{2.59}  & \second{13.43} & \second{8.65} & \second{7.69} \\
                    \myTopRule
                \end{tabular}
            }
        \end{table}         
        \begin{table}[!tb]
            \caption{\textbf{Results on \kitti{} \valOne{} cars}. 
            Comparison with bigger CNN backbones in \refSupTable{16}. %{tab:detection_with_bigger_cnn}. 
            [Key: \firstkey{Best}, \secondkey{Second Best}, $^{-}$= No \pretrain] %, Rep= Reported, $^\dagger$= Retrained]
            }
            \label{tab:detection_results_kitti_val1}
            \centering
            \scalebox{\scaleFraction}{
            \footnotesize
            \rowcolors{4}{lightgray}{white}
            \setlength\tabcolsep{0.05cm}
            \begin{tabular}{ml m c m ccc t ccc m ccc t cccm}
                \myTopRule
                \addlinespace[0.01cm]
                \multirow{3}{*}{Method} & \multirow{3}{*}{Extra} & \multicolumn{6}{cm}{\iouThreeD{} $\geq 0.7$} & \multicolumn{6}{cm}{\iouThreeD{} $\geq 0.5$}\\\cline{3-14}
                & & \multicolumn{3}{ct}{\apThreeDForty \bracketPercentage(\uparrowRHDSmall)} & \multicolumn{3}{cm}{\apBevForty \bracketPercentage(\uparrowRHDSmall)} & \multicolumn{3}{ct}{\apThreeDForty \bracketPercentage(\uparrowRHDSmall)} & \multicolumn{3}{cm}{\apBevForty \bracketPercentage(\uparrowRHDSmall)}\\%\cline{2-13}
                & & Easy & Mod & Hard & Easy & Mod & Hard & Easy & Mod & Hard & Easy & Mod & Hard\\ 
                \myTopRule
                %MonoDR \cite{beker2020monocular}   & Mask                                            & $12.50$        & $7.34$         & $4.98$        & $19.49$        & $11.51$        & $8.72$       & \mathDash{}           & \mathDash{}           & \mathDash{}           & \mathDash{}           & \mathDash{}           & \mathDash{}           \\
                %DFR-Net \cite{zou2021devil} & Depth & $19.55$ & $14.79$ & $11.04$ & $26.60$ & $19.80$ & $15.34$ & \mathDash{}           & \mathDash{}           & \mathDash{}           & \mathDash{}           & \mathDash{}           & \mathDash\\
                \ddmp \cite{wang2021depth} & Depth & $28.12$ & $20.39$ & $16.34$ & \mathDash{}           & \mathDash{}           & \mathDash{}& \mathDash{}           & \mathDash{}           & \mathDash{}           & \mathDash{}           & \mathDash{}           & \mathDash\\
                PCT \cite{wang2021progressive} & Depth & $38.39$ & $27.53$ & $24.44$ & $47.16$            & $34.65$            & $28.47$& \mathDash{}           & \mathDash{}           & \mathDash{}           & \mathDash{}           & \mathDash{}           & \mathDash\\
                MonoDistill \cite{chong2022monodistill} & Depth & $24.31$ & $18.47$ & $15.76$ &  $33.09$ & $25.40$ & $22.16$       & $65.69$ & $49.35$        & $43.49$        & $71.45$ & $53.11$        & $46.94$ \\
                %MonoRUn \cite{chen2021monorun} & \lidar & $20.02$ & $14.65$ & $12.61$ & \mathDash{}           & \mathDash{}           & \mathDash{}& \mathDash{}           & \mathDash{}           & \mathDash{}           & \mathDash{}           & \mathDash{}           & \mathDash\\
                CaDDN \cite{reading2021categorical} & \lidar & $23.57$ & $16.31$ & $13.84$& \mathDash{}           & \mathDash{}           & \mathDash{}& \mathDash{}           & \mathDash{}           & \mathDash{}           & \mathDash{}           & \mathDash{}           & \mathDash\\
                PatchNet-C \cite{simonelli2021we} & \lidar & $24.51$ & $17.03$ & $13.25$ & \mathDash{}           & \mathDash{}           & \mathDash{}& \mathDash{}           & \mathDash{}           & \mathDash{}           & \mathDash{}           & \mathDash{}           & \mathDash\\
                \ddThreeD~(DLA34) \cite{park2021pseudo} & \lidar & \mathDash{}& \mathDash{}& \mathDash{}& $33.5$ & $26.0$ & $22.6$ & \mathDash{}           & \mathDash{}           & \mathDash{}           & \mathDash{}           & \mathDash{}           & \mathDash\\
                \ddThreeD$^{-}$\!(DLA34)\cite{park2021pseudo} & \lidar & \mathDash{}& \mathDash{}& \mathDash{}& $26.8$ & $20.2$ & $16.7$ & \mathDash{}          & \mathDash{}           & \mathDash{}           & \mathDash{}           & \mathDash{}           & \mathDash\\
                MonoEF \cite{zhou2021monoef} & Odometry & $18.26$& $16.30$ & $15.24$& $26.07$ & $25.21$ & $21.61$ & $57.98$     & $51.80$            & $49.34$            & $63.40$            & $61.13$            & $53.22$\\        
                Kinematic \cite{brazil2020kinematic}  & Video                           & $19.76$ & $14.10$ & $10.47$ &  $27.83$ & $19.72$ & $15.10$       & $55.44$ & $39.47$        & $31.26$        & $61.79$ & $44.68$        & $34.56$\\
                \myTopRule
                MonoRCNN \cite{shi2021geometry}  & \mathDash{} & $16.61$        & $13.19$ & $10.65$ & $25.29$ & $19.22$ & $15.30$ & \mathDash{}           & \mathDash{}           & \mathDash{}           & \mathDash{}           & \mathDash{}           & \mathDash\\
                % UR3D \cite{shi2020distance} & \mathDash{} & $23.24$        & $13.35$ & $10.15$ & \first{33.07} & $20.84$ & $15.25$ & \mathDash{}           & \mathDash{}           & \mathDash{}           & \mathDash{}           & \mathDash{}           & \mathDash\\
                MonoDLE \cite{ma2021delving}   & \mathDash{}& $17.45$ & $13.66$ & $11.68$ & $24.97$ & $19.33$ & $17.01$ & $55.41$ & $43.42$ & $37.81$ & $60.73$ & $46.87$ & $41.89$\\
                \groomedNMS \cite{kumar2021groomed}  & \mathDash{}                                                 & $19.67$ & $14.32$ & $11.27$ & $27.38$ & $19.75$ & $15.92$ & $55.62$ & $41.07$ & $32.89$ & $61.83$ & $44.98$ & $36.29$\\
                Ground-Aware \cite{liu2021ground} & \mathDash{}& $23.63$ & $16.16$ & $12.06$ & \mathDash{}& \mathDash{}& \mathDash{}& \second{60.92} & $42.18$ & $32.02$ & \mathDash{}           & \mathDash{}           & \mathDash{}\\ 
                MonoFlex \cite{zhang2021objects} & \mathDash{}& \second{23.64} & \first{17.51} & \first{14.83} & \mathDash{}           & \mathDash{}           & \mathDash{}& \mathDash{}           & \mathDash{}           & \mathDash{}           & \mathDash{}           & \mathDash{}           & \mathDash\\
                \gupNet{} (Reported)\!\cite{lu2021geometry}\!& \mathDash{}& $22.76$ & $16.46$ & $13.72$ & \second{31.07} & \second{22.94} & \second{19.75} & 57.62 & 42.33 & 37.59 & 61.78 & 47.06 & 40.88 \\
                %\gupNet{} (Released) \cite{lu2021geometry} & \mathDash{}& $23.19$	& $16.23$	& $13.57$	& \second{30.14}	& \second{22.38} & \second{19.29} & $59.01$ & $43.63$ & $39.34$ & \second{65.16} & $48.88$ & $42.93$ \\
                \gupNet{} (Retrained)\!\cite{lu2021geometry}\!& \mathDash{}& $21.10$ & $15.48$ & $12.88$ & $28.58$ & $20.92$ & $17.83$ & $58.95$ & \second{43.99} & \second{38.07} & \second{64.60} & \second{47.76} &	\second{42.97} \\
                \hline
                \rowcolor{white}
                \textbf{\methodName{} (Ours)} & \mathDash{}& \first{24.63} & \second{16.54} & \second{14.52} & \first{32.60} &	\first{23.04} & \first{19.99} &	\first{61.00} & \first{46.00}	& \first{40.18} & \first{65.28} & \first{49.63} & \first{43.50} \\
                \myTopRule
            \end{tabular}
            }
        \end{table}

    %============================================================================
    %============================================================================
    \subsection{\kitti{} Test Monocular 3D Detection}\label{sec:detection_results_kitti_test}

        %============================================================================
        \noIndentHeading{Cars.}
            \cref{tab:detection_results_kitti_test_cars} lists out the results of monocular \threeD{} detection and BEV evaluation on \kitti{} Test cars. 
            \cref{tab:detection_results_kitti_test_cars} results show that \methodName{} outperforms the \gupNet{} and several other \sota{} methods on both  tasks. 
            Except \ddThreeD \cite{park2021pseudo} and MonoDistill \cite{chong2022monodistill}, \methodName, an image-based method, also outperforms other methods that use extra information.

        %============================================================================
        \noIndentHeading{Cyclists and Pedestrians.} 
            \cref{tab:detection_results_kitti_test_ped_cyclist} lists out the results of monocular \threeD{} detection on \kitti{} Test Cyclist and Pedestrians. 
            The results show that \methodName{} achieves \sota{} results in the \imageOnly{} category on the challenging Cyclists, and is competitive on Pedestrians.

    %============================================================================
    %============================================================================
    \subsection{\kitti{} \valOne{} Monocular 3D Detection}\label{sec:detection_results_kitti_val1}
    
        %============================================================================
        \noIndentHeading{Cars.}
            \cref{tab:detection_results_kitti_val1} summarizes the results of monocular \threeD{} detection and BEV evaluation on \kitti{} \valOne{} split at two \iouThreeD{} thresholds of $0.7$ and $0.5$ \cite{chen2020monopair,kumar2021groomed}. 
            We report the \textbf{median} model over 5 runs.
            The results show that \methodName{} outperforms the \gupNet{} \cite{lu2021geometry} baseline by a significant margin. 
            The biggest improvements shows up on the Easy set. 
            Significant improvements are also on the Moderate and Hard sets. %, where objects are more distant and occluded.
            Interestingly, \methodName{} also outperforms \ddThreeD \cite{park2021pseudo} by a large margin when the large-dataset \pretrain{}ing is not done (denoted by \ddThreeD$^{-}$).

        \begin{figure}[!tb]
            \centering
            \begin{subfigure}{.4\linewidth}
              \centering
              \includegraphics[width=\linewidth]{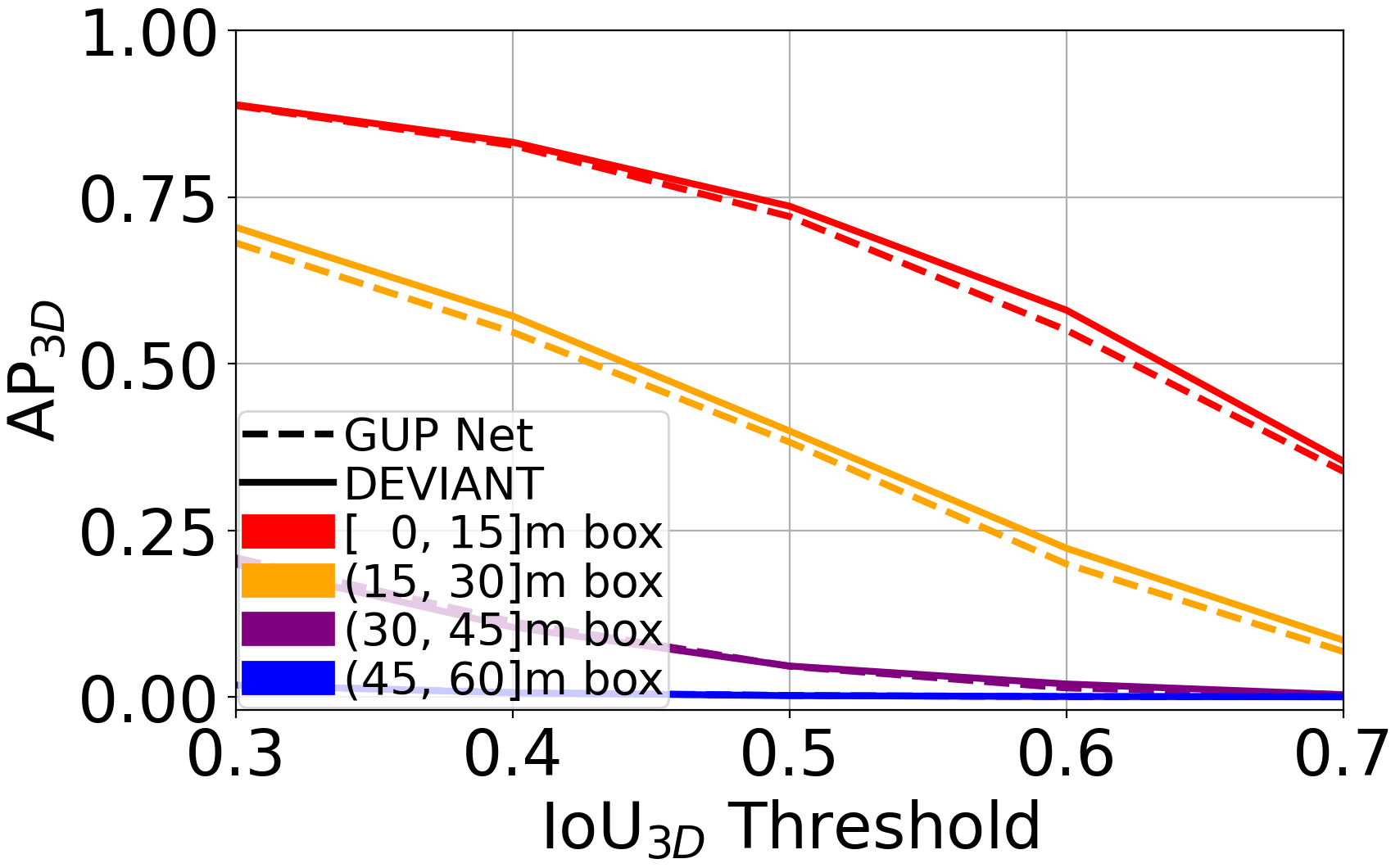}
              \caption{Linear Scale}
            \end{subfigure}%
            \begin{subfigure}{.4\linewidth}
              \centering
              \includegraphics[width=\linewidth]{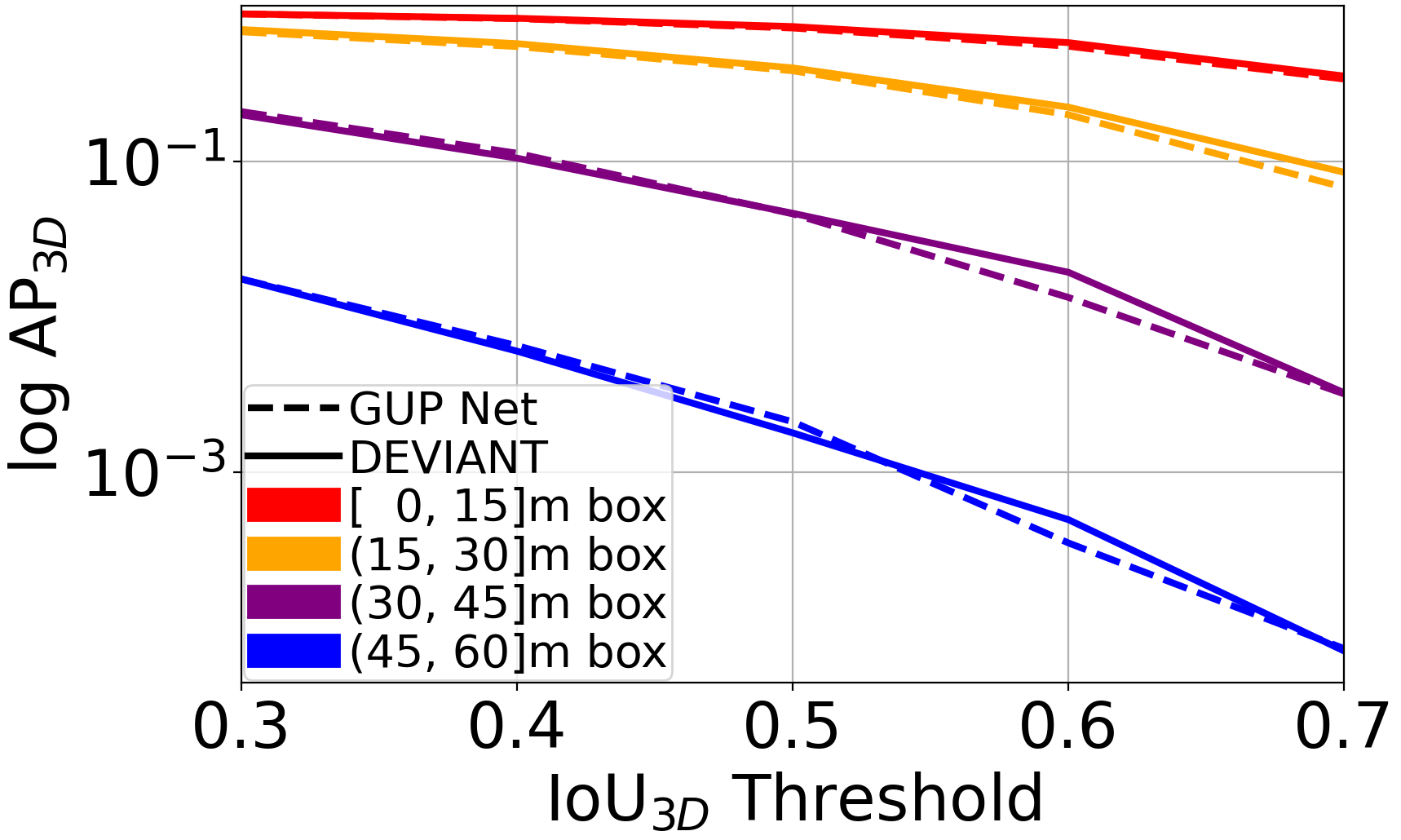}
              \caption{Log Scale}
            \end{subfigure}
            \caption{\textbf{\apThreeD{} at different depths and \iouThreeD{} thresholds} on \kitti{} \valOne{} Split. }
            \label{fig:ap_ground_truth_threshold}
        \end{figure}
        \begin{table}[!tb]
            \caption{\textbf{Cross-dataset evaluation}  of the \kitti{} \valOne{} model on \kitti{} \valOne{} and \nuscenes{} \frontal~\val{} cars with depth MAE (\downarrowRHDSmall). [Key: \firstkey{Best}, \secondkey{Second Best}]}
            \label{tab:detection_cross_dataset}
            \centering
            \scalebox{\scaleFraction}{
                \footnotesize
                \rowcolors{4}{white}{lightgray}
                \setlength{\tabcolsep}{0.1cm}
                \begin{tabular}{tlt ccc | c t ccc | ct}
                    \myTopRule
                    \addlinespace[0.01cm]
                    \multirow{2}{*}{Method} & \multicolumn{4}{ct}{\kitti{} \valOne} & \multicolumn{4}{ct}{\nuscenes{} \frontal~\val}\\\cline{2-9}
                    &$0\!-\!20$&$20\!-\!40$&$40\!-\!\infty$& All & $0\!-\!20$&$20\!-\!40$& $40\!-\!\infty$&All\\
                    \myTopRule
                    \mthreeDRPN \cite{brazil2019m3d} & $0.56$ & $1.33$ & $2.73$ & $1.26$ & $0.94$ & $3.06$ & $10.36$ & $2.67$\\
                    MonoRCNN \cite{shi2021geometry} &$0.46$& $1.27$ & $2.59$ & $1.14$ & $0.94$ & $2.84$ & $8.65$ & $2.39$\\
                    %\groomedNMS \cite{kumar2021groomed} &\\
                    \gupNet{} \cite{lu2021geometry} & \second{0.45} & \second{1.10} & \second{1.85} & \second{0.89} & \second{0.82} & \second{1.70} & \second{6.20} & \second{1.45}\\
                    \hline
                    \bestKey{\methodName} & \first{0.40} & \first{1.09} & \first{1.80} & \first{0.87} & \first{0.76} & \first{1.60} & \first{4.50} & \first{1.26}\\        
                    \myTopRule
                \end{tabular}
            }
        \end{table}

        %============================================================================
        \noIndentHeading{\apThreeD{} at different depths and \iouThreeD{} thresholds.}
            We next compare the \apThreeD{} of \methodName{} and \gupNet{} in \cref{fig:ap_ground_truth_threshold} at different distances in meters and \iouThreeD{} matching criteria of $0.3\!\rightarrowRHD\!0.7$ as in \cite{kumar2021groomed}. 
            \cref{fig:ap_ground_truth_threshold} shows that \methodName{} is effective over \gupNet{} \cite{lu2021geometry} at all depths and higher \iouThreeD{} thresholds.
                            
        %============================================================================
        \noIndentHeading{Cross-Dataset Evaluation.}
            \cref{tab:detection_cross_dataset} shows the result of our \kitti{} \valOne{} model on the \kitti{} \valOne{} and \nuscenes{} \cite{caesar2020nuscenes} \frontal{} \val{} images, using mean absolute error (MAE) of the depth of the boxes \cite{shi2021geometry}.
            More details are in \refSupApp{3.1}. %\cref{sec:results_cross_dataset_additional}.
            \methodName{} outperforms \gupNet{} on most of the metrics on both the datasets, which confirms that \methodName{} generalizes better than CNNs. 
            \methodName{} performs exceedingly well in the cross-dataset evaluation than \cite{brazil2019m3d, shi2021geometry, lu2021geometry}.
            We believe this happens because \cite{brazil2019m3d, shi2021geometry, lu2021geometry} rely on data or geometry to get the depth, while \methodName{} is \equivariant{} to the depth translations, and therefore, outputs consistent depth. 
            So, \methodName{} is more robust to data distribution changes. 

        %============================================================================
        \noIndentHeading{Alternatives to \Equivariance.} We now compare with alternatives to \equivariance{} in the following paragraphs.
                
        %============================================================================
        \noIndentHeading{(a) Scale Augmentation.}
            A withstanding question in machine learning is the choice between \equivariance{} and data augmentation \cite{gandikota2021training}.
            \cref{tab:kitti_aug_vs_eq} compares \scaleEquivariance{} and scale augmentation.
            \gupNet{} \cite{lu2021geometry} uses scale-augmentation and therefore, \cref{tab:kitti_aug_vs_eq} shows that \equivariance{} also benefits models which use scale-augmentation. 
            This agrees with Tab. 2 of \cite{sosnovik2020sesn}, where they observe that both augmentation and \equivariance{} benefits classification on MNIST-scale dataset.

        \begin{table}[!tb]
            \caption{\textbf{Scale Augmentation vs Scale \Equivariance} on \kitti{} \valOne{} cars. [Key: \bestKey{Best}, Eqv= \Equivariance, Aug= Augmentation]}
            \label{tab:kitti_aug_vs_eq}
            \centering
            \scalebox{\scaleFraction}{
                \footnotesize
                \rowcolors{4}{lightgray}{white}
                \setlength{\tabcolsep}{0.08cm}
                \begin{tabular}{tl|c|cm ccc t ccc m ccc t ccct}
                    \myTopRule
                    \addlinespace[0.01cm]
                    \multirow{3}{*}{Method} & Scale & Scale & \multicolumn{6}{cm}{\iouThreeD{} $\geq 0.7$} & \multicolumn{6}{ct}{\iouThreeD{} $\geq 0.5$}\\
                    \cline{4-15}
                    & Eqv & Aug  & \multicolumn{3}{ct}{\apThreeDForty \bracketPercentage(\uparrowRHDSmall)} & \multicolumn{3}{cm}{\apBevForty \bracketPercentage(\uparrowRHDSmall)} & \multicolumn{3}{ct}{\apThreeDForty \bracketPercentage(\uparrowRHDSmall)} & \multicolumn{3}{ct}{\apBevForty \bracketPercentage(\uparrowRHDSmall)}\\%\cline{2-13}\\[0.05cm]
                    & & & Easy & Mod & Hard & Easy & Mod & Hard & Easy & Mod & Hard & Easy & Mod & Hard\\
                    \myTopRule
                    \gupNet\!\cite{lu2021geometry}&  & & $20.82$ & $14.15$ & $12.44$ & $29.93$ & $20.90$ & $17.87$ & \best{62.37} & $44.40$ & $39.61$ & \best{66.81} & $48.09$ &	$43.14$\\
                    & & \checkmark & $21.10$ & $15.48$ & $12.88$ & $28.58$ & $20.92$ & $17.83$ & $58.95$ & $43.99$ & $38.07$ & $64.60$ & $47.76$ &	$42.97$\\
                    \myTopRule
                    \textbf{\methodName} & \checkmark & & $21.33$ & $14.77$ & $12.57$ & $28.79$ & $20.28$ & $17.59$ & $59.31$ & $43.25$ & $37.64$ & $63.94$ & $47.02$ & $41.12$\\
                    & \checkmark & \checkmark & \best{24.63} & \best{16.54} & \best{14.52} & \best{32.60} &	\best{23.04} & \best{19.99} &	$61.00$ & \best{46.00}	& \best{40.18} & $65.28$ & \best{49.63} & \best{43.50} \\
                    \myTopRule
                \end{tabular}
            }
        \end{table}
        \begin{table}[!tb]
            \caption{\textbf{Comparison of \Equivariant{} Architectures} on~\kitti{} \valOne{} cars. [Key: \bestKey{Best}, Eqv= \Equivariance, $^\dagger$= Retrained]}
            \label{tab:kitti_compare_eq}
            \centering
            \scalebox{\scaleFraction}{
                \footnotesize
                \rowcolors{4}{lightgray}{white}
                \setlength\tabcolsep{0.1cm}
                \begin{tabular}{ml m c m ccc t ccc m ccc t cccm}
                    \myTopRule
                    \addlinespace[0.01cm]
                    \multirow{3}{*}{Method} & \multirow{3}{*}{Eqv} & \multicolumn{6}{cm}{\iouThreeD{} $\geq 0.7$} & \multicolumn{6}{cm}{\iouThreeD{} $\geq 0.5$}\\
                    \cline{3-14}
                    & & \multicolumn{3}{ct}{\apThreeDForty \bracketPercentage(\uparrowRHDSmall)} & \multicolumn{3}{cm}{\apBevForty \bracketPercentage(\uparrowRHDSmall)} & \multicolumn{3}{ct}{\apThreeDForty \bracketPercentage(\uparrowRHDSmall)} & \multicolumn{3}{cm}{\apBevForty \bracketPercentage(\uparrowRHDSmall)}\\%\cline{2-13}
                    & & Easy & Mod & Hard & Easy & Mod & Hard & Easy & Mod & Hard & Easy & Mod & Hard\\ 
                    \myTopRule
                    DETR3D$^\dagger$\!\cite{wang2021detr3d} & Learned & $1.94$ & $1.26$ & $1.09$ & $4.41$ & $3.06$ & $2.79$ & $20.09$ & $13.80$ & $12.78$  & $26.51$ & $18.49$ & 	$17.36$\\
                    \gupNet\!\cite{lu2021geometry}& \twoD & $21.10$ & $15.48$ & $12.88$ & $28.58$ & $20.92$ & $17.83$ & $58.95$ & $43.99$ & $38.07$ & $64.60$ & $47.76$ &	$42.97$\\
                    \bestKey{\methodName} & \twoD+Depth & \best{24.63} & \best{16.54} & \best{14.52} & \best{32.60} &	\best{23.04} & \best{19.99} &	\best{61.00} & \best{46.00}	& \best{40.18} & \best{65.28} & \best{49.63} & \best{43.50} \\
                    \myTopRule
                \end{tabular}
            }
        \end{table}
        \begin{table}[!tb]
            \caption{\textbf{Comparison with Dilated Convolution} on~\kitti{} \valOne{} cars. [Key: \bestKey{Best}]}
            \label{tab:kitti_compare_dilation}
            \centering
            \scalebox{\scaleFraction}{
                \footnotesize
                \rowcolors{5}{lightgray}{white}
                \setlength\tabcolsep{0.1cm}
                \begin{tabular}{mc| cm ccc t ccc m ccc t cccm}
                    \myTopRule
                    \addlinespace[0.01cm]
                    \multirow{3}{*}{Method} & \multirow{3}{*}{Extra} & \multicolumn{6}{cm}{\iouThreeD{}$\geq 0.7$} & \multicolumn{6}{ct}{\iouThreeD{}$\geq 0.5$}\\\cline{3-14}
                    & & \multicolumn{3}{ct}{\apThreeDForty \bracketPercentage(\uparrowRHDSmall)} & \multicolumn{3}{cm}{\apBevForty \bracketPercentage(\uparrowRHDSmall)} & \multicolumn{3}{ct}{\apThreeDForty \bracketPercentage(\uparrowRHDSmall)} & \multicolumn{3}{cm}{\apBevForty \bracketPercentage(\uparrowRHDSmall)}\\%\cline{2-13}\\[0.05cm]
                    && Easy & Mod & Hard & Easy & Mod & Hard & Easy & Mod & Hard & Easy & Mod & Hard\\ 
                    \myTopRule
                    D4LCN \cite{ding2020learning} & Depth & $22.32$ & $16.20$ & $12.30$ & $31.53$ & $ 22.58$ & $17.87$ & \mathDash{} & \mathDash{} & \mathDash{} & \mathDash{} & \mathDash{} & \mathDash\\
                    DCNN \cite{yu2015multi} & \mathDash{}&$21.66$ & $15.49$ & $12.90$ & $30.22$ & $22.06$ & $19.01$ & $57.54$ & $43.12$ & $38.80$ & $63.29$ & $46.86$ & $42.42$\\
                    \bestKey{\methodName} & \mathDash{}& \best{24.63} & \best{16.54} & \best{14.52} & \best{32.60} &	\best{23.04} & \best{19.99} &	\best{61.00} & \best{46.00}	& \best{40.18} & \best{65.28} & \best{49.63} & \best{43.50}\\
                    \myTopRule
                \end{tabular}
            }
        \end{table}

        %============================================================================
        \noIndentHeading{(b) Other \Equivariant{} Architectures.}
            We now benchmark adding depth (scale) \equivariance{} to a \twoD{} translation \equivariant{} CNN and a transformer which learns the \equivariance. 
            Therefore, we compare \methodName{} with \gupNet{} \cite{lu2021geometry} (a CNN), and  \detrThreeD \cite{wang2021detr3d} (a transformer) in \cref{tab:kitti_compare_eq}.
            As \detrThreeD~does not report \kitti{} results, we trained \detrThreeD~on \kitti{} using their public code.
            \methodName{} outperforms \gupNet{} and also surpasses \detrThreeD~by a large margin. This happens because learning \equivariance{} requires more data \cite{worrall2018cubenet} compared to architectures which hardcode \equivariance{} like CNN or \methodName.
 
        %============================================================================
        \noIndentHeading{(c) Dilated Convolution.}
            \methodName{} adjusts the receptive field based on the object scale, and so, we compare with the dilated CNN (DCNN) \cite{yu2015multi} and D4LCN \cite{ding2020learning} in \cref{tab:kitti_compare_dilation}.
            The results show that DCNN performs sub-par to \methodName.
            This is expected because dilation corresponds to integer scales \cite{worrall2019deep} while the scaling is generally a float in monocular detection.
            D4LCN \cite{ding2020learning} uses monocular depth as input to adjust the receptive field. 
            \methodName{} (without depth) also outperforms D4LCN on Hard cars, which are more distant. 

        %============================================================================
        \noIndentHeading{(d) Other Convolutions.}
            We now compare with other known convolutions in literature such as \LogPolar{} convolution \cite{zwicke1983new}, Dilated convolution \cite{yu2015multi} convolution and DISCO \cite{sosnovik2021disco} in \cref{tab:ablation}. 
            The results show that the \logPolar{} convolution does not work well, and \ses{} convolutions are better suited to embed depth (scale) equivariance.
            As described in~\cref{sec:log_polar}, we investigate the behavior of \logPolar{} convolution through a small experiment.
            We calculate the SSIM \cite{wang2004image} of the original image and the image obtained after the upscaling, \logPolar, inverse \logPolar, and downscaling blocks.
            We then average the SSIM over all \kitti{} \valOne{} images. 
            We repeat this experiment for multiple image heights and scaling factors.
            The ideal SSIM should have been one.
            However, \cref{fig:log_polar_ssim} 
            shows that SSIM %is very low at small resolutions, and 
            does not reach $1$ even after upscaling by $4$. 
            This result confirms that \logPolar{} convolution loses information at low resolutions resulting in inaccurate detection.
            
            Next, the results show that dilated convolution \cite{yu2015multi} performs sub-par to \methodName. 
            Moreover, DISCO \cite{sosnovik2021disco} also does not outperform \ses{} convolution which agrees with the \twoD{} tracking results of \cite{sosnovik2021disco}.

        %============================================================================
        \noIndentHeading{(e) Feature Pyramid Network (FPN).}
            Our baseline \gupNet{} \cite{lu2021geometry} uses FPN \cite{lin2017feature} and \cref{tab:detection_results_kitti_val1} shows that \methodName{} outperforms \gupNet. Hence, we conclude that \equivariance{} also benefits models which use FPN.

        \begin{figure}[!tb]
            \centering
            \begin{subfigure}{.36\linewidth}
              \centering
              \includegraphics[width=\linewidth]{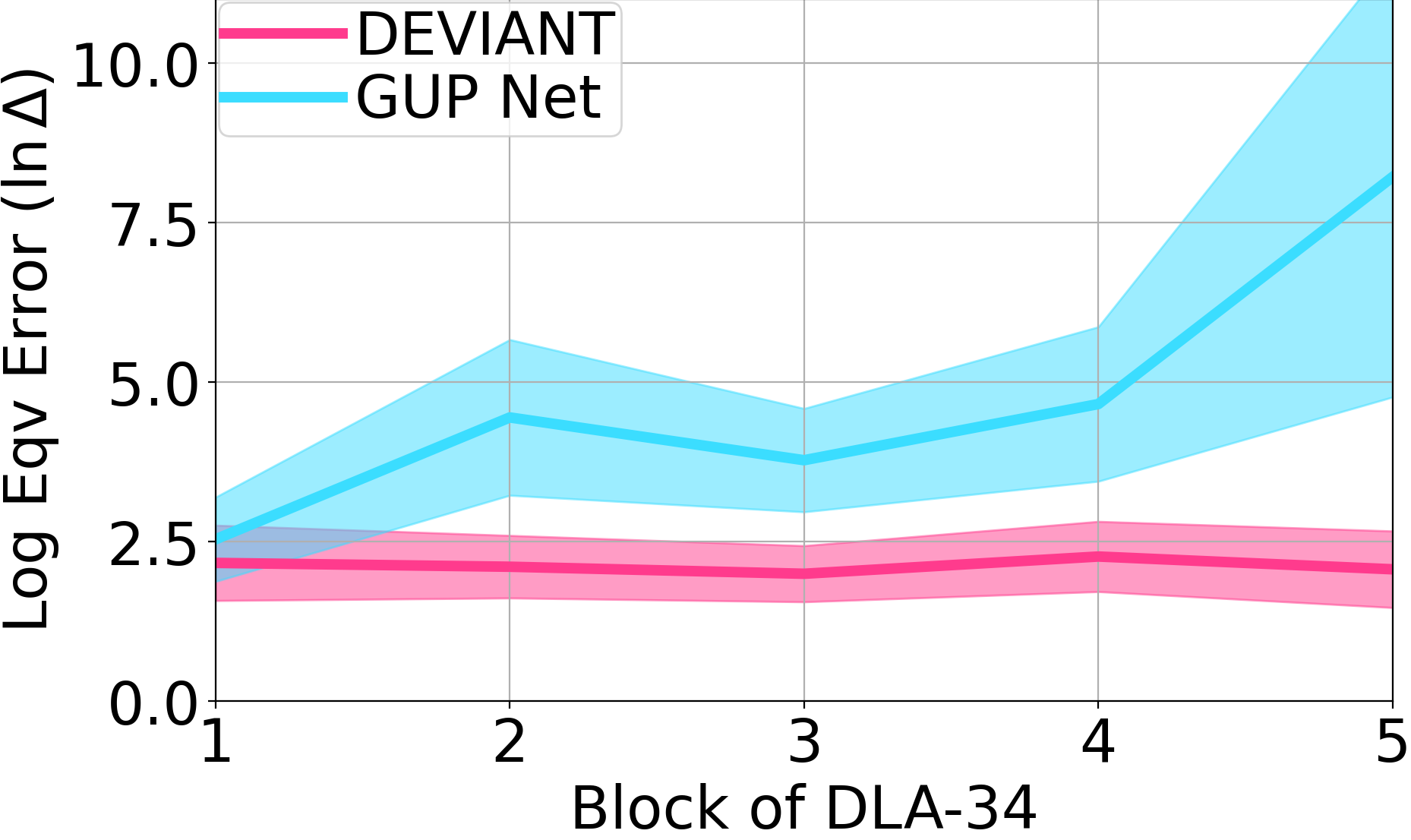}
              \caption{At blocks (depths) of backbone.}
            \end{subfigure}%
            \begin{subfigure}{.36\linewidth}
              \centering
              \includegraphics[width=0.96\linewidth]{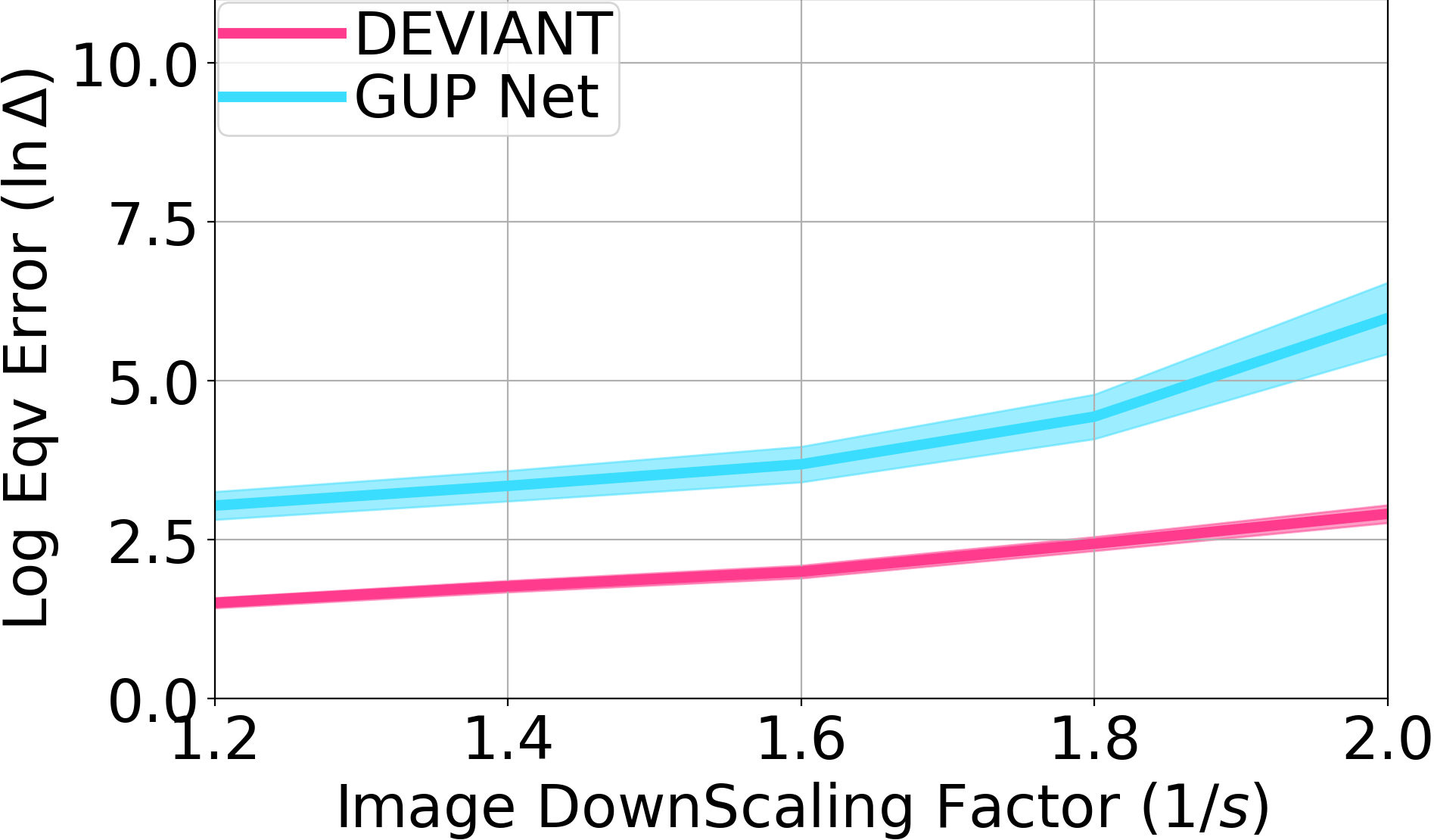}
              \caption{Varying scaling factors.}
              \label{fig:equiv_error_scaling}
            \end{subfigure}
            \caption{\textbf{Log \Equivariance{} Error $(\Delta)$} comparison for \methodName{} and \gupNet{} at \textbf{(a)} different blocks with random image scaling factors \textbf{(b)} different image scaling factors at depth $3$. \methodName{} shows \textbf{lower} \scaleEquivariance{} error than vanilla \gupNet{} \cite{lu2021geometry}.}
            \label{fig:equiv_error}
        \end{figure}

        %============================================================================
        \noIndentHeading{Comparison of \Equivariance{} Error.}
            We next quantitatively evaluate the \scaleEquivariance{} of \methodName{} vs. \gupNet{} \cite{lu2021geometry}, using the \equivariance{} error~metric\cite{sosnovik2020sesn}.
            The \equivariance{} error $\Delta$ is the normalized difference between the scaled feature map and the feature map of the scaled image, and is given by
                $\Delta = \frac{1}{N} \sum_{i=1}^N \frac{||\transformationMath_{s_i} \mapping(\projectionOneIndexed) - \mapping(\transformationMath_{s_i} \projectionOneIndexed)||_2^2}{||\transformationMath_{s_i} \mapping(\projectionOneIndexed)||_2^2},$
                % \label{eq:equiv_error}
            where $\mapping$ denotes the neural network, $\transformationMath_{s_i}$ is the scaling transformation for the image $i$, and $N$ is the total number of images.
            The \equivariance{} error is zero if the \scaleEquivariance{} is perfect.
            We plot the log of this error at different blocks of \methodName{} and \gupNet{} backbones and also plot at different downscaling of \kitti{} \valOne{} images in \cref{fig:equiv_error}.
            % We carry two experiments to demonstrate that \methodName{} achieves desired behaviour. 
            % We take \kitti{} \valOne{} images and scale them by a random number between $1$ and $2$. 
            % We then pass these scaled images through \methodName{} and \gupNet{} backbones and obtain the feature representations of the images. 
            % We finally calculate the  the $\Delta$ at different blocks in the backbones, and show it in \cref{fig:equiv_error}.
            % As an other experiment, we fix the scales and calculate the equivariance error at level 2 (same as output resolution), and plot them in \cref{fig:equiv_error}.
            The plots show that \methodName{} has low \equivariance{} error than \gupNet.
            This is expected since the feature maps of the proposed \methodName{} are additionally \equivariant{} to scale \transformation s (depth translations).
            We also visualize the \equivariance{} error for a validation image and for the objects of this image in \refSupFigure{12} in the supplementary. %\cref{fig:equiv_error_qualitative,fig:equiv_error_qualitative_object}.
            The qualitative plots also show a lower error for the proposed \methodName, which agrees with \cref{fig:equiv_error}.
            %\cref{fig:equiv_error_qualitative_object} 
            \refSupFigure{12a} shows that \equivariance{} error is particularly low for nearby cars which also justifies the good performance of \methodName{} on Easy (nearby) cars in \cref{tab:detection_results_kitti_test_cars,tab:detection_results_kitti_val1}.
    
        %============================================================================
        \noIndentHeading{Does 2D Detection Suffer?}
            We now investigate whether \twoD{} detection suffers from using \methodName{} backbones in \cref{tab:results_kitti_compare_2d_3d}.
            The results show that \methodName{} introduces minimal decrease in the \twoD{} detection performance.
            This is consistent with \cite{sosnovik2021siamese}, who report that \twoD{} tracking improves with the \se{} networks.          
        \begin{table*}[!tb]
            \caption{
            \textbf{3D~and 2D~detection} on \kitti{} \valOne{} cars.}
            \label{tab:results_kitti_compare_2d_3d}
            \centering
            \scalebox{\scaleFraction}{
                \footnotesize
                \rowcolors{4}{lightgray}{white}
                \setlength\tabcolsep{0.1cm}
                \begin{tabular}{tl m ccc t ccc m ccc t ccct}
                    \myTopRule
                    \addlinespace[0.01cm]
                    \multirow{3}{*}{Method} & \multicolumn{6}{cm}{\iou~$\geq 0.7$} & \multicolumn{6}{ct}{\iou~$\geq 0.5$}\\\cline{2-13}
                    & \multicolumn{3}{ct}{\apThreeDForty \bracketPercentage(\uparrowRHDSmall)} & \multicolumn{3}{cm}{\apTwoDForty \bracketPercentage(\uparrowRHDSmall)} & \multicolumn{3}{ct}{\apThreeDForty \bracketPercentage(\uparrowRHDSmall)} & \multicolumn{3}{ct}{\apTwoDForty \bracketPercentage(\uparrowRHDSmall)}\\
                    & Easy & Mod & Hard & Easy & Mod & Hard & Easy & Mod & Hard & Easy & Mod & Hard\\ \myTopRule                
                    \gupNet{} \cite{lu2021geometry} & $21.10$ & $15.48$ & $12.88$ & $96.78$ & $88.87$ & $79.02$ & $58.95$ & $43.99$ & $38.07$ & $99.52$ & $91.89$ &	$81.99$ \\
                    \bestKey{\methodName{} (Ours)} & $24.63$ & $16.54$ & $14.52$ & $96.68$	& $88.66$	& $78.87$ & $61.00$ &	$46.00$ & $40.18$ & $97.12$ & $91.77$ &	$81.93$\\
                    \myTopRule
                \end{tabular}
            }
        \end{table*} 
        \begin{table*}[!tb]
            \caption{\textbf{Ablation studies} on \kitti{} \valOne{} cars.}
            \label{tab:ablation}
            \centering
            \scalebox{\scaleFraction}{
                \footnotesize
                \rowcolors{5}{lightgray}{white}
                \setlength{\tabcolsep}{0.05cm}
                \begin{tabular}{m c m l m ccc t ccc m ccc t ccc m}
                    \myTopRule
                    \addlinespace[0.01cm]
                    \multicolumn{2}{mcm}{\textbf{Change from \methodName{} :}} & \multicolumn{6}{cm}{\iouThreeD{} $\geq 0.7$} & \multicolumn{6}{cm}{\iouThreeD{} $\geq 0.5$}\\
                    \cline{1-14}
                    \multirow{2}{*}{Changed} & \multirow{2}{*}{From $\longrightarrowRHD$To} & \multicolumn{3}{ct}{\apThreeDForty \bracketPercentage(\uparrowRHDSmall)} & \multicolumn{3}{cm}{\apBevForty \bracketPercentage(\uparrowRHDSmall)} & \multicolumn{3}{ct}{\apThreeDForty \bracketPercentage(\uparrowRHDSmall)} & \multicolumn{3}{cm}{\apBevForty \bracketPercentage(\uparrowRHDSmall)}\\%\cline{2-13}\\[0.05cm]
                    && Easy & Mod & Hard & Easy & Mod & Hard & Easy & Mod & Hard & Easy & Mod & Hard\\
                    \myTopRule
                    & \ses$\rightarrowRHD$Vanilla            &$21.10$ & $15.48$ & $12.88$ & $28.58$ & $20.92$ & $17.83$ & $58.95$ & $43.99$ & $38.07$ & $64.60$ & $47.76$ &	$42.97$\\
                    %Depth-Aware \cite{brazil2019m3d} \\
                    %Depth-Guided \cite{ding2020learning} \\
                    Convolution &\ses$\rightarrowRHD$\LogPolar\!\cite{zwicke1983new}& $9.19$ & $6.77$ & $5.78$ & $16.39$ & $11.15$ & $9.80$ & $40.51$ & $27.62$ & $23.90$ & $45.66$ & $31.34$ & $25.80$\\
                    & \ses$\rightarrowRHD$Dilated\!\cite{yu2015multi} & $21.66$ & $15.49$ & $12.90$ & $30.22$ & $22.06$ & $19.01$ & $57.54$ & $43.12$ & $38.80$ & $63.29$ & $46.86$ & $42.42$\\ 
                    & \ses$\rightarrowRHD$DISCO\!\cite{sosnovik2021disco} & $20.21$ & $13.84$ & $11.46$ & $28.56$ & $19.38$ & $16.41$ & $55.22$ & $39.76$ & $35.37$ & $59.46$ & $43.16$ & $38.52$\\
                    \hline
                    Downscale
                    & $10\%\rightarrowRHD 5\%$& $24.24$	& $16.51$	& $14.43$	& $31.94$	& $22.86$	& $19.82$	& $60.64$	& $44.46$	& $40.02$	& $64.68$	& $49.30$	& $43.49$\\
                    $\alpha$ & $10\%\rightarrowRHD 20\%$& $22.19$	& $15.85$	& $13.48$	& $31.15$	& $23.01$	& $19.90$	& $61.24$	& $44.93$	& $40.22$	& $67.46$	& $50.10$	& $43.83$\\
                    \hline
                    BNP & \se$\rightarrowRHD$ Vanilla & $24.39$	& $16.20$	& $14.36$	& $32.43$	& $22.53$	& $19.70$	& $62.81$	& $46.14$	& $40.38$	& $67.87$	& $50.23$	& $44.08$\\  
                    \hline   
                    Scales
                    & $3\rightarrowRHD 1$& 
                    $23.20$	& $16.29$	& $13.63$	& $31.76$	& $23.23$	& $19.97$ & $61.90$	& $46.66$	& $40.61$	& $67.37$	& $50.31$	& $43.93$\\
                    & $3\rightarrowRHD 2$& 
                    $24.15$	& $16.48$	& $14.55$	& $32.42$	& $23.17$	& $20.07$	& $61.05$	& $46.34$	& $40.46$	& $67.36$	& $50.32$	& $44.07$\\
                    \hline
                    {---} & \textbf{\methodName{} (best)} & \best{24.63} & \best{16.54} & \best{14.52} & \best{32.60} &	\best{23.04} & \best{19.99} &	\best{61.00} & \best{46.00}	& \best{40.18} & \best{65.28} & \best{49.63} & \best{43.50}    \\
                    \myTopRule
                \end{tabular}
            }
        \end{table*}

    %============================================================================    
    %============================================================================
    \noIndentHeading{Ablation Studies.}\label{sec:results_ablation}
        \cref{tab:ablation} compares the modifications of our approach on \kitti{} \valOne{} cars based on the experimental settings of \cref{sec:experiments}.
        
        \noIndentHeading{(a) Floating or Integer Downscaling?}
        We next investigate the question that whether one should use floating or integer downscaling factors for \methodName.
        We vary the downscaling factors as $(1\!+\!2\alpha, 1\!+\!\alpha, 1)$ and therefore, our scaling factor $s\!=\!\left(\frac{1}{1+2\alpha}, \frac{1}{1+\alpha}, 1\right)$.
        We find that $\alpha$ of $10\%$ works the best. 
        We again bring up the dilated convolution (Dilated) results at this point because dilation is a \scaleEquivariant{} operation for integer downscaling factors \cite{worrall2019deep} $(\alpha\!=\!100\%, s\!=\!0.5)$.
        \cref{tab:ablation} results suggest that the downscaling factors should be floating numbers.
        %and not integers for the best performance.
        
        \noIndentHeading{(b) \se{} BNP.}
        As described in \cref{sec:steerable}, we ablate \methodName{} against the case when only convolutions are \se{} but BNP layers are not. 
        So, we place \MaxScale \cite{sosnovik2020sesn} immediately after every \ses{} convolution. 
        \cref{tab:ablation} shows that such a network performs slightly sub-optimal to our final model.

        \noIndentHeading{(c) Number of Scales.}
        We next ablate against the usage of Hermite scales. 
        Using three scales performs better than using only one scale especially on Mod and Hard objects, and slightly better than using two scales.

    %============================================================================
    %============================================================================
    \subsection{\waymo{} \val{} Monocular 3D Detection}
        We also benchmark our method on the \waymo{} dataset \cite{sun2020scalability} which has more variability than \kitti.
        \cref{tab:waymo_val} shows the results on \waymo{} \val{} split. 
        The results show that \methodName{} outperforms the baseline \gupNet{} \cite{lu2021geometry} on multiple levels and multiple thresholds. 
        The biggest gains are on the nearby objects which is consistent with \cref{tab:detection_results_kitti_test_cars,tab:detection_results_kitti_val1}.
        Interestingly, \methodName{} also outperforms PatchNet \cite{ma2020rethinking} and PCT \cite{wang2021progressive} without using depth. 
        Although the performance of \methodName{} lags \caddn \cite{reading2021categorical}, it is important to stress that \caddn~uses \lidar{} data in training, while \methodName{} is an \imageOnly{} method. 

    \begin{table*}[!tb]
        \caption{\small \textbf{Results on \waymo{} \val{} vehicles}. %using \apThreeD{} and~\aphThreeD. 
        [Key: \firstkey{Best}, \secondkey{Second Best}]}
        \label{tab:waymo_val}
        \centering
        \scalebox{\scaleFraction}{
            \footnotesize
            \rowcolors{4}{lightgray}{white}
            \setlength{\tabcolsep}{0.1cm}
            \begin{tabular}{m c m c m c m c m c | ccc m c | ccc m}
                \myTopRule
                & &  &  & \multicolumn{4}{cm}{\apThreeD{} \bracketPercentage(\uparrowRHDSmall)} & \multicolumn{4}{cm}{\aphThreeD~\bracketPercentage(\uparrowRHDSmall)} \\
                \cline{5-12}
                \multirow{-2}{*}{\iouThreeD} & \multirow{-2}{*}{Difficulty} & \multirow{-2}{*}{Method} & \multirow{-2}{*}{Extra} & All & 0-30 & 30-50 & 50-$\infty$ & All & 0-30 & 30-50 & 50-$\infty$ \\ 
                \myTopRule
                & & \caddn \cite{reading2021categorical} & \lidar & $5.03$ & $14.54$ & $1.47$ & $0.10$ & $4.99$ & $14.43$ & $1.45$ & $0.10$ \\
                & & \patchNet \cite{ma2020rethinking} in \cite{wang2021progressive} & Depth & $0.39$ & $1.67$ & $0.13$ & $0.03$ & $0.39$ & $1.63$ & $0.12$ & $0.03$ \\
                & & PCT \cite{wang2021progressive} & Depth & $0.89$ & $3.18$ & $0.27$ & $0.07$ & $0.88$ & $3.15$ & $0.27$ & $0.07$ \\
                $0.7$ & \levelOne & \mthreeDRPN \cite{brazil2019m3d} in \cite{reading2021categorical} & \mathDash{}& $0.35$ & $1.12$ & $0.18$ & $0.02$ & $0.34$ & $1.10$ & $0.18$ & $0.02$ \\
                & & \gupNet{} (Retrained)\cite{lu2021geometry} & \mathDash{}& \second{2.28} & \second{6.15} & \second{0.81} &	\first{0.03} & \second{2.27} & \second{6.11} & \second{0.80} &	\first{0.03}\\
                & & \textbf{\methodName{} (Ours)} & \mathDash{}& \first{2.69}	& \first{6.95} &	\first{0.99} &	\second{0.02} & \first{2.67}	& \first{6.90} &	\first{0.98} &	\second{0.02}\\
                \myTopRule
                & & \caddn \cite{reading2021categorical} & \lidar & $4.49$ & $14.50$ & $1.42$ & $0.09$ & $4.45$ & $14.38$ & $1.41$ & $0.09$ \\
                & & \patchNet \cite{ma2020rethinking} in \cite{wang2021progressive} & Depth & $0.38$ & $1.67$ & $0.13$ & $0.03$ & $0.36$ & $1.63$ & $0.11$ & $0.03$ \\
                &  & PCT \cite{wang2021progressive} & Depth & $0.66$ & $3.18$ & $0.27$ & $0.07$ & $0.66$ & $3.15$ & $0.26$ & $0.07$\\
                $0.7$ & \levelTwo & \mthreeDRPN \cite{brazil2019m3d} in \cite{reading2021categorical} & \mathDash{}& $0.33$ & {1.12} & {0.18} & \first{0.02} & $0.33$ & $1.10$ & $0.17$ & \first{0.02} \\
                & & \gupNet{} (Retrained)\cite{lu2021geometry} & \mathDash{}& \second{2.14} & \second{6.13} & \second{0.78} &	\first{0.02} & \second{2.12} & \second{6.08} & \second{0.77} &	\first{0.02} \\ 
                & & \textbf{\methodName{} (Ours)} & \mathDash{}& \first{2.52}	& \first{6.93} &	\first{0.95} &	\first{0.02} & \first{2.50}	& \first{6.87} &	\first{0.94} &	\first{0.02}\\
                \myTopRule
                & & \caddn \cite{reading2021categorical} & \lidar & $17.54$ & $45.00$ & $9.24$ & $0.64$ & $17.31$ & $44.46$ & $9.11$ & $0.62$ \\
                & & \patchNet \cite{ma2020rethinking} in \cite{wang2021progressive} & Depth & $2.92$ & $10.03$ & $1.09$ & $0.23$ & $2.74$ & $9.75$ & $0.96$ & $0.18$ \\
                & & PCT \cite{wang2021progressive} & Depth & $4.20$ & $14.70$ & $1.78$ & $0.39$ & $4.15$ & $14.54$ & $1.75$ & $0.39$\\
                $0.5$ & \levelOne & \mthreeDRPN \cite{brazil2019m3d} in \cite{reading2021categorical} & \mathDash{}& $3.79$ & $11.14$ & $2.16$ & \first{0.26} & $3.63$ & $10.70$ & $2.09$ & \second{0.21} \\
                & & \gupNet{} (Retrained)\cite{lu2021geometry} & \mathDash{}& \second{10.02} & \second{24.78} & \second{4.84} &	\second{0.22} & \second{9.94} & \second{24.59} & \second{4.78} &	\first{0.22} \\ 
                & & \textbf{\methodName{} (Ours)} & \mathDash{}& \first{10.98} & \first{26.85} & \first{5.13} &	$0.18$ & \first{10.89} & \first{26.64} & \first{5.08} & $0.18$  \\
                \myTopRule
                & & \caddn \cite{reading2021categorical} & \lidar & $16.51$ & $44.87$ & $8.99$ & $0.58$ & $16.28$ & $44.33$ & $8.86$ & $0.55$ \\
                & & \patchNet \cite{ma2020rethinking} in \cite{wang2021progressive} & Depth & $2.42$ & $10.01$ & $1.07$ & $0.22$ & $2.28$ & $9.73$ & $0.97$ & $0.16$ \\
                & & PCT \cite{wang2021progressive} & Depth & $4.03$ & $14.67$ & $1.74$ & $0.36$ & $4.15$ & $14.51$ & $1.71$ & $0.35$\\
                $0.5$ & \levelTwo  & \mthreeDRPN \cite{brazil2019m3d} in \cite{reading2021categorical} & \mathDash{}& $3.61$ & $11.12$ & $2.12$ & \first{0.24} & $3.46$ & $10.67$ & $2.04$ & \first{0.20} \\
                & & \gupNet{} (Retrained)\cite{lu2021geometry} & \mathDash{}& \second{9.39} & \second{24.69} & \second{4.67} &	\second{0.19} & \second{9.31} & \second{24.50} & \second{4.62} &	\second{0.19} \\ 
                & & \textbf{\methodName{} (Ours)} & \mathDash{}& \first{10.29} & \first{26.75} & \first{4.95} &	$0.16$ & \first{10.20} & \first{26.54} & \first{4.90} & $0.16$  \\
                \myTopRule
            \end{tabular}
        }
    \end{table*}

%============================================================================
%============================================================================
%============================================================================
\section{Conclusions}
    This paper studies the modeling error in monocular \threeD{} detection in detail and takes the first step towards convolutions equivariant to arbitrary \threeD{} translations in the projective manifold.
    Since the depth is the hardest to estimate for this task, this
    paper proposes \methodNameFull{} (\methodName) built with existing \scaleEquivariant{} steerable blocks. 
    As a result, \methodName{} is \equivariant{} to the depth translations in the projective \manifold{} whereas vanilla networks are not.
    The additional \depthEquivariance{} forces the \methodName{} to learn consistent depth estimates and therefore, \methodName{} achieves \sota{} detection results on \kitti{} and \waymo{} datasets in the \imageOnly{} category and performs competitively to methods using extra information.
    Moreover, \methodName{} works better than vanilla networks in cross-dataset evaluation.
    Future works include applying the idea to \pseudoLidar{} \cite{wang2019pseudo}, and monocular \threeD{} tracking.

%============================================================================
%============================================================================
%============================================================================
\clearpage
\bibliographystyle{splncs04}
\bibliography{references}

%============================================================================
%============================================================================
%============================================================================
\clearpage

\appendix % Restarts sections at 1

% Start appendices sections with A1, A2 instead of 1,2
\renewcommand{\thesection}{A\arabic{section}}

\begin{center}
\textbf{\Large \paperTitle\\ Supplementary Material\\}
\end{center}

%============================================================================
%============================================================================
%============================================================================
\section{Supportive Explanations}\label{sec:supplementary_explanation}
    
    We now add some explanations which we could not put in the main paper because of the space constraints.
    
    %===============================================================================
    \subsection{\Equivariance{} vs Augmentation} 
        \Equivariance{} adds suitable inductive bias to the backbone \cite{cohen2016group, dieleman2016exploiting} and is not learnt.
        Augmentation adds transformations to the input data during training or inference.
        
        \Equivariance{} and data augmentation have their own pros and cons.
        \Equivariance{} models the physics better, is mathematically principled and is so more agnostic to data distribution shift compared to the data augmentation.
        A downside of \equivariance{} compared to the augmentation is \equivariance{} requires mathematical modelling, may not always exist \cite{burns1992non}, is not so intuitive and generally requires more flops for inference.
        On the other hand, data augmentation is simple, intuitive and fast, but is not mathematically principled. 
        The choice between \equivariance{} and data augmentation is a withstanding question in machine learning \cite{gandikota2021training}.
    
    %============================================================================
    \subsection{Why do \twoD{} CNN detectors generalize?}
        We now try to understand why \twoD{} CNN detectors generalize well.
        Consider an image $\projectionOne(\pixU, \pixV)$ and $\mapping$ be the CNN.
        Let $\transformationMath_\translation$ denote the translation in the $(\pixU,\pixV)$ space.
        The \twoD{} translation \equivariance{} \cite{bronstein2021convolution, bronstein2021geometric, rath2020boosting} of the CNN means that 
        \begin{align}
            \mapping(\transformationMath_\translation\projectionOne(\pixU, \pixV)) &=
            \transformationMath_\translation\mapping(\projectionOne(\pixU, \pixV)) \nonumber \\
            \implies \mapping(\projectionOne(\pixU + \transU, \pixV + \transV)) &=
            \mapping(\projectionOne(\pixU, \pixV)) + (\transU, \transV)
            \label{eq:translation_equivariance}
        \end{align}
        where $(\transU, \transV)$ is the translation in the $(\pixU,\pixV)$ space.
        
        Assume the CNN predicts the object position in the image as $(\pixUTwo, \pixVTwo)$. 
        Then, we write
            \begin{align}
                \mapping(\projectionOne(\pixU, \pixV)) &= (\predictionU, \predictionV)
            \end{align}
        
        Now, we want the CNN to predict the output the position of the same object translated by $(\transU, \transV)$.
        The new image is thus $\projectionOne(\pixU + \transU, \pixV + \transV)$.
        The CNN easily predicts the translated position of the object because all CNN is to do is to invoke its \twoD{} translation \equivariance{} of \cref{eq:translation_equivariance}, and translate the previous prediction by the same amount.
        In other words,
        \begin{align*}
            \mapping(\projectionOne(\pixU + \transU, \pixV + \transV)) &=    
            \mapping(\projectionOne(\pixU, \pixV)) + (\transU, \transV)\\
            &=  (\predictionU, \predictionV) + (\transU, \transV) \\
            &=  (\predictionU + \transU, \predictionV + \transV)
        \end{align*}
        Intuitively, \equivariance{} is a disentaglement method. 
        The \twoD{} translation \equivariance{} disentangles the \twoD{} translations $(\transU, \transV)$ from the original image $\projectionOne$ and therefore, the network generalizes to unseen \twoD{} translations.

    %============================================================================
    \subsection{Existence and Non-existence of \Equivariance{}}
        The result from \cite{burns1992non} says that generic projective \equivariance{} does not exist in particular with rotation transformations.
        We now show an example of when the \equivariance{} exists and does not exist in the projective manifold in \cref{fig:eqv_exists,fig:non_existence} respectively.
    
    \begin{figure}[!tb]
        \centering
        \begin{tikzpicture}[scale=0.28, every node/.style={scale=0.50}, every edge/.style={scale=0.50}]
\tikzset{vertex/.style = {shape=circle, draw=black!70, line width=0.06em, minimum size=1.4em}}
\tikzset{edge/.style = {-{Triangle[angle=60:.06cm 1]},> = latex'}}

    \input{images/teaser_common}

    % =============== Plane we are taking about ====================
    \draw [draw=black!100, line width=0.04em, fill=gray!50, opacity=0.7]    (6.75, 9.25) rectangle (8.25, 10.75) node[]{};
    \node [inner sep=1pt, scale= 1.5,text width=3cm] at (14.5, 8.0)  {Patch Plane\\ $m\varX+n\varY+o\varZ+p= 0$};

    %================ Circles ===============
    \draw[draw=rayShade, fill=rayShade, thick](7.5,10.0) circle (0.35) node[scale= 1.25]{};

    %============ Left Shooting rays (Top) ================= 
    \draw [draw=rayShade, line width=0.1em, shorten <=0.5pt, shorten >=0.5pt, >=stealth]
       (7.5,10.0) node[]{}
    -- (4.65,6.2) node[]{};
    
    %============ Plane pointing ================= 
    \draw [draw=black!100, line width=0.05em, shorten <=0.5pt, shorten >=0.5pt, >=stealth]
       (7.8,9.5) node[]{}
    -- (10.3,8.0) node[]{};

    %============ Names if any =================
    \node [inner sep=1pt, scale= 1.75] at (8.5, 4.8)   {$\projectionOne(\pixU, \pixV)$};
    \node [inner sep=1pt, scale= 1.75] at (5.0, -1.1) {$\projectionTwo(\pixUTwo, \pixVTwo)$};
    \node [inner sep=1pt, scale= 1.25] at (10.1,10.0)   {$(\posX,\posY,\posZ)$};
    
    %================  Principal Points ================ 
    %(2.2, 3.2)
    \draw[draw=black, fill=black, thick](4.7, 5.5) circle (0.12) node[scale= 1.1]{~\quad\quad\quad$(\!\ppointU,\!\ppointV\!)$};
    %(-1.65, -2.6)
    \draw[draw=black, fill=black, thick](0.55, -0.5) circle (0.12) node[scale= 1.1]{~\quad\quad\quad$(\!\ppointU,\!\ppointV\!)$};

    %================  Focal Lengths ================ 
    \draw [{|}-{|}, draw=black!60, line width=0.05em, shorten <=0.5pt, shorten >=0.5pt, >=stealth]
       (2.1,4.8) node[]{}
    -- (0.45,2.8) node[pos=0.5, scale= 1.75, align= center]{\color{black}{$\focal$}~~~~\\};

    \draw [{|}-{|}, draw=black!60, line width=0.05em, shorten <=0.5pt, shorten >=0.5pt, >=stealth]
       (-2.5,-1.2) node[]{}
    -- (-4.05,-3.2) node[pos=0.5, scale= 1.75, align= center]{\color{black}{$\focal$}~~~~\\};

% (1.5,2.0)
% (-3,-4.0)

    % \draw [draw=black, line width=0.1em, shorten <=0.5pt, shorten >=0.5pt, >=stealth]
    %   (-1.65,-2.6) node[]{}
    % -- (-5.01,-2.68) node[pos=0.5, scale= 1.75, align= center]{\color{black}{$\focal$}~~~~\\};
    
    %================ Axis Convention ===================
    \draw[draw=axisShadeDark, fill=axisShadeDark, thick](-1.65,10.0) circle (0.08) node[]{};
    
    \draw [-{Triangle[angle=60:.1cm 1]}, draw=axisShadeDark, line width=0.05em, shorten <=0.5pt, shorten >=0.5pt, >=stealth]
           (-1.65,10.0) node[]{}
        -- (0.3,10.0) node[scale= 1.75]{~~$x$};
        
    \draw [-{Triangle[angle=60:.1cm 1]}, draw=axisShadeDark, line width=0.05em, shorten <=0.5pt, shorten >=0.5pt, >=stealth]
           (-1.65,10.0) node[]{}
        -- (-1.65,8.35) node[text width=1cm,align=center,scale= 1.75]{~\\$y$};
    
    \draw [-{Triangle[angle=60:.1cm 1]}, draw=axisShadeDark, line width=0.05em, shorten <=0.5pt, shorten >=0.5pt, >=stealth]
           (-1.65,10.0) node[]{}
        -- (-0.75,11.0) node[scale= 1.75]{~~$z$};
    
\end{tikzpicture}
        \caption{
            \textbf{\Equivariance{} exists} for the \plane{} when there is depth translation of the ego camera. 
            Downscaling converts image $\projectionOne$ to image $\projectionTwo$.
        }
        \label{fig:eqv_exists}
    \end{figure}
    \begin{figure}[!tb]
        \centering
        \input{images/translational_equiv_not_exist}
        \caption{
            \textbf{Example of non-existence of \equivariance{} }\cite{burns1992non} when there is $180\degree$ rotation of the ego camera. 
            No transformation can convert image $\projectionOne$ to image $\projectionTwo$.
        }
        \label{fig:non_existence}
    \end{figure}

    %============================================================================
    \subsection{Why do not Monocular 3D CNN detectors generalize?}
        Monocular \threeD{} CNN detectors do not generalize well because they are not \equivariant{} to arbitrary \threeD{} translations in the projective manifold.
        To show this, let $\pointCloudOne(\varX, \varY, \varZ)$ denote a \threeD{} point cloud.
        The monocular detection network $\mapping$ operates on the projection $\projectionOne(\pixU, \pixV)$ of this point cloud $\pointCloudOne$ to output the position $(\predictionX, \predictionY, \predictionZ)$ as
        \begin{align*}
            \mapping( \projectionOperator \pointCloudOne(\varX, \varY, \varZ)) &= (\predictionX, \predictionY, \predictionZ) \\
            \implies \mapping(   \projectionOne(\pixU, \pixV) ) &= (\predictionX, \predictionY, \predictionZ),   
        \end{align*}
        where $\projectionOperator$ denotes the projection operator.
        We translate this point cloud by an arbitrary \threeD{} translation of $(\transX,\transY,\transZ)$ to obtain the new point cloud  $\pointCloudOne(\varX + \transX, \varY + \transY, \varZ + \transZ)$. 
        Then, we again ask the monocular detector $\mapping$ to do prediction over the translated point cloud.
        However, we find that
        \begin{align*}
            \mapping( \projectionOperator \pointCloudOne(\varX + \transX, \varY + \transY, \varZ + \transZ)) &\ne \mapping(\projectionOne(\pixU + \projectionOperator (\transX, \transY,\transZ), \pixV + \projectionOperator (\transX, \transY, \transZ))) \\
            &= \mapping(\projectionOne(\pixU, \pixV)) + \projectionOperator (\transX, \transY, \transZ)  \\
            \implies \mapping( \projectionOperator \pointCloudOne(\varX + \transX, \varY + \transY, \varZ + \transZ)) &\ne \mapping(\projectionOperator \pointCloudOne(\varX, \varY, \varZ)) + \projectionOperator (\transX, \transY, \transZ)
        \end{align*}
        In other words, the projection operator $\projectionOperator$ does not distribute over the point cloud $\pointCloudOne$ and arbitrary \threeD{} translation of $(\transX,\transY,\transZ)$.
        Hence, if the network $\mapping$ is a vanilla CNN (existing monocular backbone), it can no longer invoke its \twoD{} translation \equivariance{} of  \cref{eq:translation_equivariance} to get the new \threeD{} coordinates $(\predictionX + \transX, \predictionY + \transY, \predictionZ + \transZ)$.
        
        Note that the \lidar{} based \threeD{} detectors with \threeD{} convolutions do not suffer from this problem because they do not involve any projection operator $\projectionOperator$.
        Thus, this problem exists only in monocular \threeD{} detection. 
        This makes monocular \threeD{} detection different from \twoD{} and \lidar{} based \threeD{} object detection.

    %============================================================================
    \subsection{Overview of \cref{th:projective_bigboss}}
    
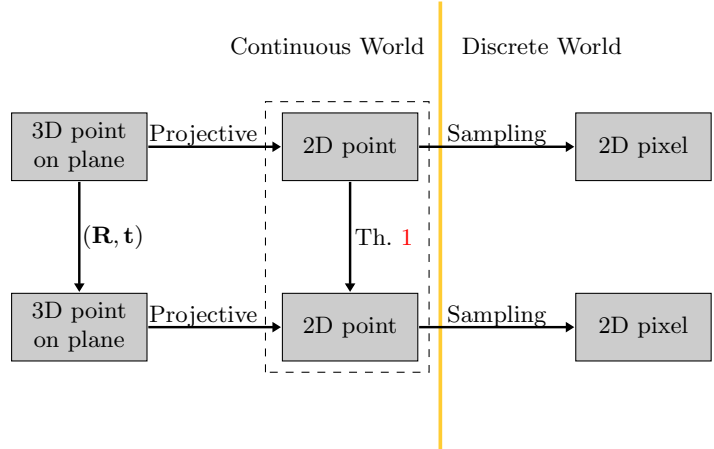
\begin{figure}[!htb]
            \centering
            \begin{tikzpicture}[scale=0.30, every node/.style={scale=0.5}, every edge/.style={scale=0.50}]
\tikzset{vertex/.style = {shape=circle, draw=black!70, line width=0.06em, minimum size=1.4em}}
\tikzset{edge/.style = {-{Triangle[angle=60:.06cm 1]},> = latex'}}

%================ 3D points =========================
\draw[black,fill=black!20] (-2, 0) rectangle (4, 3);
\node [inner sep=1pt, scale= 2, align= center] at (1, 1.5)  {\threeD{} point\\on plane};

\draw[black,fill=black!20] (-2, 8) rectangle (4, 11);
\node [inner sep=1pt, scale= 2, align= center] at (1, 9.5)  {\threeD{} point\\on plane};

% Arrow
\draw [-{Triangle[angle=60:.1cm 1]},draw=black, line width=0.1em, shorten <=0.5pt, shorten >=0.5pt, >=stealth]
      (1,8) node[]{}
    -- (1,3) node[]{};
\node [inner sep=1pt, scale= 2] at (2.5, 5.5)  {$(\rotation,\translation)$};

%================ Arrows connecting 3D to 2D =========================
\draw [-{Triangle[angle=60:.1cm 1]},draw=black, line width=0.1em, shorten <=0.5pt, shorten >=0.5pt, >=stealth]
      (4,1.5) node[]{}
    -- (10,1.5) node[]{};
\node [inner sep=1pt, scale= 2] at (6.5, 2)  {Projective};

\draw [-{Triangle[angle=60:.1cm 1]},draw=black, line width=0.1em, shorten <=0.5pt, shorten >=0.5pt, >=stealth]
      (4,9.5) node[]{}
    -- (10,9.5) node[]{};
\node [inner sep=1pt, scale= 2] at (6.5, 10)  {Projective};

%================ Continuous 2D points =========================
\draw[black,fill=black!20] (10, 0) rectangle (16, 3);
\node [inner sep=1pt, scale= 2] at (13, 1.5)  {\twoD{} point};

\draw[black,fill=black!20] (10, 8) rectangle (16, 11);
\node [inner sep=1pt, scale= 2] at (13, 9.5)  {\twoD{} point};

% Arrow
\draw [-{Triangle[angle=60:.1cm 1]},draw=black, line width=0.1em, shorten <=0.5pt, shorten >=0.5pt, >=stealth]
      (13,8) node[]{}
    -- (13,3) node[]{};
\node [inner sep=1pt, scale= 2] at (14.5, 5.5)  {Th. \ref{th:projective_bigboss}};

% Dashed rectangle around
\draw[black,dashed] (9.25, -0.5) rectangle (16.5, 11.5);
% \node [inner sep=1pt, scale= 2] at (13, -3.5)  {\ses{} with \cref{th:projective_scaled}};

% Separation between continuous and discrete world
\draw [draw=rayShade, line width=0.15em, shorten <=0.5pt, shorten >=0.5pt, >=stealth]
      (17,16) node[]{}
    -- (17,-4) node[]{};
\node [inner sep=1pt, scale= 2] at (12, 14)  {Continuous World};
\node [inner sep=1pt, scale= 2] at (21.5, 14)  {Discrete World};

%================ Arrows connecting continuous 2D to 2D pixel =========================
\draw [-{Triangle[angle=60:.1cm 1]},draw=black, line width=0.1em, shorten <=0.5pt, shorten >=0.5pt, >=stealth]
      (16,1.5) node[]{}
    -- (23,1.5) node[]{};
% \draw[black, fill= white] (18, 0) rectangle (22, 3);
\node [inner sep=1pt, scale= 2] at (19.5, 2)  {Sampling};

\draw [-{Triangle[angle=60:.1cm 1]},draw=black, line width=0.1em, shorten <=0.5pt, shorten >=0.5pt, >=stealth]
      (16,9.5) node[]{}
    -- (23,9.5) node[]{};
\node [inner sep=1pt, scale= 2] at (19.5, 10)  {Sampling};

%================ Discrete 2D points =========================
\draw[black,fill=black!20] (23, 0) rectangle (29, 3);
\node [inner sep=1pt, scale= 2] at (26, 1.5)  {\twoD{} pixel};

\draw[black,fill=black!20] (23, 8) rectangle (29, 11);
\node [inner sep=1pt, scale= 2] at (26, 9.5)  {\twoD{} pixel};

\end{tikzpicture}
            \caption{
                \textbf{Overview} of \cref{th:projective_bigboss} (Example 13.2 from \cite{hartley2003multiple}), which links the planarity and projective transformations in the continuous world.
            }
            \label{fig:high_level}
        \end{figure}

        We now pictorially provide the overview of \cref{th:projective_bigboss} (Example 13.2 from \cite{hartley2003multiple}), which links the planarity and projective transformations in the continuous world in \cref{fig:high_level}.
    
    %============================================================================
    \subsection{Approximation of \cref{th:projective_scaled}}\label{sec:approximation_proof}
        We now give the approximation under which \cref{th:projective_scaled} is valid.
        We assume that the ego camera does not undergo any rotation. Hence, we substitute $\rotation=\identity$ in \cref{eq:bigboss} to get
            \begin{align}
                \projectionOne(\pixU-\ppointU, \pixV-\ppointV)
                &= \projectionTwo\left(\focal\dfrac{\left(1\mySign\transX\frac{m}{p}\right)\pixUMinusPointU\mySign\transX\frac{n}{p}\pixVMinusPointV 
                \mySign\transX\frac{o}{p}\focal}{\transZ\frac{m}{p}\pixUMinusPointU \mySign\transZ\frac{n}{p}\pixVMinusPointV+\left(1\mySign\transZ\frac{o}{p}
                \right)\focal} ,
                \right. \nonumber \\
                &\quad\quad\left.\focal\dfrac
                {\transY\frac{m}{p}\pixUMinusPointU +  \left(1\mySign\transY\frac{n}{p}\right)\pixVMinusPointV \mySign \transY\frac{o}{p}\focal
                }
                {\transZ\frac{m}{p}\pixUMinusPointU 
                + \transZ\frac{n}{p} \pixVMinusPointV+\left(1\mySign\transZ\frac{o}{p}\right)\focal} \right).
                \label{eq:projective_no_rotation}
            \end{align}
        Next, we use the assumption that the ego vehicle moves in the $z$-direction as in \cite{brazil2020kinematic},  \thatIs, substitute $\transX\!=\!\transY\!=\!0$ to get
        \begin{align}
            \projectionOne(\pixU\!-\!\ppointU, \pixV\!-\!\ppointV) %\nonumber \\
            &= \projectionTwo\left(\dfrac{\pixU-\ppointU}{\frac{\transZ}{\focal}\frac{m}{p}\pixUMinusPointU \mySign\frac{\transZ}{\focal}\frac{n}{p}\pixVMinusPointV +\left(1\mySign\transZ\frac{o}{p}\right)},  \right. \nonumber \\
            &\quad\quad\quad~\left.\dfrac{\pixV-\ppointV}{\frac{\transZ}{\focal}\frac{m}{p}\pixUMinusPointU \mySign\frac{\transZ}{f}\frac{n}{p}\pixVMinusPointV +\left(1\mySign\transZ\frac{o}{p}\right)} \right).
            \label{eq:projective_only_z}
        \end{align}

        The \plane{} is $mx + ny + oz + p= 0$. 
        We consider the planes in the front of camera. 
        Without loss of generality, consider $p < 0$ and $o > 0$.
    
        We first write the denominator $D$ of RHS term in \cref{eq:projective_only_z} as
        \begin{align}
            D &= \frac{\transZ}{\focal}\frac{m}{p}\pixUMinusPointU \mySign\frac{\transZ}{f}\frac{n}{p}\pixVMinusPointV +\left(1\mySign\transZ\frac{o}{p}\right) \nonumber \\
            &= 1 + \frac{\transZ}{p} \left(\frac{m}{f}\pixUMinusPointU + \frac{n}{f}\pixVMinusPointV + o \right) \nonumber
        \end{align}
        Because we considered \plane s in front of the camera, $p < 0$. 
        Also consider $\transZ < 0$, which implies $\transZ / p > 0$.
        Now, we bound the term in the parantheses of the above equation as
        \begin{align*}
            D &\le 1 +  \frac{\transZ}{p} \norm{\frac{m}{f}\pixUMinusPointU + \frac{n}{f}\pixVMinusPointV + o } \\
              &\le 1 + \frac{\transZ}{p} \left( \norm{\frac{m}{f}\pixUMinusPointU} + \norm{\frac{n}{f}\pixVMinusPointV} + \norm{o} \right) \quad \text{by Triangle inequality}\\
              &\le 1 + \frac{\transZ}{p} \left(\frac{\norm{m}}{f}\frac{W}{2} + \frac{\norm{n}}{f}\frac{H}{2} + o \right), \pixUMinusPointU \le \frac{W}{2}, \pixVMinusPointV \le \frac{H}{2}, \norm{o}= o\\
              &\le 1 + \frac{\transZ}{p} \left(\frac{\norm{m}}{f}\frac{W}{2} + \frac{\norm{n}}{f}\frac{W}{2} + o \right), H \le W\\
              &\le 1 + \frac{\transZ}{p} \left( \frac{(\norm{m} + \norm{n}) W}{2f} + o \right) , \quad 
        \end{align*}
        
\begin{figure}[!htb]
            \centering
            \begin{tikzpicture}[scale=0.36, every node/.style={scale=0.6}, every edge/.style={scale=0.60}]
\tikzset{vertex/.style = {shape=circle, draw=black!70, line width=0.06em, minimum size=1.4em}}
\tikzset{edge/.style = {-{Triangle[angle=60:.06cm 1]},> = latex'}}

%================ Camera =========================
\draw[black,fill=black!50] (-1.6,10.4) rectangle (-0.4,9.6);
\coordinate (c10) at (-0.4,10.0);
\coordinate (c11) at (0.18,9.6);
\coordinate (c12) at (0.18,10.4);
\filldraw[draw=black, fill=gray!20] (c10) -- (c11) -- (c12) -- cycle;

%============ Shooting rays ================= 
\draw [draw=rayShade, line width=0.1em, shorten <=0.5pt, shorten >=0.5pt, >=stealth]
       (0.2,10.0) node[]{}
    -- (8.6,10.0) node[]{};

%============ Names if any =================
% \node [inner sep=1pt, scale= 2] at (8.5, 13)  {Perfect Parallel};

\draw [, draw=axisShadeDark, line width=0.1em]
       (8.5,7.5) node[]{}
    -- (8.5,12.5) node[text width=1cm,align=center,scale= 1.75]{};

\draw [dashed, draw=axisShadeDark, line width=0.07em, shorten <=0.5pt, shorten >=0.5pt, >=stealth]
       (8.0,8.0) node[]{}
    -- (9.0,12.0) node[text width=1cm,align=center,scale= 1.75]{};

\draw [draw=my_magenta, line width=0.05em, shorten <=0.5pt, shorten >=0.5pt, >=stealth]
       (8.45,11.4) node[]{}
    -- (9.0,11.4) node[text width=1cm,align=center,scale= 1.75]{~~~~~\textcolor{my_magenta}{$\theta$}};

\draw [dashed, draw=axisShadeDark, line width=0.07em, shorten <=0.5pt, shorten >=0.5pt, >=stealth]
       (9.0,8.0) node[]{}
    -- (8.0,12.0) node[text width=1cm,align=center,scale= 1.75]{};

%================ Axis Convention ===================
    \draw[draw=axisShadeDark, fill=axisShadeDark, thick](-6.65,10.0) circle (0.08) node[]{};
    
    \draw [-{Triangle[angle=60:.1cm 1]}, draw=axisShadeDark, line width=0.05em, shorten <=0.5pt, shorten >=0.5pt, >=stealth]
           (-6.65,10.0) node[]{}
        -- (-4.7,10.0) node[scale= 1.75]{~~$z$};
        
    \draw [-{Triangle[angle=60:.1cm 1]}, draw=axisShadeDark, line width=0.05em, shorten <=0.5pt, shorten >=0.5pt, >=stealth]
           (-6.65,10.0) node[]{}
        -- (-6.65,8.35) node[text width=1cm,align=center,scale= 1.75]{~\\$y$};
    
    % \draw [-{Triangle[angle=60:.1cm 1]}, draw=axisShadeDark, line width=0.05em, shorten <=0.5pt, shorten >=0.5pt, >=stealth]
    %       (-1.65,10.0) node[]{}
    %     -- (-0.75,11.0) node[scale= 1.75]{~~$z$};

\end{tikzpicture}
            \caption{
                \textbf{Approximation of \cref{th:projective_scaled}}. Bold shows the \plane{}  parallel to the image plane. The dotted line shows the approximated \plane{}.
            }
            \label{fig:assumption}
        \end{figure}
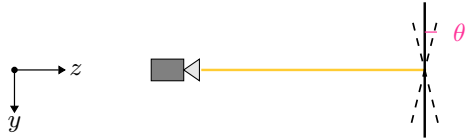

        If the coefficients of the patch plane $m,n,o$, its width $W$ and focal length $f$ follow the relationship $\frac{(\norm{m} + \norm{n}) W}{2f} << o$, the \plane{} is ``approximately''  parallel to the image plane.
        Then, a few quantities can be ignored in the denominator $D$ to get
        \begin{align}
            \text{D} &\approx 1 + \transZ\frac{o}{p} 
        \end{align}
        Therefore, the RHS of \cref{eq:projective_only_z} gets simplified and we obtain
        \begin{align}
            &\transformationMath_\scaleNotation :\projectionOne(\pixU-\ppointU, \pixV-\ppointV)
            \approx \projectionTwo\left(\dfrac{\pixU-\ppointU}{1\mySign\transZ\frac{o}{p}}, \dfrac{\pixV-\ppointV}{1\mySign\transZ\frac{o}{p}}\right)
        \end{align}
        An immediate benefit of using the approximation is \cref{eq:projective_scaled} does not depend on the distance of the \plane{} from the camera. 
        This is different from wide-angle camera assumption, where the ego camera is assumed to be far from the \plane{}.
        Moreover, \plane s need not be perfectly aligned with the image plane for \cref{eq:projective_scaled}. Even small enough perturbed \plane s work.
        We next show the approximation in the \cref{fig:assumption} with $\theta$ denoting the deviation from the perfect parallel plane.
        The deviation $\theta$ is about $3$ degrees for the \kitti{} dataset while it is $6$ degrees for the \waymo{} dataset.
        
        \forExample{} The following are valid \plane s for \kitti{} images whose focal length $\focal = 707$ and width $W = 1242$.
        \begin{align}
            -0.05x + 0.05y + z &= 30 \nonumber \\
            0.05x - 0.05y + z &= 30 
        \end{align}
        The following are valid \plane s for \waymo{} images whose focal length $\focal = 2059$ and width $W = 1920$.
        \begin{align}
            -0.1x + 0.1y + z &= 30 \nonumber \\
            0.1x - 0.1y + z &= 30 
        \end{align}
        
        Although the assumption is slightly restrictive, we believe our method shows improvements on both \kitti{} and \waymo{} datasets because the car patches are approximately parallel to image planes and also because the depth remains the hardest parameter to estimate \cite{ma2021delving}.

    %===============================================================================
    \subsection{\ScaleEquivariance{} of \ses{} Convolution for Images}\label{sec:scale_eqv_proof}
        \cite{sosnovik2020sesn} derive the \scaleEquivariance{} of \ses{} convolution for a \oneD{} signal.
        We simply follow on their footsteps to get the \scaleEquivariance{} of \ses{} convolution for a \twoD{} image $\projectionOne(\pixU,\pixV)$ for the sake of completeness. 
        Let the scaling of the image $\projectionOne$ be $\scaleNotation$. 
        Let $\conv$ denote the standard vanilla convolution and $\filter$ denote the convolution filter.
        Then, the convolution of the downscaled image $\transformationMath_{\scaleNotation}(\projectionOne)$ with the filter $\filter$ is given by
        \begin{align}
            &\left[\transformationMath_{\scaleNotation}(\projectionOne)\conv\filter \right](\pixU,\pixV) \nonumber\\
            &= \int\int\projectionOne\left(\frac{\pixUTwo}{\scaleNotation}, \frac{\pixVTwo}{\scaleNotation}\right)\filter(\pixUTwo-\pixU, \pixVTwo-\pixV) d\pixUTwo d\pixVTwo \nonumber\\
            &= \scaleNotation^2 \int\int\projectionOne\left(\frac{\pixUTwo}{\scaleNotation}, \frac{\pixVTwo}{\scaleNotation}\right)\filter\left( \scaleNotation\frac{\pixUTwo-\pixU}{\scaleNotation}, \scaleNotation\frac{\pixVTwo-\pixV}{\scaleNotation}\right) d\left(\frac{\pixUTwo}{\scaleNotation}\right) d\left(\frac{\pixVTwo}{\scaleNotation} \right)\nonumber\\
            &= \scaleNotation^2 \int\int\projectionOne\left(\frac{\pixUTwo}{\scaleNotation}, \frac{\pixVTwo}{\scaleNotation}\right)\transformationMath_{\scaleNotation^{-1}}\left[\filter\left( \frac{\pixUTwo-\pixU}{\scaleNotation}, \frac{\pixVTwo-\pixV}{\scaleNotation}\right)\right] d\left(\frac{\pixUTwo}{\scaleNotation}\right) d\left(\frac{\pixVTwo}{\scaleNotation} \right)\nonumber\\
            &= \scaleNotation^2 \int\int\projectionOne\left(\frac{\pixUTwo}{\scaleNotation}, \frac{\pixVTwo}{\scaleNotation}\right)\transformationMath_{\scaleNotation^{-1}}\left[\filter\left( \frac{\pixUTwo}{\scaleNotation}-\frac{\pixU}{\scaleNotation}, \frac{\pixVTwo}{\scaleNotation}-\frac{\pixV}{\scaleNotation}\right)\right] d\left(\frac{\pixUTwo}{\scaleNotation}\right) d\left(\frac{\pixVTwo}{\scaleNotation} \right)\nonumber\\
            &= \scaleNotation^2 \left[\projectionOne \conv \transformationMath_{\scaleNotation^{-1}}(\filter)\right]\left(\frac{\pixU}{\scaleNotation},\frac{\pixV}{\scaleNotation}\right) \nonumber\\
            &= \scaleNotation^2 \transformationMath_{\scaleNotation}\left[\projectionOne \conv\transformationMath_{\scaleNotation^{-1}}(\filter)\right](\pixU,\pixV).
            \label{eq:scale_eq_orig}
        \end{align}

        Next, \cite{sosnovik2020sesn} re-parametrize the SES filters by writing $\filter_\sigma(\pixU,\pixV) = \frac{1}{\sigma^2}\filter\left(\frac{\pixU}{\sigma},\frac{\pixV}{\sigma}\right)$. 
        Substituting in \cref{eq:scale_eq_orig}, we get
        \begin{align}
            \left[\transformationMath_{\scaleNotation}(\projectionOne)\conv\filter_\sigma \right](\pixU,\pixV)
            &= \scaleNotation^2 \transformationMath_{\scaleNotation}\left[\projectionOne \conv\transformationMath_{\scaleNotation^{-1}}(\filter_\sigma)\right](\pixU,\pixV)
            \label{eq:scale_eq_interm}
        \end{align}
        
        Moreover, the re-parametrized filters are separable \cite{sosnovik2020sesn} by construction and so, one can write
        \begin{align}
            \filter_\sigma(\pixU,\pixV) &= \filter_\sigma(\pixU)\filter_\sigma(\pixV).
        \end{align}
        
        The re-parametrization and separability leads to the important property that
        \begin{align}
            \transformationMath_{\scaleNotation^{-1}}\left( \filter_\sigma(\pixU,\pixV) \right)
            &= \transformationMath_{\scaleNotation^{-1}}\left( \filter_\sigma(\pixU)\filter_\sigma(\pixV) \right) \nonumber\\
            &= \transformationMath_{\scaleNotation^{-1}}\left( \filter_\sigma(\pixU)\right)\transformationMath_{\scaleNotation^{-1}}\left( \filter_\sigma(\pixV) \right) \nonumber\\
            &= \scaleNotation^{-2} \filter_{\scaleNotation^{-1} \sigma}(\pixU)\filter_{\scaleNotation^{-1} \sigma}(\pixV) \nonumber \\
            &= \scaleNotation^{-2} \filter_{\scaleNotation^{-1} \sigma}(\pixU, \pixV).
        \end{align}
        
        Substituting above in the RHS of \cref{eq:scale_eq_interm}, we get
        \begin{align}
            \left[\transformationMath_{\scaleNotation}(\projectionOne)\conv\filter_\sigma \right](\pixU,\pixV)
            &= \scaleNotation^2 \transformationMath_{\scaleNotation}\left[\projectionOne \conv \scaleNotation^{-2} \filter_{\scaleNotation^{-1} \sigma}\right](\pixU,\pixV) \nonumber \\
            \implies \left[\transformationMath_{\scaleNotation}(\projectionOne)\conv\filter_\sigma \right](\pixU,\pixV) &= \transformationMath_{\scaleNotation}\left[\projectionOne \conv \filter_{\scaleNotation^{-1} \sigma}\right](\pixU,\pixV),
            \label{eq:scale_eq_final}
        \end{align}
        which is a cleaner form of \cref{eq:scale_eq_orig}.
        \cref{eq:scale_eq_final} says that convolving the downscaled image with a filter is same as the
        downscaling the result of convolving the image with the upscaled filter \cite{sosnovik2020sesn}. 
        This additional constraint regularizes the scale (depth) predictions for the image, leading to better generalization.
        
        % From \cref{th:projective_scaled}, the scale is a linear function of depth translation \thatIs{} $\scaleNotation = 1\!+\!\transZ\frac{o}{p}$.
        % Hence, we have,
        % \begin{align}
        %     h\left(\frac{\pixU}{1\!+\!\transZ\frac{o}{p}},\frac{\pixV}{1\!+\!\transZ\frac{o}{p}}\right) \conv \filter_\sigma(\pixU,\pixV) 
        %     &= \transformationMath_{1\!+\!\transZ\frac{o}{p}}\left[\projectionOne (\pixU,\pixV) \conv \filter_{\left(1\!+\!\transZ\frac{o}{p}\right)^{-1} \sigma}(\pixU,\pixV)\right]
        % \end{align}

    %===============================================================================
    \subsection{Why does \methodName{} generalize better compared to CNN backbone?}\label{sec:why_better_generalize}
        \methodName{} models the physics better compared to the CNN backbone. 
        CNN generalizes better for \twoD{} detection because of the \twoD{} translation \equivariance{} in the Euclidean manifold.
        However, monocular \threeD{} detection does not belong to the Euclidean manifold but is a task of the projective manifold.
        Modeling translation \equivariance{} in the correct manifold improves generalization.
        For monocular \threeD{} detection, we take the first step towards the general \threeD{} translation \equivariance{} by embedding \equivariance{} to depth translations. 
        The \threeD{} \depthEquivariance{} in \methodName{} uses \cref{eq:scale_eq_interm} and thus imposes an additional constraint on the feature maps.
        This additional constraint results in consistent depth estimates from the current image and a virtual image (obtained by translating the ego camera), and therefore, better generalization than CNNs. 
        On the other hand, CNNs, by design, do not constrain the depth estimates from the current image and a virtual image (obtained by translating the ego camera), and thus, their depth estimates are entirely data-driven. 
        % are no longer free or data-driven as in CNNs but are tied together by \cref{eq:scale_eq_interm} in . 
        % This additional \equivariance{} of \cref{eq:scale_eq_interm} forces the network to output 
        % consistent depth estimates (had the ego camera moved), enhancing generalization.

    %============================================================================
    \subsection{Why not Fixed Scale Assumption?}
        We now answer the question of keeping the fixed scale assumption. If we assume fixed scale assumption, then vanilla convolutional layers have the right equivariance. 
        However, we do not keep this assumption because the ego camera translates along the depth in driving scenes and also, because the depth is the hardest parameter to estimate \cite{ma2021delving} for monocular detection. 
        So, zero depth translation or fixed scale assumption is always violated.

    %============================================================================
    \subsection{Comparisons with Other Methods}
        We now list out the differences between different convolutions and monocular detection methods in \cref{tab:compare_methods}.
        Kinematic3D \cite{brazil2020kinematic} does not constrain the output at feature map level, but at system level using Kalman Filters.
        The closest to our method is the Dilated CNN (DCNN) \cite{yu2015multi}. 
        We show in \cref{tab:kitti_compare_dilation} that \methodName{} outperforms Dilated CNN.

        \begin{table}[!tb]
            \caption{\textbf{Comparison of Methods} on the basis of inputs, convolution kernels, outputs and whether output are scale-constrained.
            }
            \label{tab:compare_methods}
            \centering
            \scalebox{\scaleFraction}{
                \footnotesize
                \rowcolors{3}{lightgray}{white}
                \setlength\tabcolsep{0.1cm}
                \begin{tabular}{ml m c m ccc  m cccm}
                    \myTopRule
                    \addlinespace[0.01cm]
                    \multirow{2}{*}{Method} & Input & \#Conv & \multirow{2}{*}{Output} & Output Constrained\\
                     & Frame & Kernel & & for Scales?\\
                    \myTopRule
                    Vanilla CNN & 1 & 1 & 4D & \xmark \\
                    Depth-Aware \cite{brazil2019m3d} & 1 & $>1$ & \fourD & \xmark \\
                    Dilated CNN \cite{yu2015multi} & 1 & $>1$ & \fiveD & Integer \cite{worrall2019deep}\\
                    \textbf{\methodName{}} & 1 & $>1$ & \fiveD & Float\\
                    \hline
                    Depth-guided\cite{ding2020learning} & 1 + Depth & 1 & \fourD & Integer \cite{worrall2019deep}\\
                    Kinematic3D \cite{brazil2020kinematic} & $>1$ & 1 & \fiveD & \xmark \\
                    \myTopRule
                \end{tabular}
            }
        \end{table}

    %============================================================================
    \subsection{Why is Depth the hardest among all parameters?}
        Images are the \twoD{} projections of the \threeD{} scene, and therefore, the depth is lost during projection.
        Recovering this depth is the most difficult to estimate, as shown in Tab.~1 of \cite{ma2021delving}.
        Monocular detection task involves estimating \threeD{} center, \threeD{} dimensions and the yaw angle. 
        The right half of Tab.~1 in \cite{ma2021delving} shows that if the ground truth \threeD{} center is replaced with the predicted center, the detection reaches a minimum. 
        Hence, \threeD{} center is the most difficult to estimate among center, dimensions and pose.
        Most monocular \threeD{} detectors further decompose the \threeD{} center into projected (\twoD{}) center and depth. 
        Out of projected center and depth, Tab.~1 of \cite{ma2021delving} shows that replacing ground truth depth with the predicted depth leads to inferior detection compared to replacing ground truth projected center with the predicted projected center.
        Hence, we conclude that depth is the hardest parameter to estimate.

\clearpage
%============================================================================
%============================================================================
%============================================================================
\section{Implementation Details}\label{sec:implement_details}
    We now provide some additional implementation details for facilitating reproduction of this work.

    \begin{figure}[!t]
        \centering
        \input{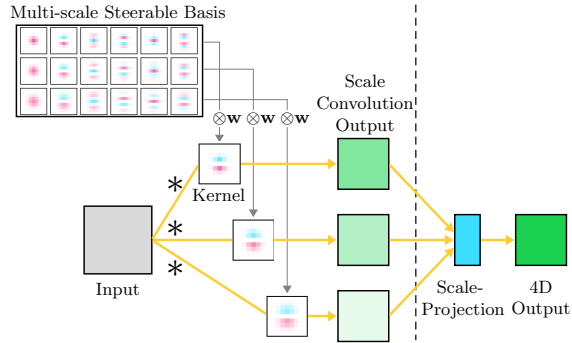}
        \caption{\textbf{(a) \ses{} convolution} \cite{ghosh2019scale,sosnovik2020sesn} The non-trainable basis functions multiply with learnable weights $\weight$ to get kernels. The input then convolves with these kernels to get multi-scale \fiveD{} output. \textbf{(b) \MaxScale} \cite{sosnovik2020sesn} takes $\max$ over the scale dimension of the \fiveD{} output and converts it to \fourD. [Key: $*$ = Vanilla convolution.]}
        \label{fig:steerable_idea}
    \end{figure}

    \begin{figure}[!t]
        \centering
        \includegraphics[width=0.7\linewidth]{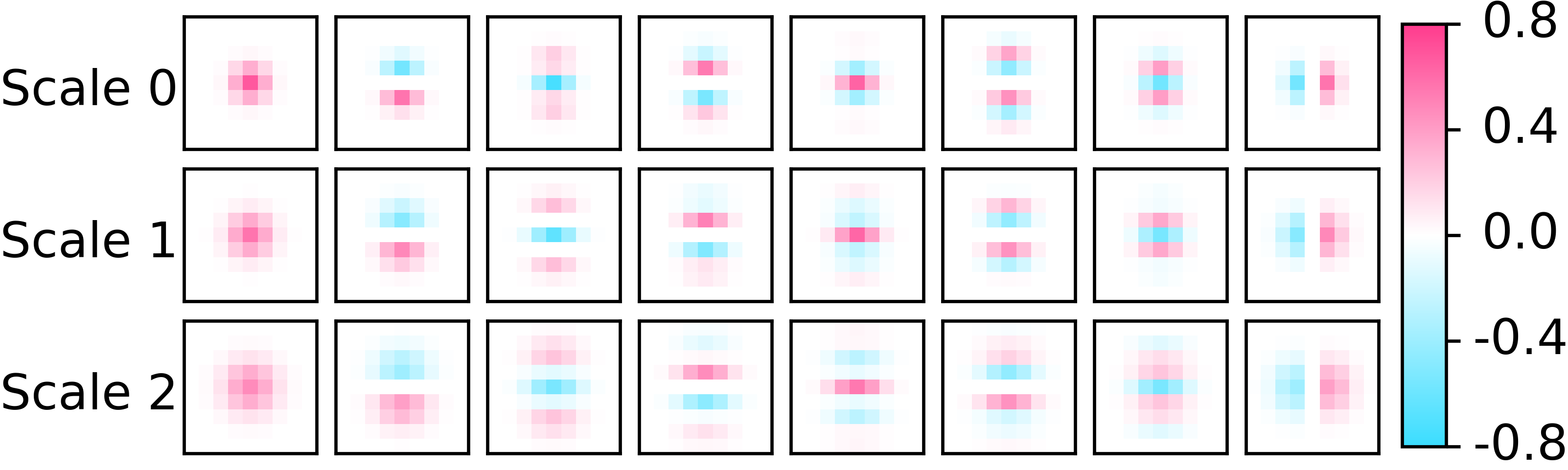}
        \caption{\textbf{Steerable Basis} \cite{sosnovik2020sesn} for $7\!\times\!7$ \ses ~convolution filters. (Showing only $8$ of the $49$ members for each scale).}
        \label{fig:steerable_basis}
    \end{figure}

    %============================================================================
    %============================================================================
    \subsection{Steerable Filters of \ses{} Convolution}\label{sec:steerable_additional}

        We use the \scaleEquivariant{} steerable blocks proposed by \cite{sosnovik2021siamese} for our \methodName{} backbone. 
        We now share the implementation details of these steerable filters.
        
        %============================================================================
        \noIndentHeading{Basis.}
        Although steerable filters can use any linearly independent functions as their basis, we stick with the Hermite polynomials as the basis \cite{sosnovik2021siamese}.
        Let $(0,0)$ denote the center of the function and $(\pixU, \pixV)$ denote the pixel coordinates. 
        Then, the filter coefficients~$\psi_{\sigma n m}$ \cite{sosnovik2021siamese} are         
        \begin{align}
            \psi_{\sigma n m} &= \frac{A}{\sigma^2} H_n \left(\frac{u}{\sigma}\right) H_m \left(\frac{v}{\sigma}\right) e^{-\frac{u^2 + v^2}{\sigma^2}} 
        \end{align}
        $H_n$ denotes the Probabilist's Hermite polynomial of the $n$th order, and $A$ is the normalization constant. 
        The first six Probabilist's Hermite polynomials are
        \begin{align}
            H_0(x) &= 1\\
            H_1(x) &= x\\
            H_2(x) &= x^2 - 1\\
            H_3(x) &= x^3 - 3x\\
            H_4(x) &= x^4 - 6x^2+3
        \end{align}
        \cref{fig:steerable_basis} visualizes some of the \ses{} filters and shows that the basis is indeed at different scales.

    %============================================================================
    %============================================================================
    \subsection{Monocular 3D Detection}\label{sec:detection_training_additional}

        %============================================================================
        \noIndentHeading{Architecture.}
            We use the \dla{} \cite{yu2018deep} configuration, with the standard Feature Pyramid Network (FPN) \cite{lin2017feature}, binning and ensemble of uncertainties.
            FPN is a bottom-up feed-forward CNN that computes feature maps with a downscaling factor of $2$, and a top-down network that brings them back to the high-resolution ones.
            There are total six feature maps levels in this FPN.
            
            We use \dla{} as the backbone for our baseline \gupNet{} \cite{lu2021geometry}, while we use \ses-\dla{} as the backbone for \methodName{}.
            We also replace the \twoD{} pools by \threeD{} pools with pool along the scale dimensions as $1$ for \methodName{}.
            
            We initialize the vanilla CNN from \imageNet{} weights. 
            For \methodName, we use the regularized least squares \cite{sosnovik2021siamese} to initialize the trainable weights in all the Hermite scales from the \imageNet{} \cite{deng2009imagenet} weights.
            Compared to initializing one of the scales as proposed in \cite{sosnovik2021siamese}, we observed more stable convergence in initializing all the Hermite scales.
            
            We output three foreground classes for \kitti{} dataset. 
            We also output three foreground classes for \waymo{} dataset ignoring the Sign class \cite{reading2021categorical}.

        %============================================================================
        \noIndentHeading{Datasets.}
            We use the publicly available \kitti{},\waymo{} and \nuscenes{} datasets for our experiments.
            \kitti{} is available at \url{http://www.cvlibs.net/datasets/kitti/eval_object.php?obj_benchmark=3d} under %CCA-NonCommercial-ShareAlike 
            CC BY-NC-SA 3.0 License.
            \waymo{} is available at \url{https://waymo.com/intl/en_us/dataset-download-terms/} under the Apache License, Version 2.0.
            \nuscenes{} is available at \url{https://www.nuscenes.org/nuscenes} under 
            %Creative Commons Attribution-NonCommercial-ShareAlike 
            CC BY-NC-SA 
            4.0 International Public License.

        %============================================================================
        \noIndentHeading{Augmentation.}
            Unless otherwise stated, we horizontal flip the training images with probability $0.5$, and use scale augmentation as $0.4$ as well for all the models \cite{lu2021geometry} in training.

        %============================================================================
        \noIndentHeading{Pre-processing.}
            The only pre-processing step we use is image resizing.
            \begin{itemize}
                \item \textit{\kitti{}.}
            We resize the $[370, 1242]$ sized~\kitti{} images, and bring them to the $[384, 1280]$ resolution \cite{lu2021geometry}.
            
                \item \textit{\waymo{}.}
            We resize the $[1280, 1920]$ sized~\waymo{} images, and bring them to the $[512, 768]$ resolution. This resolution preserves their aspect ratio. 
            \end{itemize}
        
        %============================================================================
        \noIndentHeading{Box Filtering.}
            We apply simple hand-crafted rules for filtering out the boxes. 
            We ignore the box if it belongs to a class different from the detection class.
            \begin{itemize}
                \item \textit{\kitti{}.} 
                We train with boxes which are atleast $2m$ distant from the ego camera, and with visibility $> 0.5$ \cite{lu2021geometry}. 
            
                \item \textit{\waymo{}.} 
                We train with boxes which are atleast $2m$ distant from the ego camera.
                The \waymo{} dataset does not have any occlusion based labels. 
                However, \waymo{} provides the number of \lidar{}  points inside each \threeD{} box which serves as a proxy for the occlusion.
                We train the boxes which have more than $100$ \lidar{} points for the vehicle class and have more than $50$ \lidar{} points for the cyclist and pedestrian class. 
            %     \item \textit{\nuscenes .} We train with boxes which are atleast $2m$ distant from the ego camera.
            %     In addition, we also drop the boxes if their height in the original resolution is less than $60$px or if the $< 20\%$ of the bounding box of the object lies inside the image.
            %     The \nuscenes{} dataset does not have any occlusion based labels. 
            %     However, \nuscenes{} provides the number of \lidar{} and radar points associated with each box. 
            %     Therefore, we use the sum of the \lidar{} and radar points as a proxy for the occlusion.
            %     If this sum is less than $3$, we drop this box for training. 
            \end{itemize}

        %============================================================================
        \noIndentHeading{Training.}
            We use the training protocol of \gupNet{} \cite{lu2021geometry} for all our experiments.
            Training uses the Adam optimizer \cite{kingma2014adam} and weight-decay $1\times10^{-5}$ .
            Training dynamically weighs the losses using Hierarchical Task Learning (HTL)  \cite{lu2021geometry} strategy keeping $K$ as $5$ \cite{lu2021geometry}. 
            Training also uses a linear warmup strategy in the first $5$ epochs to stabilize the training.
            We choose the model saved in the last epoch as our final model for all our experiments.
            \begin{itemize}
                \item \textit{\kitti .} 
                We train with a batch size of $12$ on single Nvidia A100 (40GB) GPU for $140$ epochs. 
                Training starts with a learning rate $1.25 \times 10^{-3}$ with a step decay of $0.1$ at the $90$th and the $120$th epoch.

                \item \textit{\waymo .}
                We train with a batch size of $40$ on single Nvidia A100 (40GB) GPU for $30$ epochs because of the large size of the \waymo{} dataset.
                Training starts with a learning rate $1.25 \times 10^{-3}$ with a step decay of $0.1$ at the $18$th and the $26$th epoch.
            \end{itemize}

        %============================================================================
        \noIndentHeading{Losses.}
            We use the \gupNet{} \cite{lu2021geometry} multi-task losses before the NMS for training. The total loss $\loss$ is given by
            \begin{align}
                \loss &= \loss_\heatmap + \loss_{\twoDMath ,\offset} + \loss_{\twoDMath,\size} + \loss_{\threeDMath\twoDMath,\offset}+ \loss_{\threeDMath,angle}\nonumber\\
                &\quad\quad + \loss_{\threeDMath,l} + \loss_{\threeDMath,w} + \loss_{\threeDMath,h} + \loss_{\threeDMath,depth}.
            \end{align}
            The individual terms are given by
            \begin{align}
                \loss_\heatmap &= \text{Focal}(\class^b, \class^g), \\
                \loss_{\twoDMath ,\offset} &= \lOne (\delta_{\twoDMath}^b, \delta_{\twoDMath}^g), \\   
                \loss_{\twoDMath,\size} &= \lOne (w_{\twoDMath}^b, w_{\twoDMath}^g) + \lOne (h_{\twoDMath}^b, h_{\twoDMath}^g), \\
                \loss_{\threeDMath\twoDMath,\offset} &= \lOne(\delta_{\threeDMath\twoDMath}^b, \delta_{\threeDMath\twoDMath}^g)\\
                \loss_{\threeDMath,angle} &= \text{CE}(\alpha^b, \alpha^g) \\
                \loss_{\threeDMath,l} &= \lOne(\mu_{l{\threeDMath}}^b, \delta_{l\threeDMath}^g) \\
                \loss_{\threeDMath,w} &= \lOne(\mu_{w{\threeDMath}}^b, \delta_{w{\threeDMath}}^g) \\
                \loss_{\threeDMath,h} &= \frac{\sqrt{2}}{\sigma_{h\threeDMath}}\lOne(\mu^b_{h\threeDMath}, \delta_{h\threeDMath}^g) + \ln(\sigma_{h\threeDMath}) \\
                \loss_{\threeDMath,depth} &= \frac{\sqrt{2}}{\sigma_d}\lOne(\mu^b_{d}, \mu_{d}^g) + \ln(\sigma_d),
            \end{align}
            where, \begin{align}
                \mu^b_d &= \focal \frac{\mu_{h\threeDMath}^b}{h_{\twoDMath}^b} + \mu_{d,pred} \\
                \sigma_d &= \sqrt{ \left( \focal \frac{\sigma_{h\threeDMath}}{h_{\twoDMath}^b} \right)^2 + \sigma_{d,pred}^2 }.
            \end{align}
            
            The superscripts $b$ and $g$ denote the predicted box and ground truth box respectively. 
            CE and Focal denote the Cross Entropy and Focal loss respectively.
            
            The number of heatmaps depends on the number of output classes.
            $\delta_{\twoDMath}$ denotes the deviation of the \twoD{} center from the center of the heatmap.
            $\delta_{\threeDMath\twoDMath, \offset}$ denotes the deviation of the projected \threeD{} center from the center of the heatmap.
            The orientation loss is the cross entropy loss between the binned observation angle of the prediction and the ground truth.
            The observation angle $\alpha$ is split into $12$ bins covering $30\degree$ range.
            $\delta_{l\threeDMath}, \delta_{w\threeDMath}$ and $\delta_{h\threeDMath}$ denote the deviation of the \threeD{} length, width and height of the box from the class dependent mean size respectively.
            
            The depth is the hardest parameter to estimate\cite{ma2021delving}. 
            So, \gupNet{} uses in-network ensembles to predict the depth. 
            It obtains a Laplacian estimate of depth from the \twoD{} height, while it obtains another estimate of depth from the prediction of depth. 
            It then adds these two depth estimates.
    
        %============================================================================
        \noIndentHeading{Inference.}\label{sec:det_testing_additional}
            Our testing resolution is same as the training resolution. 
            We do not use any augmentation for test/validation.
            We keep the maximum number of objects to $50$ in an image, and we multiply the class and predicted confidence to get the box's overall score in inference as in \cite{kumar2021groomed}.
            We consider output boxes~with scores greater than a threshold of $0.2$ for \kitti{} \cite{lu2021geometry} and $0.1$ for \waymo{} \cite{reading2021categorical}.

\clearpage
%============================================================================
%============================================================================
%============================================================================
\section{Additional Experiments and Results}\label{sec:additional_exp}
    We now provide additional details and results of the experiments evaluating our system's performance.

    %============================================================================
    %============================================================================
    \subsection{\kitti{} \valOne{} Split}

        \begin{table*}[!tb]
            \caption{\textbf{Generalization gap} (\downarrowRHDSmall) on~\kitti{} \valOne{} cars. Monocular detection has huge generalization gap between training and inference sets. [Key: \bestKey{Best}]}
            \label{tab:kitti_compare_generalization_big}
            \centering
            \scalebox{\scaleFraction}{
                \footnotesize
                \setlength{\tabcolsep}{0.1cm}
                    % \multicolumn{3}{tcm}{} & \multicolumn{6}{cm}{\iouThreeD{} $\geq 0.7$} & \multicolumn{6}{ct}{\iouThreeD{} $\geq 0.5$}\\\cline{1-15}
                    % \multirow{2}{*}{Method} & Scale & \multirow{2}{*}{Split} & \multicolumn{3}{ct}{\apThreeDForty \bracketPercentage(\uparrowRHDSmall)} & \multicolumn{3}{cm}{\apBevForty \bracketPercentage(\uparrowRHDSmall)} & \multicolumn{3}{ct}{\apThreeDForty \bracketPercentage(\uparrowRHDSmall)} & \multicolumn{3}{ct}{\apBevForty \bracketPercentage(\uparrowRHDSmall)}\\
                \begin{tabular}{tl|c m c m ccc t ccc m ccc t ccct}
                    \myTopRule
                    \addlinespace[0.01cm]
                    \multirow{3}{*}{Method} & Scale  & \multirow{3}{*}{Set} & \multicolumn{6}{cm}{\iouThreeD{} $\geq 0.7$} & \multicolumn{6}{ct}{\iouThreeD{} $\geq 0.5$}\\
                    \cline{4-15}
                    & Eqv & & \multicolumn{3}{ct}{\apThreeDForty \bracketPercentage(\uparrowRHDSmall)} & \multicolumn{3}{cm}{\apBevForty \bracketPercentage(\uparrowRHDSmall)} & \multicolumn{3}{ct}{\apThreeDForty \bracketPercentage(\uparrowRHDSmall)} & \multicolumn{3}{ct}{\apBevForty \bracketPercentage(\uparrowRHDSmall)}\\%\cline{2-13}\\[0.05cm]
                    & & & Easy & Mod & Hard & Easy & Mod & Hard & Easy & Mod & Hard & Easy & Mod & Hard\\
                    % \myTopRule
                    % \multirow{3}{*}{GrooMeD\cite{kumar2021groomed}}  &  & Train & $86.08$ & $69.98$ & $52.18$ & $88.44$ & $71.68$ & $55.32$ & $97.07$ & $84.56$ & $65.21$ & $97.22$ & $86.85$ & $67.47$\\
                    % & & Val &$19.67$ & $14.32$ & $11.27$ & $27.38$ & $19.75$ & $15.92$ & $55.62$ & $41.07$ & $32.89$ & $61.83$ & $44.98$ & $36.29$\\
                    % \cline{4-15}
                    % & & \COFix Gap & \COFix$66.41$ & \COFix$55.66$ & \COFix$40.91$ & \COFix$61.06$ & \COFix$51.93$ & \COFix$39.40$ & \COFix$41.45$ & \COFix$43.49$ & \COFix$32.32$ & \COFix$35.39$ & \COFix$41.87$ & \COFix$31.27$\\
                    \myTopRule
                    \multirow{3}{*}{\gupNet{} \cite{lu2021geometry}} & & Train & $91.83$ & $74.87$ & $67.43$ & $95.19$ & $80.95$ & $73.55$ & $99.50$ & $93.62$ & $86.22$ & $99.56$ & $93.88$ & $86.46$\\
                    & & Val & $21.10$ & $15.48$ & $12.88$ & $28.58$ & $20.92$ & $17.83$ & $58.95$ & $43.99$ & $38.07$ & $64.60$ & $47.76$ &	$42.97$\\
                    \cline{4-15}
                    & & \COFix Gap & \COFix$70.73$ & \COFix\best{59.39} & \COFix$54.55$ & \COFix$66.61$ & \COFix$60.03$ & \COFix$55.72$ & \COFix$40.55$ & \COFix$49.63$ & \COFix\best{48.15} & \COFix$34.96$ & \COFix$46.12$ & \COFix\best{43.49}\\
                    \myTopRule
                    \multirow{3}{*}{\textbf{\methodNameShort}}  & \multirow{3}{*}{\checkmark} & Train & $91.09$ & $76.19$ & $67.16$ & $94.76$ & $82.61$ & $75.51$ & $99.37$ & $93.56$ & $88.57$ & $99.50$ & $93.87$ & $88.90$\\
                    & & Val & $24.63$ & $16.54$ & $14.52$ & $32.60$ & $23.04$ & $19.99$ & $61.00$ & $46.00$ & $40.18$ & $65.28$ & $49.63$ & $43.50$\\
                    \cline{4-15}
                    & & \COFix Gap & \COFix\best{66.46} & \COFix$59.65$ & \COFix\best{52.64} & \COFix\best{62.16} & \COFix\best{59.57} & \COFix\best{55.52} & \COFix\best{38.37} & \COFix\best{47.56} & \COFix$48.39$ & \COFix\best{34.22} & \COFix\best{44.24} & \COFix$45.40$\\
                    \myTopRule
                \end{tabular}
            }
        \end{table*}
    
        %============================================================================
        \noIndentHeading{Monocular Detection has Huge Generalization Gap.}
            As mentioned in \cref{sec:intro}, we now show that the monocular detection has huge generalization gap between training and inference.
            We report the object detection performance on the train and validation (val) set for the two models on \kitti{} \valOne{} split in \cref{tab:kitti_compare_generalization_big}. 
            \cref{tab:kitti_compare_generalization_big} shows that the performance of our baseline \gupNet{} \cite{lu2021geometry} and our \methodName{} is huge on the training set, while it is less than one-fourth of the train performance on the val set.
            
            We also report the \ccolorbox{my_magenta_highlight}{generalization~gap} metric \cite{wu2021rethinking} in \cref{tab:kitti_compare_generalization_big}, which is the difference between training and validation performance.
            The generalization gap at both the thresholds of $0.7$ and $0.5$ is huge.
    
        \begin{table}[!tb]
            \caption{\textbf{Comparison on multiple backbones} on \kitti{} \valOne{} cars. [Key: \bestKey{Best}]}
            \label{tab:kitti_compare_backbones}
            \centering
            \scalebox{\scaleFraction}{
                \footnotesize
                \rowcolors{4}{lightgray}{white}
                \setlength\tabcolsep{0.1cm}
                \begin{tabular}{ml m c m ccc t ccc m ccc t cccm}
                    \myTopRule
                    \addlinespace[0.01cm]
                    \multirow{3}{*}{BackBone} & \multirow{3}{*}{Method} & \multicolumn{6}{cm}{\iouThreeD{} $\geq 0.7$} & \multicolumn{6}{cm}{\iouThreeD{} $\geq 0.5$}\\
                    \cline{3-14}
                    & & \multicolumn{3}{ct}{\apThreeDForty \bracketPercentage(\uparrowRHDSmall)} & \multicolumn{3}{cm}{\apBevForty \bracketPercentage(\uparrowRHDSmall)} & \multicolumn{3}{ct}{\apThreeDForty \bracketPercentage(\uparrowRHDSmall)} & \multicolumn{3}{cm}{\apBevForty \bracketPercentage(\uparrowRHDSmall)}\\
                    & & Easy & Mod & Hard & Easy & Mod & Hard & Easy & Mod & Hard & Easy & Mod & Hard\\ 
                    \myTopRule
                    \resNetEighteen & \gupNet\!\cite{lu2021geometry}& $18.86$ & $13.20$ & $11.01$ & $26.05$ & $19.37$ & $16.57$ & $54.90$ & $40.65$ & $34.98$ & $60.54$ & $46.13$ & $40.12$\\
                    & \bestKey{\methodName} & $20.27$ & $14.21$ & $12.56$ & $28.09$ & $20.32$ & $17.49$ & $55.75$ & $42.41$ & $36.97$ & $60.82$ & $46.43$ & $40.59$\\
                    \myTopRule
                    \dla &\gupNet\!\cite{lu2021geometry} & $21.10$ & $15.48$ & $12.88$ & $28.58$ & $20.92$ & $17.83$ & $58.95$ & $43.99$ & $38.07$ & $64.60$ & $47.76$ &	$42.97$\\
                    & \bestKey{\methodName} & \best{24.63} & \best{16.54} & \best{14.52} & \best{32.60} &	\best{23.04} & \best{19.99} &	\best{61.00} & \best{46.00}	& \best{40.18} & \best{65.28} & \best{49.63} & \best{43.50} \\
                    \myTopRule
                \end{tabular}
            }
        \end{table}

        %============================================================================
        \noIndentHeading{Comparison on Multiple Backbones.}
            A common trend in \twoD{} object detection community is to show improvements on multiple backbones \cite{wang2020scale}. 
            \ddThreeD \cite{park2021pseudo} follows this trend and also reports their numbers on multiple backbones.
            Therefore, we follow the same and compare with our baseline on multiple backbones on~\kitti{} \valOne{} cars in \cref{tab:kitti_compare_backbones}.
            \cref{tab:kitti_compare_backbones} shows that \methodName{} shows consistent improvements over \gupNet{} \cite{lu2021geometry} in \threeD{} object detection on multiple backbones, proving the effectiveness of our proposal.
               
        %===============================================================================
        \noIndentHeading{Comparison with Bigger CNN Backbones.}
            Since the SES blocks increase the Flop counts significantly compared to the vanilla convolution block, we next compare \methodName{} with bigger CNN backbones with comparable GFLOPs and FPS/ wall-clock time (instead of same configuration) in \cref{tab:detection_with_bigger_cnn}.
            We compare \methodName{} with \dlaOneZeroTwo{} and \dlaOneSixNine{} - two biggest DLA networks with \imageNet{} weights\footnote{Available at \url{http://dl.yf.io/dla/models/imagenet/}} on \kitti{} \valOne{} split. 
            We use the fvcore library\footnote{\url{https://github.com/facebookresearch/fvcore}} to get the parameters and flops.
            \cref{tab:detection_with_bigger_cnn} shows that \methodName{} again outperforms the bigger CNN backbones, especially on nearby objects. 
            We believe this happens because the bigger CNN backbones have more trainable parameters than \methodName{}, which leads to overfitting.
            Although \methodName{} takes more time compared to the CNN backbones, \methodName{} still keeps the inference almost real-time.
            
        \begin{table}[!tb]
            \caption{\textbf{Results with bigger CNNs having similar flops} on \kitti{} \valOne{} cars. 
            [Key: \firstkey{Best}]
            }
            \label{tab:detection_with_bigger_cnn}
            \centering
            \scalebox{\scaleFraction}{
                \footnotesize
                \setlength\tabcolsep{1.00pt}
                \rowcolors{3}{lightgray}{white}
                \begin{tabular}{tl m c m c m c m c m c m ccc m ccct}
                    \myTopRule
                    \multirow{2}{*}{Method} & \multirow{2}{*}{BackBone} & Param (\downarrowRHDSmall) & Disk Size (\downarrowRHDSmall) & Flops (\downarrowRHDSmall) & Infer (\downarrowRHDSmall)& \multicolumn{3}{cm}{\apThreeD{} \iouThreeD{}$\ge 0.7$ (\uparrowRHDSmall)} & \multicolumn{3}{ct}{\apThreeD{} \iouThreeD{}$\ge 0.5$ (\uparrowRHDSmall)}\\ 
                    & & (M) & (MB) & (G) & (ms) & Easy & Mod & Hard & Easy & Mod & Hard\\
                    \myTopRule
                    \gupNet{} \cite{lu2021geometry} & \dlaThirtyFour & \first{16} & \first{235} & \first{30} & \first{20} & $21.10$ & $15.48$ & $12.88$ & $58.95$ & $43.99$ & $38.07$\\
                    \gupNet{} \cite{lu2021geometry} & \dlaOneZeroTwo & $34$ & $583$ & $70$ & $25$ & $20.96$ & $14.64$ & $12.80$ & $57.06$	& $41.78$	& $37.26$\\
                    \gupNet{} \cite{lu2021geometry} & \dlaOneSixNine & $54$ & $814$ & $114$ & $30$ & $21.76$ & $15.35$ & $12.72$ & $57.60$	& $43.27$ & $37.32$\\
                    \hline
                    \methodNameShort & SES-\dlaThirtyFour & \first{16} & $236$ & $235$ & $40$ & \first{24.63} & \first{16.54} & \first{14.52} & \first{61.00} & \first{46.00}	& \first{40.18}\\
                    \myTopRule
                \end{tabular}
            }
        \end{table}

        %============================================================================
        \noIndentHeading{Performance on Cyclists and Pedestrians.}
            \cref{tab:detection_results_kitti_valone_ped_cyclist} lists out the results of \threeD{} object detection on \kitti{} \valOne{} Cyclist and Pedestrians. 
            The results show that \methodName{} is competitive on challenging Cyclist and achieves \sota{} results on Pedestrians on the \kitti{} \valOne{} split.

        \begin{table}[!tb]
            \caption{\textbf{Results on \kitti{} \valOne{} cyclists and pedestrians} (Cyc/Ped) (\iouThreeD$\geq\!0.5$). [Key: \firstkey{Best}, \secondkey{Second Best}]
            }
            \label{tab:detection_results_kitti_valone_ped_cyclist}
            \centering
            \scalebox{\scaleFraction}{
                \footnotesize
                \rowcolors{3}{lightgray}{white}
                \setlength\tabcolsep{0.1cm}
                \begin{tabular}{ml m c m ccc  m cccm}
                    \myTopRule
                    \addlinespace[0.01cm]
                    \multirow{2}{*}{Method} & \multirow{2}{*}{Extra} &\multicolumn{3}{cm}{Cyc \apThreeDForty \bracketPercentage(\uparrowRHDSmall)} & \multicolumn{3}{cm}{Ped \apThreeDForty \bracketPercentage(\uparrowRHDSmall)}\\ 
                    & & Easy & Mod & Hard & Easy & Mod & Hard\\ 
                    \myTopRule
                    \groomedNMS \cite{kumar2021groomed} & \mathDash{}& $0.00$ & $0.00$ & $0.00$ & $3.79$ & $2.71$ & $2.61$\\
                    MonoDIS \cite{simonelli2019disentangling}& \mathDash{}& $1.52$ & $0.73$ & $0.71$ & $3.20$ & $2.28$ & $1.71$\\ 
                    \monoDISMulti \cite{simonelli2020disentangling}& \mathDash{}   & $2.70$        & $1.50$        & $1.30$    & $9.50$        & $7.10$        & $5.70$          \\
                    \gupNet{} (Retrained) \cite{lu2021geometry} & \mathDash{}  & \first{4.41} & \second{2.17} & \second{2.03} & \second{9.37} & \second{6.84} & \first{5.73} \\
                    \hline	 
                    \rowcolor{white}
                    \bestKey{\methodName{} (Ours)}                        & \mathDash{}  & \second{4.05} & \first{2.20} & \first{2.14}  & \first{9.85} & \first{7.18} & \second{5.42} \\
                    \myTopRule
                \end{tabular}
            }
        \end{table}

        %============================================================================
        \noIndentHeading{Cross-Dataset Evaluation Details.}\label{sec:results_cross_dataset_additional}
            For cross-dataset evaluation, we test on all $3{,}769$ images of the \kitti{} \valOne{} split, as well as all frontal $6{,}019$ images of the \nuscenes{} \val{} split \cite{caesar2020nuscenes}, as in \cite{shi2021geometry}.
            We first convert the \nuscenes{} \val{} images to the KITTI format using the \texttt{export\_kitti}\footnote{\url{https://github.com/nutonomy/nuscenes-devkit/blob/master/python-sdk/nuscenes/scripts/export\_kitti.py}} function in the nuscenes devkit.
            We keep \kitti{} \valOne{} images in the $[384, 1280]$ resolution, while we keep the \nuscenes{} \val{} images in the $[384, 672]$ resolution to preserve the aspect ratio. 
            For \mthreeDRPN \cite{brazil2019m3d}, we bring the \nuscenes{} \val{} images in the $[512, 910]$ resolution.
            
            Monocular \threeD{} object detection relies on the camera focal length to back-project the projected centers into the \threeD{} space. 
            Therefore, the \threeD{} centers depends on the focal length of the camera used in the dataset. 
            Hence, one should take the camera focal length into account while doing cross-dataset evaluation. 
            We now calculate the camera focal length of a dataset as follows. 
            We take the camera matrix $\projectionOperator$ and calculate the normalized focal length $\focalNormalized = \frac{2 f_y}{H}$, where $H$ denotes the height of the image. 
            The normalized focal length $\focalNormalized$ for the \kitti{} dataset is $3.82$, while the normalized focal length $\focalNormalized$ for the \nuscenes{} dataset is $2.82$.
            Thus, the \kitti{} and the \nuscenes{} images have a different focal length \cite{wang2020train}.
            
            \mthreeDRPN \cite{brazil2019m3d} does not normalize w.r.t. the focal length. 
            So, we explicitly correct and divide the depth predictions of \nuscenes{} images from the \kitti{} model by $3.82/2.82= 1.361$ in the  \mthreeDRPN \cite{brazil2019m3d} codebase.
            The \gupNet{} \cite{lu2021geometry} and \methodName{} codebases use normalized coordinates \thatIs{} they normalize w.r.t. the focal length. 
            So, we do not explicitly correct the focal length for \gupNet{} and \methodName{} predictions.
            
            We match predictions to the ground truths using the \iouTwoD{} overlap threshold of $0.7$ \cite{shi2021geometry}. 
            After this matching, we calculate the Mean Average Error (MAE) of the depths of the predicted and the ground truth boxes \cite{shi2021geometry}.

        %===============================================================================
        \noIndentHeading{Stress Test with Rotational and/or xy-translation Ego Movement.} 
            \cref{th:projective_scaled} uses translation along the depth as the sole ego movement. 
            This assumption might be valid for the current outdoor datasets and benchmarks, but is not the case in the real world. 
            Therefore, we conduct stress tests on how tolerable \methodName{} and \gupNet{} \cite{lu2021geometry} are when there is rotational and/or $xy$-translation movement on the vehicle.
    
            % We evaluate the robustness as follows. 
            First, note that \kitti{} and \waymo{} are already large-scale real-world datasets, and our own dataset might not be a good choice.
            So, we stick with \kitti{} and \waymo{} datasets.
            We manually choose $306$ \kitti{} \valOne{} images with such ego movements and again compare performance of \methodName{} and \gupNet{} on this subset in \cref{tab:detection_with_rotational_movement}.
            The average distance of the car in this subset is $27.69$ m $(\pm 16.59$ m$)$, which suggests a good variance and unbiasedness in the subset.
            \cref{tab:detection_with_rotational_movement} shows that both the \methodName{} backbone and the CNN backbone show a drop in the detection performance by about $4$ AP points on the Mod cars of ego-rotated subset compared to the all set. 
            This drop experimentally confirms the theory that both the \methodName{} backbone and the CNN backbone do not handle arbitrary \threeD{} rotations. 
            More importantly, the table shows that \methodName{} maintains the performance improvement over \gupNet{} \cite{lu2021geometry} under such movements.

            Also, \waymo{} has many images in which the ego camera shakes.
            Improvements on \waymo{} (\cref{tab:waymo_val}) also confirms that \methodNameShort{} outperforms \gupNet{} \cite{lu2021geometry} even when there is rotational or $xy$-translation ego movement.

        \begin{table}[!tb]
            \caption{\textbf{Stress Test} with rotational and $xy$-translation ego movement on \kitti{} \valOne{} cars.
            [Key: \firstkey{Best}]
            }
            \label{tab:detection_with_rotational_movement}
            \centering
            \centering
            \scalebox{\scaleFraction}{
                \footnotesize
                \setlength\tabcolsep{0.1cm}
                \rowcolors{3}{lightgray}{white}
                \begin{tabular}{tl m l m ccc  m ccct}
                    \myTopRule
                    \multirow{2}{*}{Set }  & \multirow{2}{*}{Method }  & \multicolumn{3}{cm}{\apThreeD{} \iouThreeD{}$\ge 0.7$ (\uparrowRHDSmall)} & \multicolumn{3}{ct}{\apThreeD{} \iouThreeD{}$\ge 0.5$ (\uparrowRHDSmall)}\\ 
                    & & Easy & Mod & Hard & Easy & Mod & Hard\\ 
                    \myTopRule
                    Subset & \gupNet{} \cite{lu2021geometry} & $17.22$ & $11.43$ & $9.91$ & $47.47$ & $35.02$ & $32.63$\\
                    $(306)$ & \methodNameShort{} & \first{20.17} & \first{12.49} & \first{10.93} & \first{49.81} & \first{36.93} & \first{34.32}\\
                    \myTopRule
                    \kitti{} \valOne{} & \gupNet{} \cite{lu2021geometry} & $21.10$ & $15.48$ & $12.88$ & $58.95$ & $43.99$ & $38.07$\\
                    $(3769)$ & \methodNameShort{} & \first{24.63} & \first{16.54} & \first{14.52} & \first{61.00} & \first{46.00}	& \first{40.18}\\
                    \myTopRule
                \end{tabular}
            }
        \end{table}

        %===============================================================================
        \noIndentHeading{Comparison of Depth Estimates from Monocular Depth Estimators and 3D Object Detectors.}
            We next compare the depth estimates from monocular depth estimators and depth estimates from monocular \threeD{} object detectors on the foreground objects. 
            We take a monocular depth estimator BTS \cite{lee2019big} model trained on \kitti{} Eigen split. 
            We next compare the depth error for all and foreground objects (cars) on \kitti{} \valOne{} split using MAE (\downarrowRHDSmall) metric in \cref{tab:detection_depth_estimator_depth} as in \cref{tab:detection_cross_dataset}.
            We use the MSeg \cite{lambert2020mseg} to segment out cars in the driving scenes for BTS.
            \cref{tab:detection_depth_estimator_depth} shows that the depth from BTS is not good for foreground objects (cars) beyond $20+$ m range.
            Note that there is a data leakage issue between the \kitti{} Eigen train split and the \kitti{} \valOne{} split \cite{simonelli2021we} and therefore, we expect more degradation in performance of monocular depth estimators after fixing the data leakage issue.

        \begin{table}[!tb]
            \caption{\textbf{Comparison of Depth Estimates} of monocular depth estimators and 3D object detectors on \kitti{} \valOne{} cars. Depth from a depth estimator BTS is not good for foreground objects (cars) beyond $20+$ m range.
            [Key: \firstkey{Best}, \secondkey{Second Best}]
            }
            \label{tab:detection_depth_estimator_depth}
            \centering
            \scalebox{\scaleFraction}{
                \footnotesize
                \setlength\tabcolsep{0.1cm}
                \rowcolors{3}{lightgray}{white}
                \begin{tabular}{tl m cc m ccc  m ccct}
                    \myTopRule
                    \multirow{2}{*}{Method} & Depth & Ground & \multicolumn{3}{cm}{Back+ Foreground} & \multicolumn{3}{ct}{Foreground (Cars)}\\
                    & at & Truth & $0\!-\!20$ & $20\!-\!40$ & $40\!-\!\infty$ &$0\!-\!20$ & $20\!-\!40$ & $40\!-\!\infty$\\ 
                    \myTopRule
                    \gupNet{} \cite{lu2021geometry} & \threeD{} Center & \threeD{} Box & \mathDash & \mathDash & \mathDash & $0.45$ & \second{1.10} & \second{1.85}\\
                    \methodNameShort & \threeD{} Center & \threeD{} Box & \mathDash & \mathDash & \mathDash & \second{0.40} & \first{1.09} & \first{1.80}\\
                    \hline
                    BTS \cite{lee2019big} & Pixel & \lidar{} &  $0.48$ & $1.30$ & $1.83$ & \first{0.30} & $1.22$ & $2.16$\\
                    \myTopRule
                \end{tabular}
            }
        \end{table}

        \begin{figure}[!tb]
            \centering
            \includegraphics[width=0.45\linewidth]{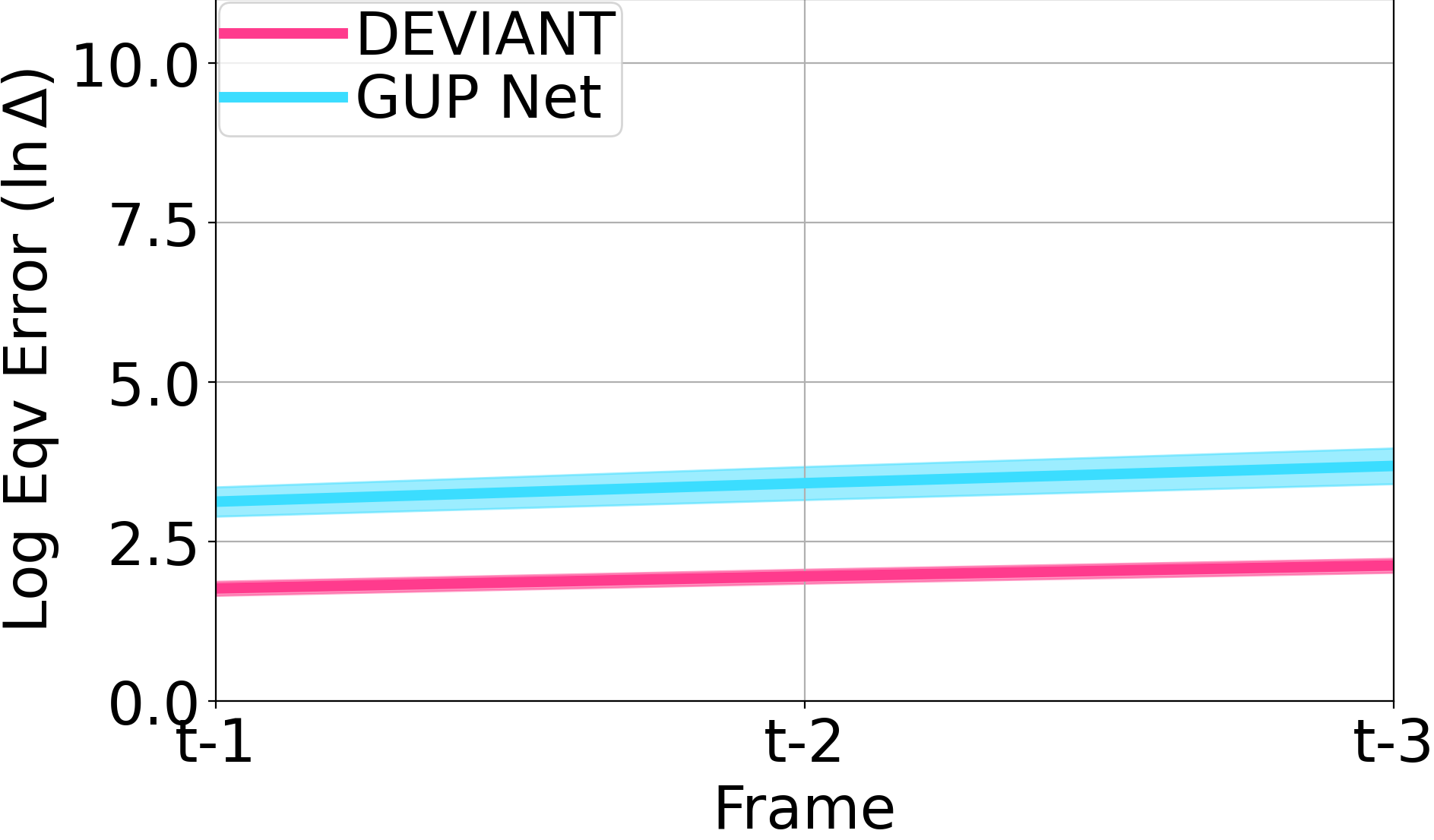}
            \caption{\textbf{\Equivariance{} error $(\Delta)$} comparison for \methodName{} and \gupNet{} on previous three frames of the \kitti{} monocular videos at block $3$ in the backbone.}
            \label{fig:equiv_error_mono_videos}
        \end{figure}

        %============================================================================
        \noIndentHeading{\Equivariance{} Error for \kitti{} Monocular Videos.}
            A better way to compare the scale \equivariance{} of the \methodName{} and \gupNet{} \cite{lu2021geometry} compared to \cref{fig:equiv_error}, is to compare \equivariance{} error on real images with depth translations of the ego camera.
            The \equivariance{} error $\Delta$ is the normalized difference between the scaled feature map and the feature map of the scaled image, and is given by
            \begin{align}
                \Delta &= \frac{1}{N} \sum_{i=1}^N \frac{||\transformationMath_{s_i} \mapping(\projectionOneIndexed) - \mapping(\transformationMath_{s_i} \projectionOneIndexed)||_2^2}{||\transformationMath_{s_i} \mapping(\projectionOneIndexed)||_2^2},
                \label{eq:equiv_error}
            \end{align}
            where $\mapping$ denotes the neural network, $\transformationMath_{s_i}$ is the scaling transformation for the image $i$, and $N$ is the total number of images.
            Although we do evaluate this error in \cref{fig:equiv_error}, the image scaling in \cref{fig:equiv_error} does not involve scene change because of the absence of the moving objects.
            Therefore, evaluating on actual depth translations of the ego camera makes the \equivariance{} error evaluation more realistic.
            We next carry out this experiment and report the \equivariance{} error on three previous frames of the val images of the \kitti{} \valOne{} split as in \cite{brazil2020kinematic}.
            We plot this \equivariance{} error in \cref{fig:equiv_error_mono_videos} at block $3$ of the backbones because the resolution at this block corresponds to the output feature map of size $[96, 320]$.
            \cref{fig:equiv_error_mono_videos} is similar to \cref{fig:equiv_error_scaling}, and shows that \methodName{} achieves lower \equivariance{} error.
            Therefore, \methodName{} has better \equivariance{} to depth translations (scale \transformation{} s) than \gupNet{}\cite{lu2021geometry} in real scenarios.

        %============================================================================
        \noIndentHeading{Model Size, Training, and Inference Times.}
            Both \methodName{} and the baseline \gupNet{} have the same number of trainable parameters, and therefore, the same model size. 
            \gupNet{} takes $4$ hours to train on \kitti{} \valOne{} and $0.02$ ms per image for inference on a single Ampere A100 (40 GB) GPU.
            \methodName{} takes $8.5$ hours for training and $0.04$ ms per image for inference on the same GPU.
            This is expected because \se{} models use more flops \cite{zhu2019scale, sosnovik2020sesn} and, therefore, \methodName{} takes roughly twice the training and inference time as \gupNet.

        %============================================================================
        \noIndentHeading{Reproducibility.}
            As described in \cref{sec:detection_results_kitti_val1}, we now list out the five runs of our baseline \gupNet{} \cite{lu2021geometry} and \methodName{} in \cref{tab:runs_results_kitti_val1}.
            \cref{tab:runs_results_kitti_val1} shows that \methodName{} outperforms \gupNet{} in all runs and in the average run.
            
        \begin{table}[!tb]
            \caption{\textbf{Five Different Runs} on \kitti{} \valOne{} cars. 
            [Key: \bestKey{Average}]
            }
            \label{tab:runs_results_kitti_val1}
            \centering
            \scalebox{\scaleFraction}{
            \footnotesize
            \rowcolors{4}{lightgray}{white}
            \setlength\tabcolsep{0.1cm}
            \begin{tabular}{ml m c m ccc t ccc m ccc t cccm}
                \myTopRule
                \addlinespace[0.01cm]
                \multirow{3}{*}{Method} & \multirow{3}{*}{Run} & \multicolumn{6}{cm}{\iouThreeD{} $\geq 0.7$} & \multicolumn{6}{cm}{\iouThreeD{} $\geq 0.5$}\\\cline{3-14}
                & & \multicolumn{3}{ct}{\apThreeDForty \bracketPercentage(\uparrowRHDSmall)} & \multicolumn{3}{cm}{\apBevForty \bracketPercentage(\uparrowRHDSmall)} & \multicolumn{3}{ct}{\apThreeDForty \bracketPercentage(\uparrowRHDSmall)} & \multicolumn{3}{cm}{\apBevForty \bracketPercentage(\uparrowRHDSmall)}\\%\cline{2-13}
                & & Easy & Mod & Hard & Easy & Mod & Hard & Easy & Mod & Hard & Easy & Mod & Hard\\
                \myTopRule
                & 1 & $21.67$ & $14.75$ & $12.68$ & $28.72$ & $20.88$ & $17.79$ & $58.27$ & $43.53$ & $37.62$ & $63.67$ & $47.37$ & $42.55$\\
                & 2& $21.26$ & $14.94$ & $12.49$ & $28.39$ & $20.40$ & $17.43$ & $59.20$ & $43.55$ & $37.63$ & $64.06$ & $47.46$ & $42.67$\\
                \gupNet{} \cite{lu2021geometry} 
                & 3 & $20.87$ & $15.03$ & $12.61$ & $28.66$ & $20.56$ & $17.48$ & $60.19$ & $44.08$ & $39.36$ & $65.26$ & $49.44$ & $43.17$\\
                % & 4 & \best{21.10} & \best{15.48} & \best{12.88} & \best{28.58} & \best{20.92} & \best{17.83} & \best{58.95} & \best{43.99} & \best{38.07} & \best{64.60} & \best{47.76} &	\best{42.97} \\
                & 4 & $21.10$ & $15.48$ & $12.88$ & $28.58$ & $20.92$ & $17.83$ & $58.95$ & $43.99$ & $38.07$ & $64.60$ & $47.76$ &	$42.97$ \\
                & 5 & $22.52$ & $15.92$ & $13.31$ & $30.77$ & $22.40$ & $19.36$ & $59.91$ & $44.00$ & $39.30$ & $64.94$ & $48.01$ & $43.08$\\
                \cline{2-14}
                & Avg & \best{21.48} & \best{15.22} & \best{12.79} & \best{29.02} & \best{21.03} & \best{17.98} & \best{59.30} & \best{43.83} & \best{38.40} & \best{64.51} & \best{48.01} & \best{42.89}\\
                \myTopRule
                & 1 & $23.19$ & $15.84$ & $14.11$ & $29.82$ & $21.93$ & $19.16$ & $60.19$ & $45.52$ & $39.86$ & $66.32$ & $49.39$ & $43.38$\\
                & 2 & $23.33$ & $16.12$ & $13.54$ & $31.22$ & $22.64$ & $19.64$ & $61.59$ & $46.33$ & $40.35$ & $67.49$ & $50.26$ & $43.98$\\
                \methodName & 3 & $24.12$ & $16.37$ & $14.48$ & $31.58$ & $22.52$ & $19.65$ & $62.51$ & $46.47$ & $40.65$ & $67.33$ & $50.24$ & $44.16$\\ 
                % & 4 & \best{24.63} & \best{16.54} & \best{14.52} & \best{32.60} &	\best{23.04} & \best{19.99} &	\best{61.00} & \best{46.00}	& \best{40.18} & \best{65.28} & \best{49.63} & \best{43.50} \\
                & 4 & $24.63$ & $16.54$ & $14.52$ & $32.60$ &	$23.04$ & $19.99$ &	$61.00$ & $46.00$	& $40.18$ & $65.28$ & $49.63$ & $43.50$ \\
                & 5 & $25.82$ & $17.69$ & $15.07$ & $33.63$ & $23.84$ & $20.60$ & $62.39$ & $46.46$ & $40.61$ & $67.55$ & $50.51$ & $45.80$\\
                \cline{2-14}
                & Avg & \best{24.22} & \best{16.51} & \best{14.34} & \best{31.77} & \best{22.79} & \best{19.81}  & \best{61.54} & \best{46.16} & \best{40.33} & \best{66.79} & \best{50.01} & \best{44.16}\\
                \myTopRule
            \end{tabular}
            }
        \end{table}

        %============================================================================
        \noIndentHeading{Experiment Comparison.}
            We now compare the experiments of different papers in \cref{tab:expt_comparison}.
            To the best of our knowledge, 
            the experimentation in \methodName{} is more than the experimentation of most monocular \threeD{} object detection papers.

        \begin{table}[!tb]
            \caption{\textbf{Experiments Comparison}.
            }
            \label{tab:expt_comparison}
            \centering
            \scalebox{\scaleFraction}{
                \footnotesize
                \rowcolors{3}{lightgray}{white}
                \setlength\tabcolsep{0.1cm}
                \begin{tabular}{m l m c m ccc m}
                    \myTopRule
                    Method & Venue & Multi-Dataset & Cross-Dataset & Multi-Backbone\\
                    \myTopRule
                    \groomedNMS \cite{kumar2021groomed} & CVPR21 & \mathDash{} & \mathDash{} & \mathDash{}\\
                    MonoFlex \cite{zhang2021objects} & CVPR21 & \mathDash{} & \mathDash{} & \mathDash{}\\
                    \caddn \cite{reading2021categorical} & CVPR21 & \cmark & \mathDash{} & \mathDash{}\\
                    \hline
                    MonoRCNN \cite{shi2021geometry} & ICCV21 & \mathDash{} & \cmark & \mathDash{}\\
                    \gupNet{} \cite{lu2021geometry} & ICCV21 & \mathDash{} & \mathDash{} & \mathDash{}\\
                    \ddThreeD \cite{park2021pseudo} & ICCV21 & \cmark & \mathDash{} & \cmark\\
                    \hline
                    PCT \cite{wang2021progressive} & NeurIPS21 & \cmark & \mathDash{} & \cmark\\
                    MonoDistill \cite{chong2022monodistill} & ICLR22 & \mathDash{} & \mathDash{} & \mathDash{}\\
                    \hline
                    \monoDISMulti \cite{simonelli2020disentangling} & TPAMI20 & \cmark & \mathDash{} & \mathDash{}\\
                    MonoEF \cite{zhou2021monoef} & TPAMI21 & \cmark & \mathDash{} & \mathDash{}\\
                    \hline
                    \best{\methodName} & - & \cmark & \cmark & \cmark\\
                    \myTopRule
                \end{tabular}
            }
        \end{table}

    %============================================================================
    %============================================================================
    \subsection{Qualitative Results}

\begin{figure*}[!tb]
            \centering
            \begin{subfigure}{\figureScaleFraction\linewidth}
                \includegraphics[width=\linewidth]{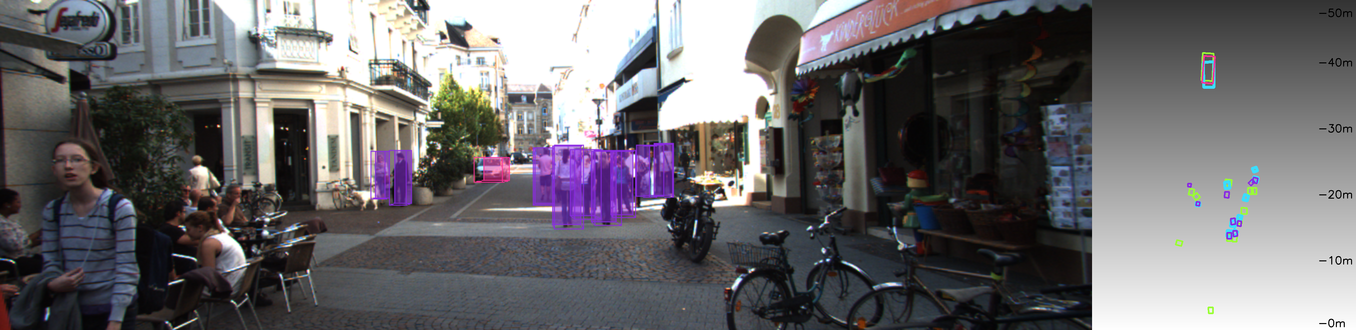}
            \end{subfigure}
            \begin{subfigure}{\figureScaleFraction\linewidth}
                \includegraphics[width=\linewidth]{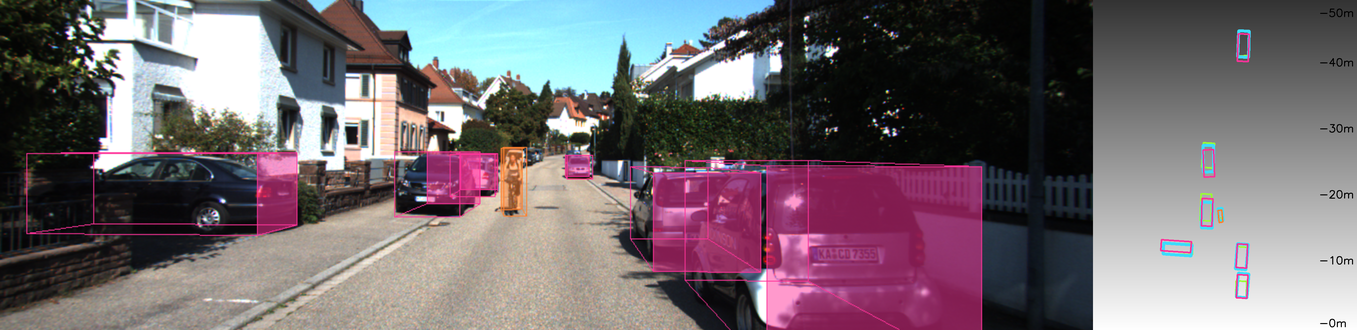}
            \end{subfigure}
            \begin{subfigure}{\figureScaleFraction\linewidth}
                \includegraphics[width=\linewidth]{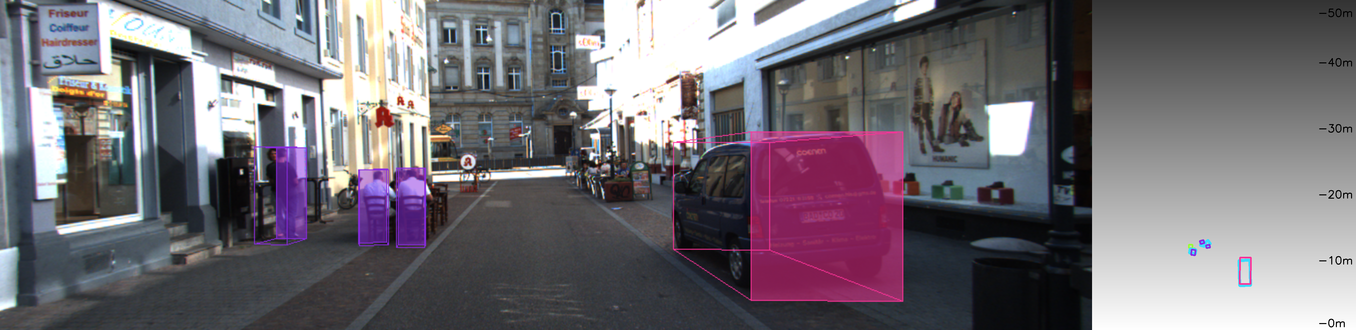}
            \end{subfigure}
            \begin{subfigure}{\figureScaleFraction\linewidth}
                \includegraphics[width=\linewidth]{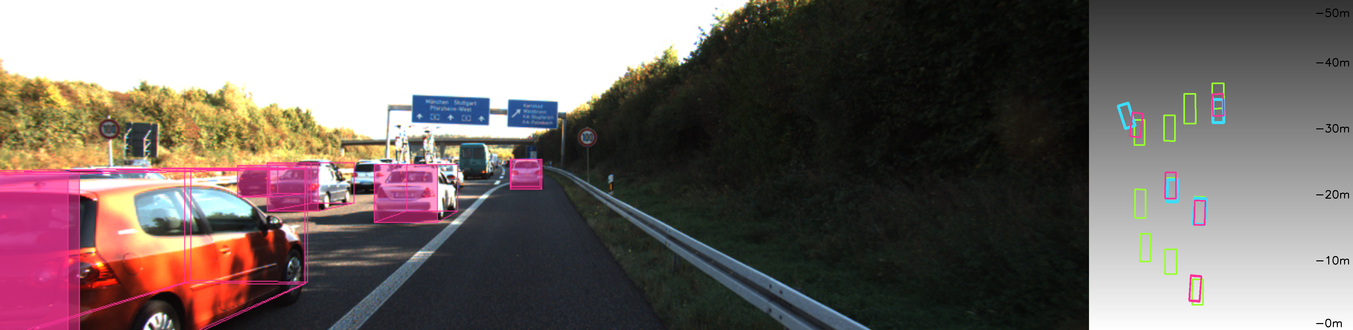}
            \end{subfigure}
            \begin{subfigure}{\figureScaleFraction\linewidth}
                \includegraphics[width=\linewidth]{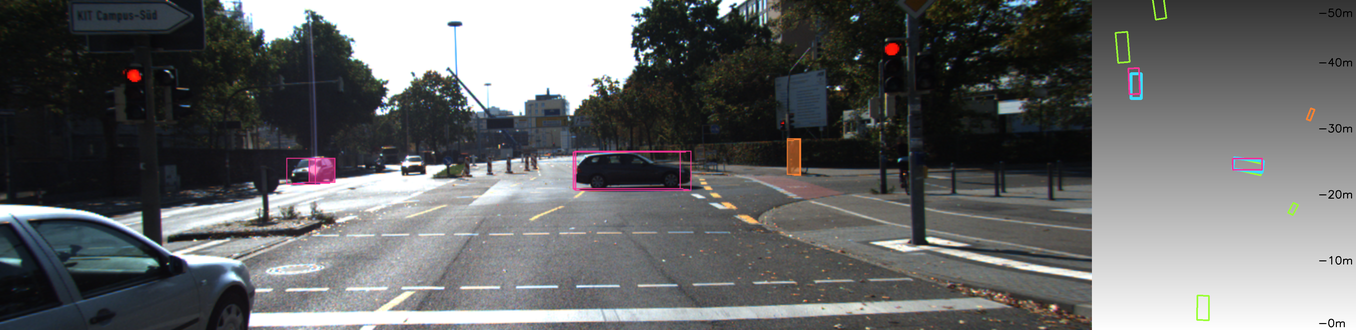}
            \end{subfigure}
            \begin{subfigure}{\figureScaleFraction\linewidth}
                \includegraphics[width=\linewidth]{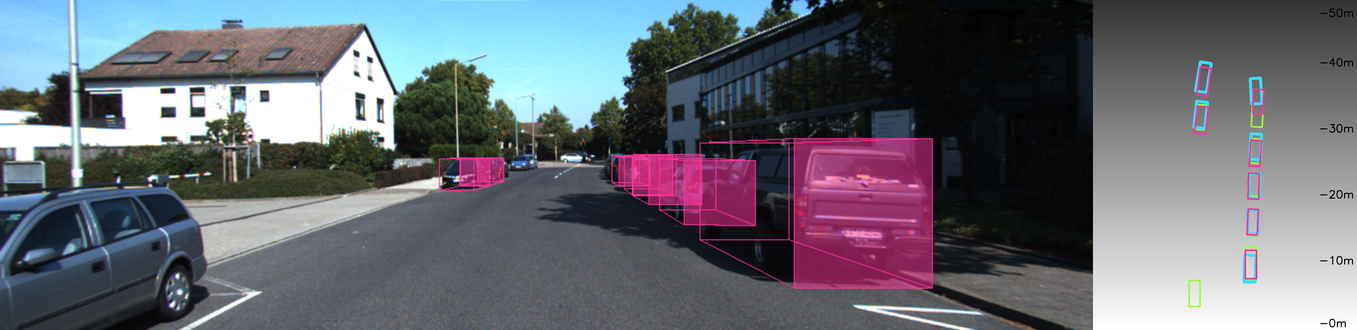}
            \end{subfigure}
            \caption{\textbf{\kitti{} Qualitative Results}. 
            \methodName{} predictions in general are more accurate than \textcolor{set1_cyan}{\gupNet} \cite{lu2021geometry}.
            [Key: \textcolor{my_magenta}{Cars}, \textcolor{orange}{Cyclists} and \textcolor{violet}{Pedestrians} of \methodName; \textcolor{set1_cyan}{all classes of \gupNet}, and \textcolor{my_green}{Ground Truth} in BEV]. }
            \label{fig:qualitative_kitti}
        \end{figure*}

        %============================================================================
        \noIndentHeading{\kitti{}.}
            We next show some more qualitative results of models trained on \kitti{} \valOne{} split in \cref{fig:qualitative_kitti}. 
            We depict the predictions of \methodName{} in image view on the left and the predictions of \methodName{} and \gupNet{} \cite{lu2021geometry}, and ground truth in BEV on the right. 
            In general, \methodName{} predictions are more closer to the ground truth than \gupNet{} \cite{lu2021geometry}.
            
\begin{figure*}[!tb]
            \centering
            % left bottom right top
            \begin{subfigure}{\figureScaleFraction\linewidth}
                \includegraphics[trim={0 2.5cm 0 5.0cm},clip,width=\linewidth]{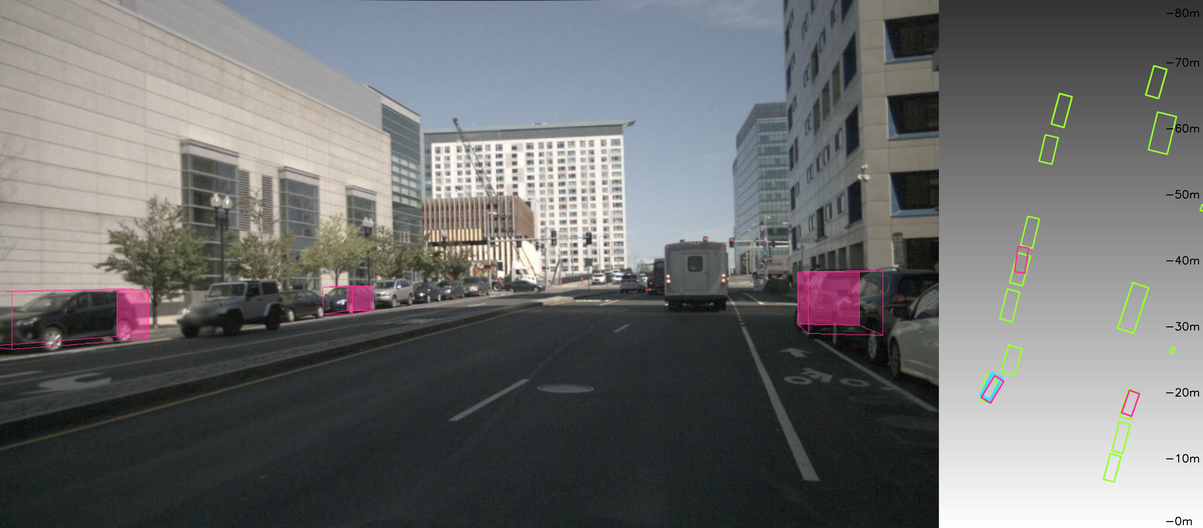}
            \end{subfigure}
            \begin{subfigure}{\figureScaleFraction\linewidth}
                \includegraphics[trim={0 2.5cm 0 5.0cm},clip,width=\linewidth]{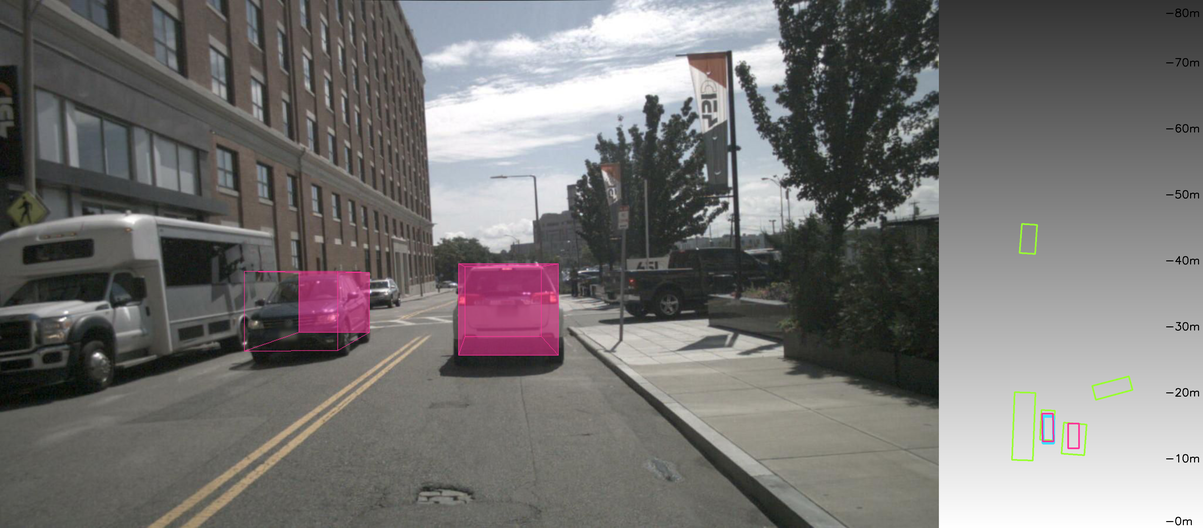}
            \end{subfigure}
            \begin{subfigure}{\figureScaleFraction\linewidth}
                \includegraphics[trim={0 2.5cm 0 5.0cm},clip,width=\linewidth]{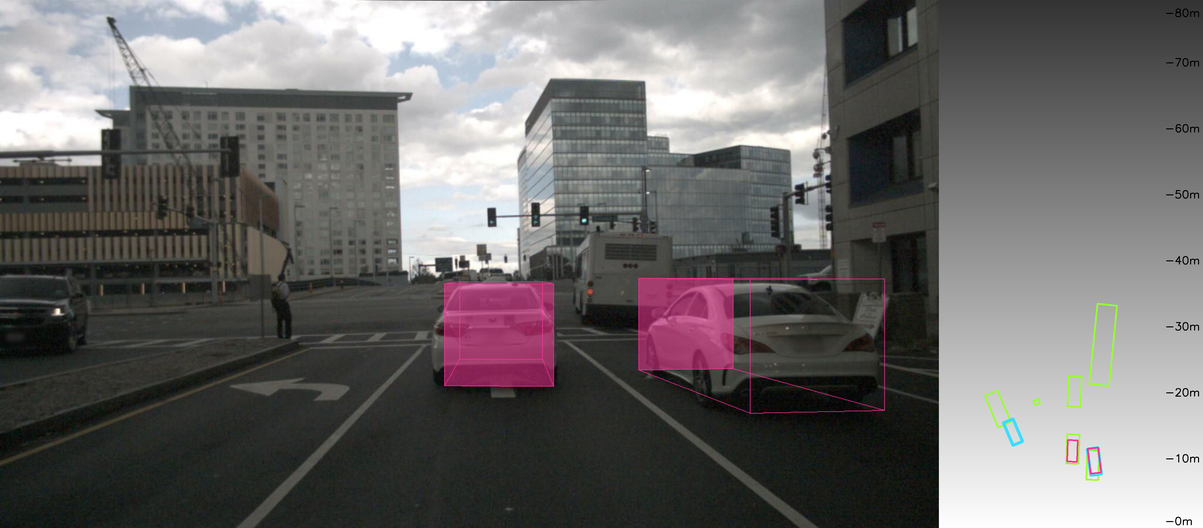}
            \end{subfigure}
            \begin{subfigure}{\figureScaleFraction\linewidth}
                \includegraphics[trim={0 2.5cm 0 5.0cm},clip,width=\linewidth]{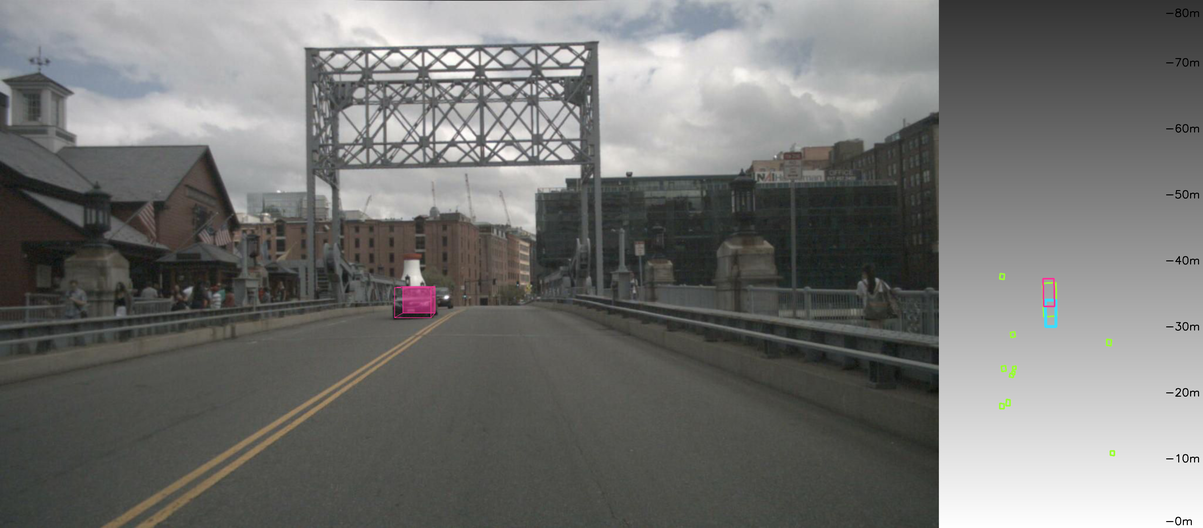}
            \end{subfigure}
            \begin{subfigure}{\figureScaleFraction\linewidth}
                \includegraphics[trim={0 2.5cm 0 5.0cm},clip,width=\linewidth]{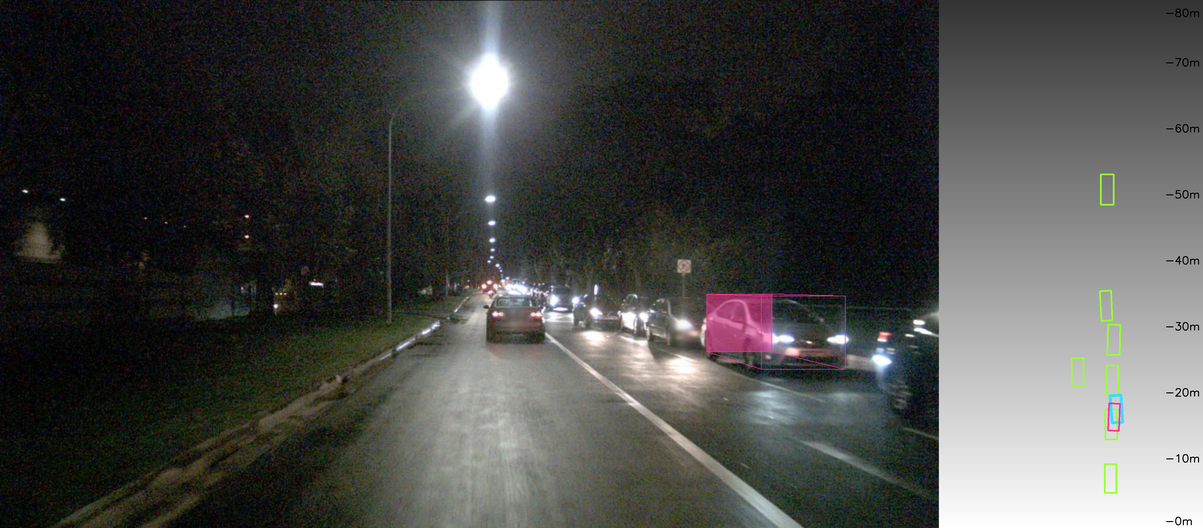}
            \end{subfigure}
            \begin{subfigure}{\figureScaleFraction\linewidth}
                \includegraphics[trim={0 2.5cm 0 5.0cm},clip,width=\linewidth]{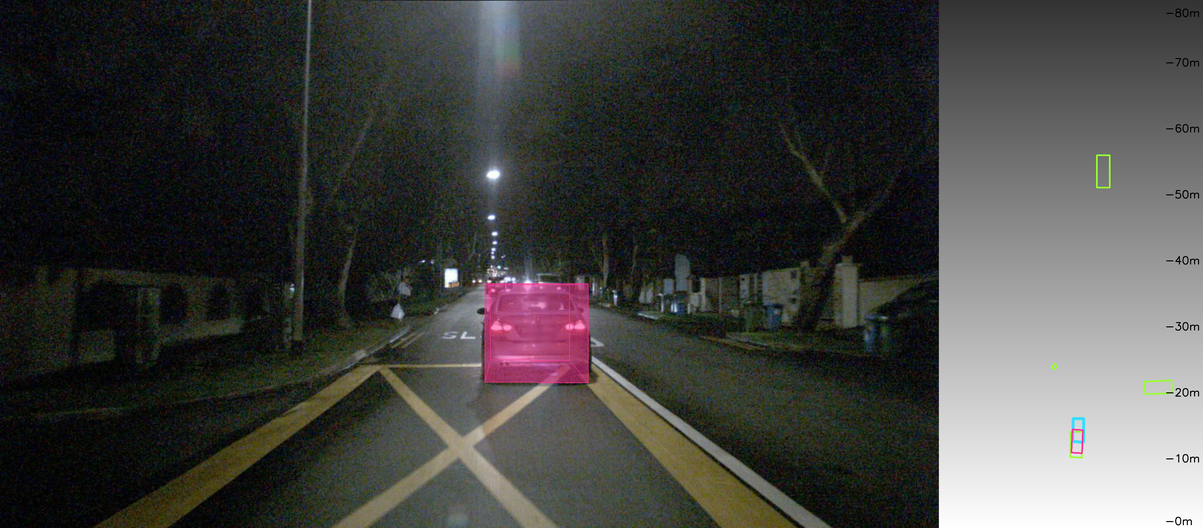}
            \end{subfigure}
            \caption{\textbf{\nuscenes{} Cross-Dataset Qualitative Results.}
            \methodName{} predictions in general are more accurate than \textcolor{set1_cyan}{\gupNet} \cite{lu2021geometry}.
            [Key: \textcolor{my_magenta}{Cars} of \methodName; \textcolor{set1_cyan}{Cars} of \gupNet, and \textcolor{my_green}{Ground Truth} in BEV]. }
            \label{fig:qualitative_nusc_kitti}
        \end{figure*}

        %============================================================================
        \noIndentHeading{\nuscenes{} Cross-Dataset Evaluation.}
            We then show some qualitative results of \kitti{} \valOne{} model evaluated on \nuscenes{} frontal in \cref{fig:qualitative_nusc_kitti}. 
            We again observe that \methodName{} predictions are more closer to the ground truth than \gupNet{} \cite{lu2021geometry}.
            Also, considerably less number of boxes are detected in the cross-dataset evaluation \thatIs{} on \nuscenes{}. 
            We believe this happens because of the domain shift.
            
\begin{figure*}[!tb]
            \centering
            % left bottom right top
            \begin{subfigure}{\waymoFigureScaleFraction\linewidth}
                \includegraphics[trim={0 2cm 0 2cm},clip,width=\linewidth]{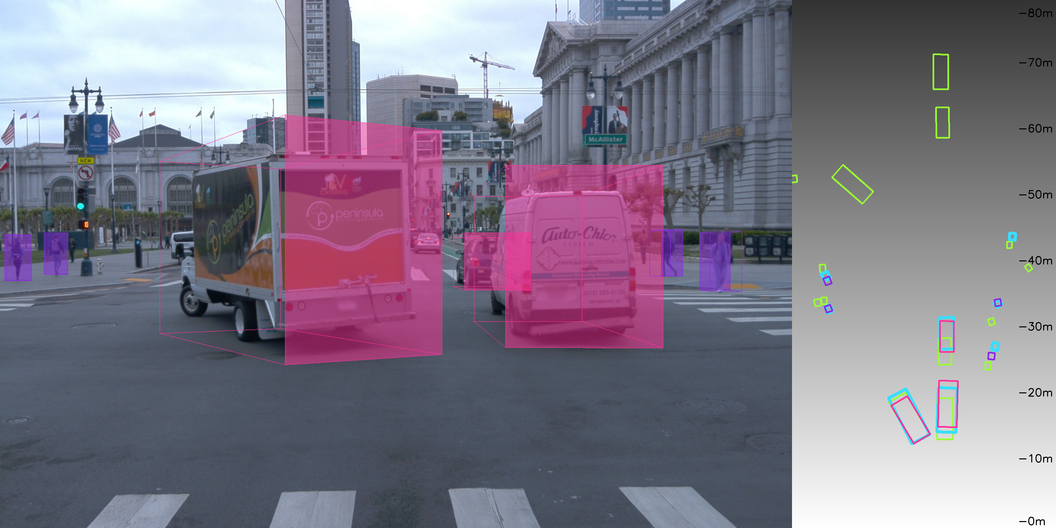}
            \end{subfigure}
            \begin{subfigure}{\waymoFigureScaleFraction\linewidth}
                \includegraphics[trim={0 2.0cm 0 2.0cm},clip,width=\linewidth]{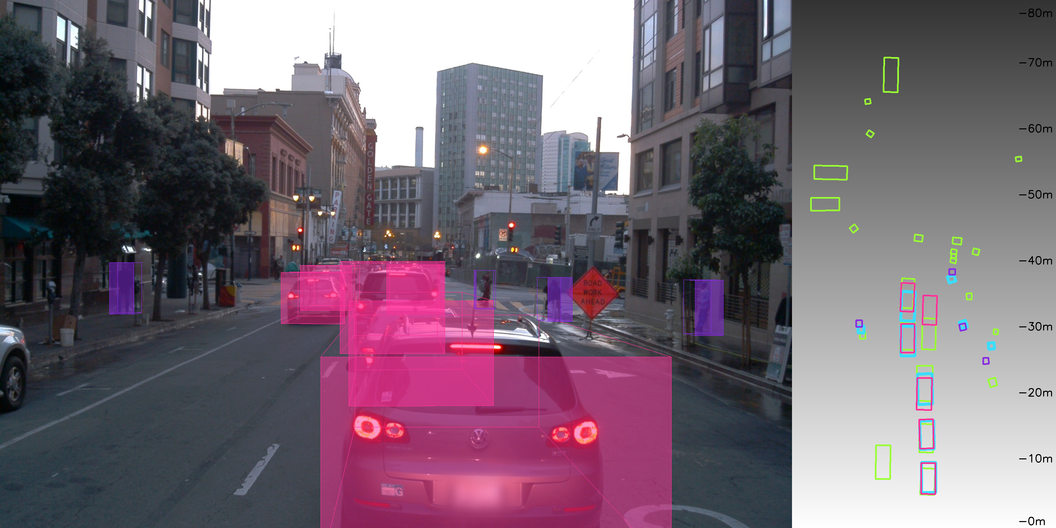}
            \end{subfigure}
            \begin{subfigure}{\waymoFigureScaleFraction\linewidth}
                \includegraphics[trim={0 2.0cm 0 2.0cm},clip,width=\linewidth]{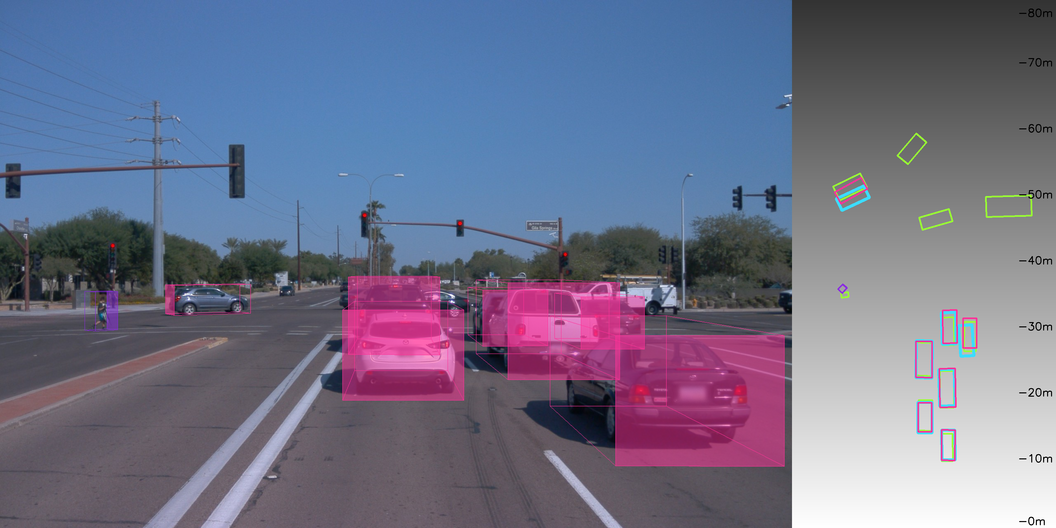}
            \end{subfigure}
            \begin{subfigure}{\waymoFigureScaleFraction\linewidth}
                \includegraphics[trim={0 2.0cm 0 2.0cm},clip,width=\linewidth]{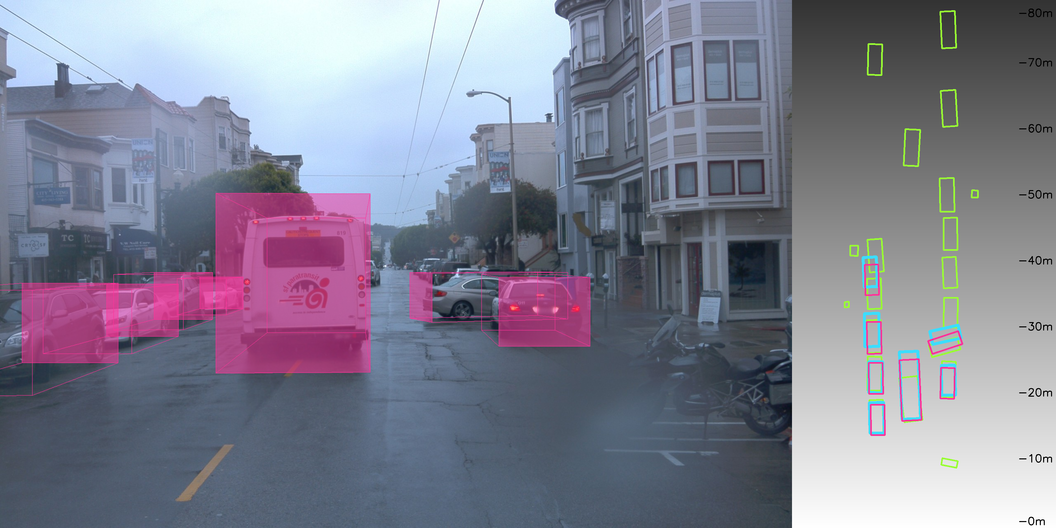}
            \end{subfigure}
            \begin{subfigure}{\waymoFigureScaleFraction\linewidth}
                \includegraphics[trim={0 2cm 0 2cm},clip,width=\linewidth]{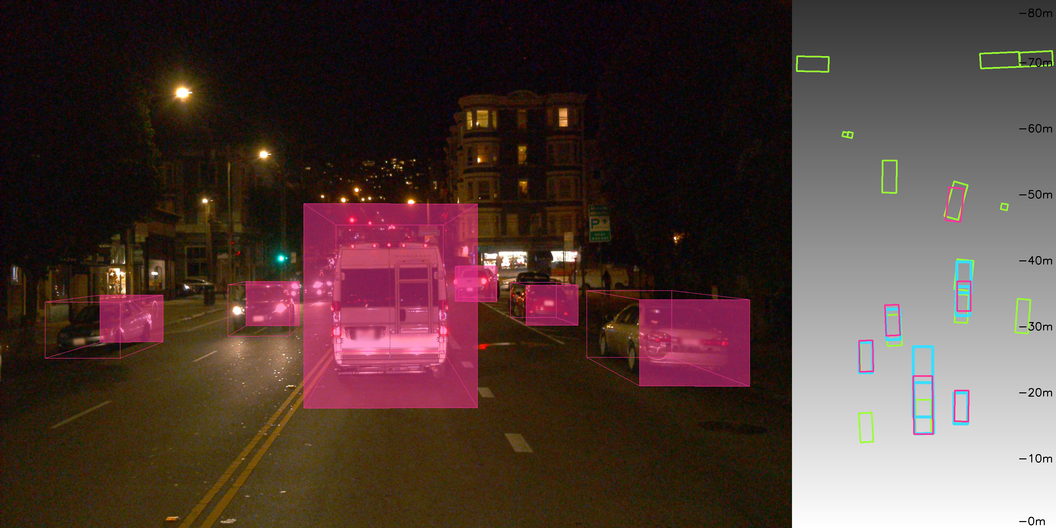}
            \end{subfigure}
            \begin{subfigure}{\waymoFigureScaleFraction\linewidth}
                \includegraphics[trim={0 2.0cm 0 2.0cm},clip,width=\linewidth]{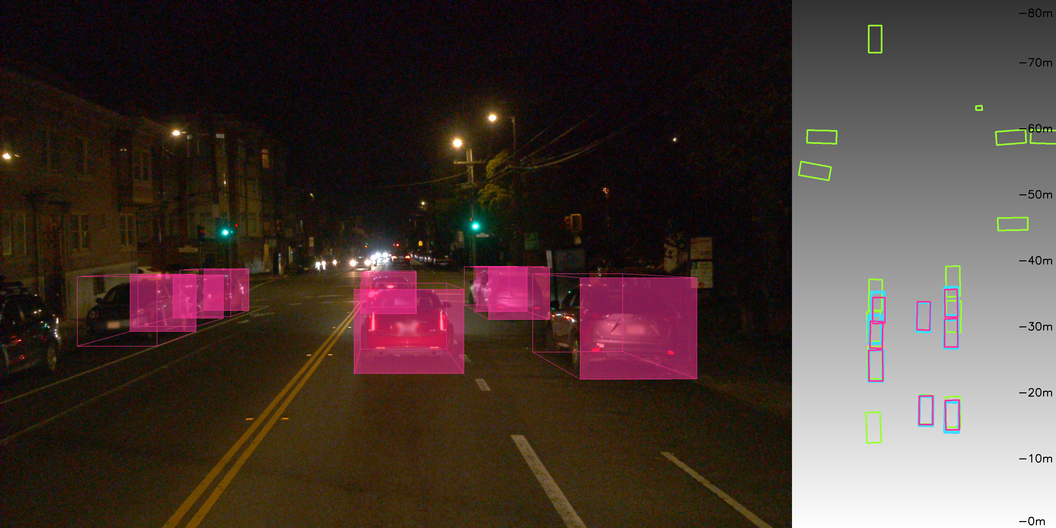}
            \end{subfigure}
            \caption{
            \textbf{\waymo{} Qualitative Results}. 
            \methodName{} predictions in general are more accurate than \textcolor{set1_cyan}{\gupNet} \cite{lu2021geometry}.
            [Key: \textcolor{my_magenta}{Cars}, \textcolor{orange}{Cyclists} and \textcolor{violet}{Pedestrians} of \methodName; \textcolor{set1_cyan}{all classes of \gupNet}, and \textcolor{my_green}{Ground Truth} in BEV].}
            \label{fig:qualitative_waymo}
        \end{figure*}

        %============================================================================
        \noIndentHeading{\waymo{}.}
            We now show some qualitative results of models trained on \waymo{} \val{} split in \cref{fig:qualitative_waymo}. 
            We again observe that \methodName{} predictions are more closer to the ground truth than \gupNet{} \cite{lu2021geometry}.
            
    \begin{figure}[!tb]
        \centering
        \begin{subfigure}[align=bottom]{.548\linewidth}
              \centering
              \includegraphics[width=\linewidth]{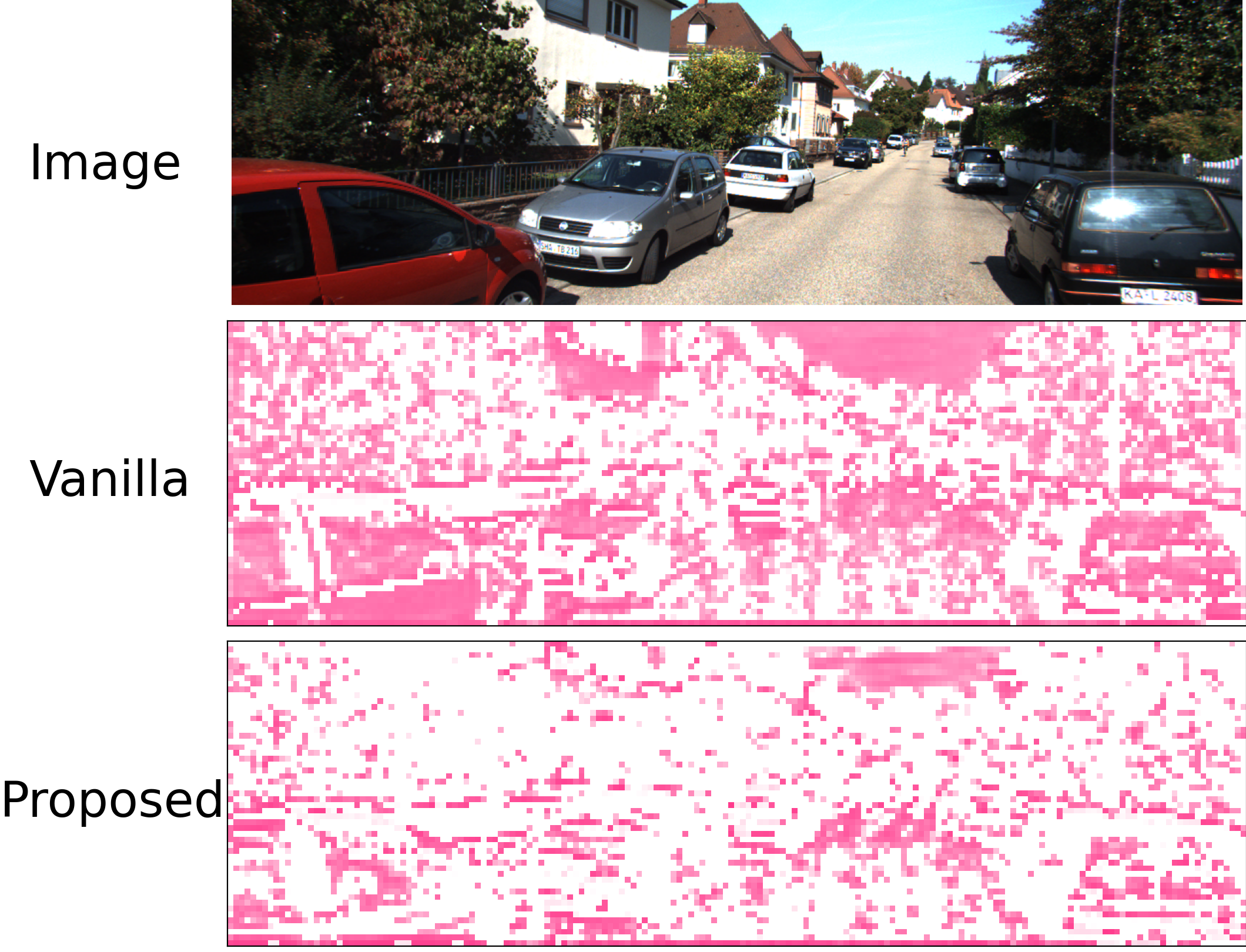}
              \caption{Depth \equivariance{} error (\downarrowRHDSmall).}
              \label{fig:equiv_error_qualitative}
        \end{subfigure}
        \begin{subfigure}[align=bottom]{.38\linewidth}
            \centering
            \includegraphics[width=\linewidth]{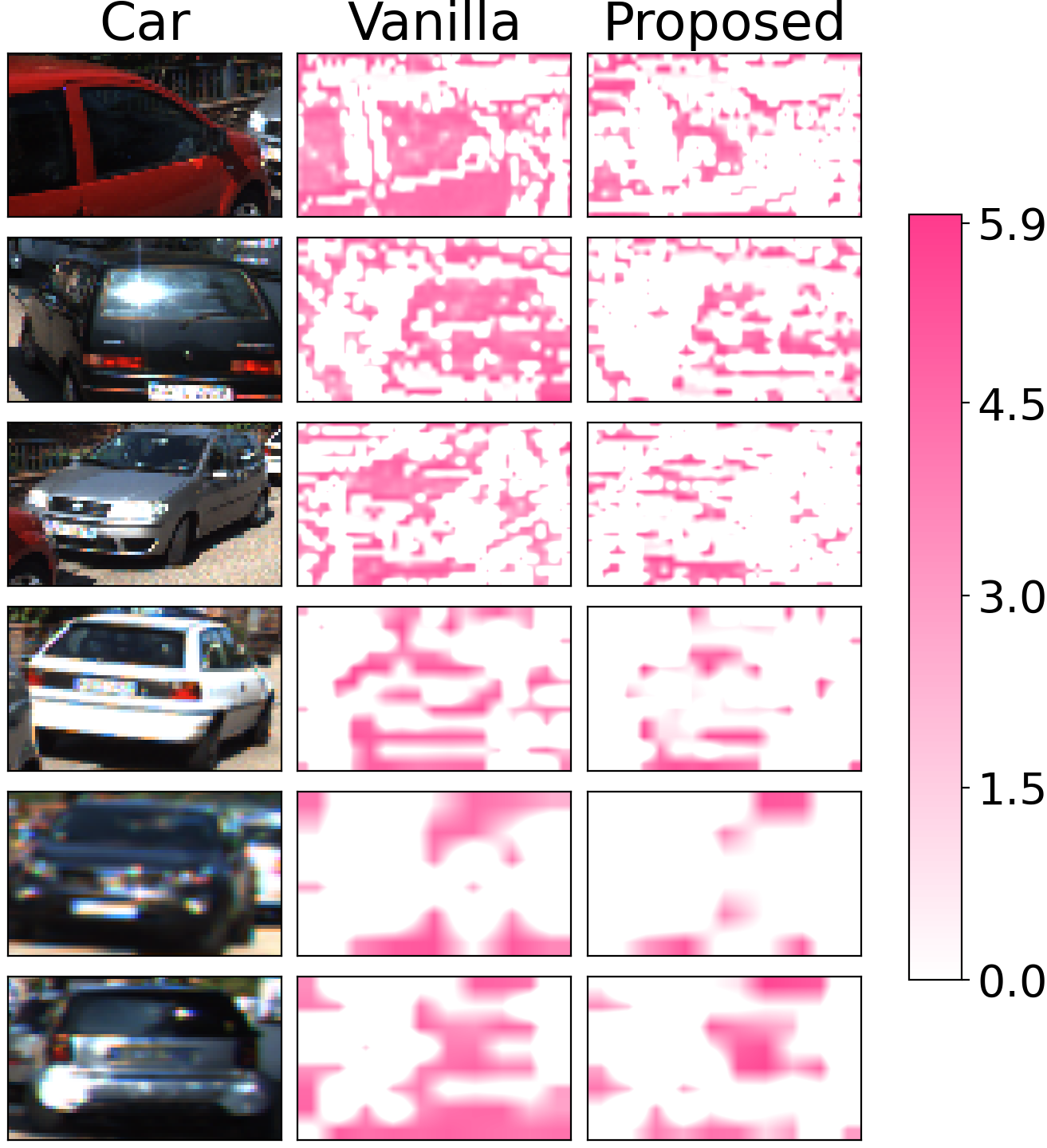}
            \caption{Error~(\downarrowRHDSmall) on objects.}
            \label{fig:equiv_error_qualitative_object}
        \end{subfigure}
        \caption{
        \textbf{(a) Depth (scale) \equivariance{} error } of vanilla \gupNet{} \cite{lu2021geometry} and proposed \methodName{}. (See \cref{sec:detection_results_kitti_val1} for details)
        \textbf{(b) Error on objects.} The proposed backbone has less depth \equivariance{} error than vanilla CNN backbone.
        }
        \label{fig:equiv_error_qualitative_both}
    \end{figure}

    %============================================================================
    %============================================================================
    \subsection{Demo Videos of \methodName}   
        
        %============================================================================
        \noIndentHeading{Detection Demo. }
            We next put a short demo video of our \methodName{} model trained on \kitti{} \valOne{} split at \url{https://www.youtube.com/watch?v=2D73ZBrU-PA}. 
            We run our trained model independently on each frame of \textsc{2011\_09\_26\_drive\_0009} \kitti{} raw \cite{geiger2013vision}.
            The video belongs to the City category of the \kitti{} raw video.
            None of the frames from the raw video appear in the training set of \kitti{} \valOne{} split \cite{kumar2021groomed}.
            We use the camera matrices available with the video but do not use any temporal information. 
            Overlaid on each frame of the raw input videos, we plot the projected \threeD{} boxes of the predictions and also plot these \threeD{} boxes in the BEV.
            We set the frame rate of this demo at $10$
            fps as in \kitti{}.
            The attached demo video demonstrates very stable and impressive results because of the additional equivariance to depth translations in \methodName{} which is absent in vanilla CNNs.
            Also, notice that the orientation of the boxes are stable despite not using any temporal information. 
            
        %============================================================================
        \noIndentHeading{\Equivariance{} Error Demo.}
            We next show the \depthEquivariance{} (\scaleEquivariance{}) error demo of one of the channels from the vanilla \gupNet{} and our proposed method at \url{https://www.youtube.com/watch?v=70DIjQkuZvw}.
            As before, we report at block $3$ of the backbones which corresponds to output feature map of the size $[96, 320]$.
            The \equivariance{} error demo indicates more white spaces which confirms that \methodName{} achieves lower equivariance~error compared to the baseline \gupNet{} \cite{lu2021geometry}. 
            Thus, this demo agrees with \cref{fig:equiv_error_qualitative}. 
            This happens because depth (scale) \equivariance{} is additionally hard-baked into \methodName, while the vanilla \gupNet{} is not \equivariant{} to depth translations (scale \transformation s).


%============================================================================
%============================================================================
%============================================================================
\section*{Acknowledgements}
    This research was partially sponsored by the Ford Motor Company and the Army Research Office (ARO) grant W911NF-18-1-0330. 
    This document's views and conclusions are those of the authors and do not represent the official policies, either expressed or implied, of the Army Research Office or the U.S. government.

    We deeply appreciate Max Welling from the University of Amsterdam for several pointers and discussions on \equivariance{} and projective transformations.
    We also thank Ivan Sosnovik, Wei Zhu, Marc Finzi and Vidit for their inputs on scale \equivariance{}.
    
    Yan Lu and Yunfei Long helped setting up the \gupNet{} \cite{lu2021geometry} and DETR3D \cite{wang2021detr3d} codebases respectively.
    Xuepeng Shi and Li Wang shared details of their cross-dataset evaluation \cite{shi2021geometry} and Waymo experiments \cite{wang2021progressive} respectively.
    Shengjie Zhu helped us with the monocular depth (BTS) \cite{lee2019big} experiments.
    Shengjie Zhu, Vishal Asnani, Yiyang Su, Masa Hu, Andrew Hou and Arka Sadhu proof-read our manuscript.
    %and suggested several changes.
    % We also thank Andrea Simonelli and Marcos Paul Gerardo Castro for discussions.
    
    We finally thank anonymous CVPR and ECCV reviewers for their feedback that shaped the final manuscript.
    One anonymous CVPR reviewer pointed out that \cref{th:projective_bigboss} exists as Example 13.2 in \cite{hartley2003multiple}, which we had wrongly claimed as ours in an earlier version.

\end{document}